\documentclass[10pt,twocolumn,letterpaper]{article}

\usepackage{cvpr}
\usepackage{times}
\usepackage{epsfig}
\usepackage{graphicx}
\usepackage{amsmath}
\usepackage{amssymb}

\usepackage{multicol}
\usepackage{caption}
\usepackage{multirow}

\usepackage{algorithmic}
\usepackage{algorithm}
\makeatletter
\newcommand{\removelatexerror}{\let\@latex@error\@gobble}
\makeatother

\usepackage[pagebackref=true,breaklinks=true,letterpaper=true,colorlinks,bookmarks=false]{hyperref}
\usepackage{appendix}

\newcommand{\figref}[1]{Fig.~\ref{#1}}
\newcommand{\tabref}[1]{Tab.~\ref{#1}}

\newcommand{\AlgRef}[1]{Algorithm~\ref{#1}}

\cvprfinalcopy 

\def\cvprPaperID{5815}

\ifcvprfinal\fi
\begin{document}

\twocolumn[{
\begin{@twocolumnfalse}

\title{Self-supervised Learning of Depth Inference for Multi-view Stereo}

\author{Jiayu Yang$^1$, \;\; Jose M. Alvarez$^2$,\;\; Miaomiao Liu$^1$\\
$^1$Australian National University, $^2$NVIDIA\\
{\tt\small \{jiayu.yang, miaomiao.liu\}@anu.edu.au,}\;\;{\tt\small josea@nvidia.com}
}

\maketitle

\begin{center}
\setlength\tabcolsep{0pt}
\vspace{-0.55cm}\begin{tabular}{ccc}
      \includegraphics[width=0.32\linewidth]{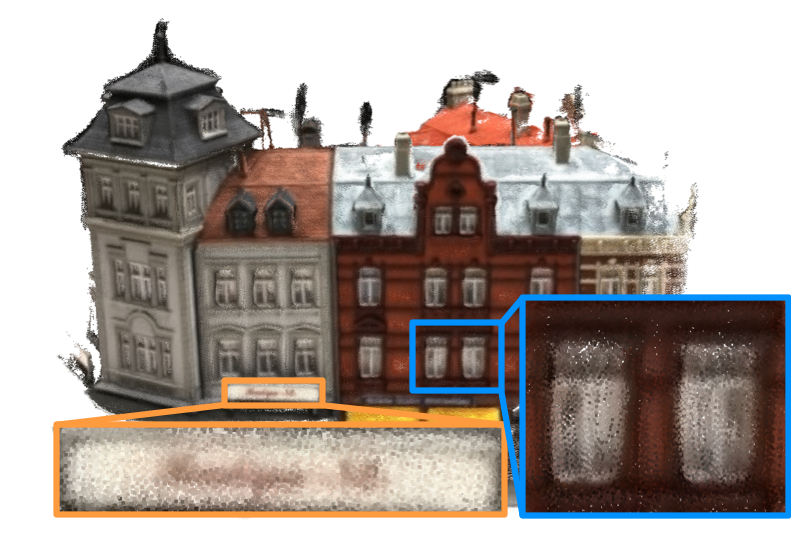}
      & \includegraphics[width=0.32\linewidth]{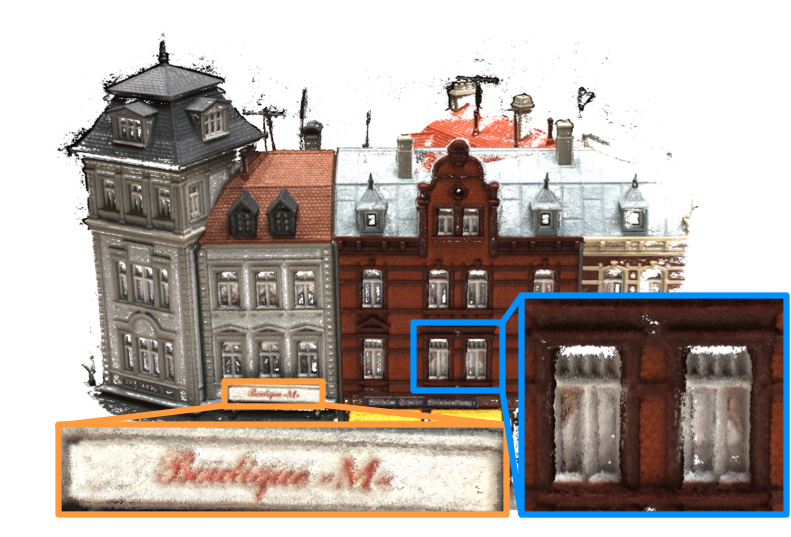}
      & \includegraphics[width=0.32\linewidth]{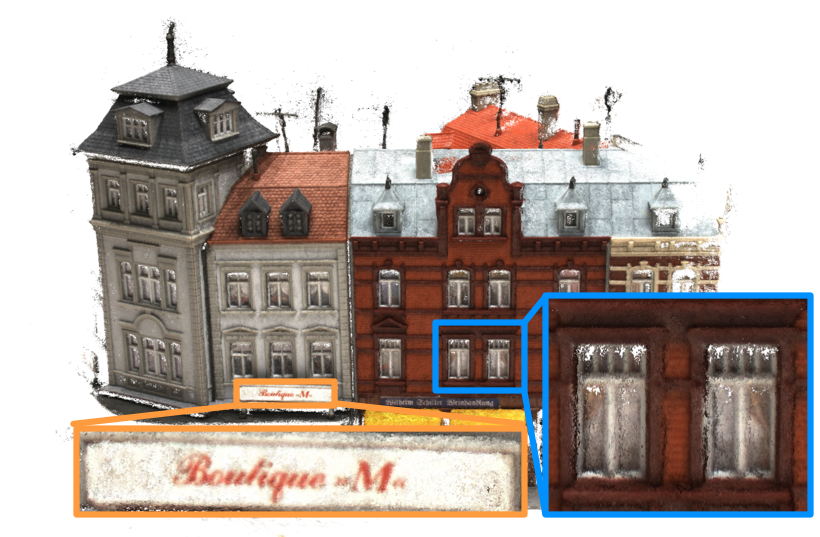}\\
      Khot \etal~\cite{Khot2019learning} & Ours (Unsupervised) & Ours (Self-supervised)
\end{tabular}
\end{center}
\vspace{-0.5cm}
\captionof{figure}{Point cloud reconstructed by existing unsupervised MVS network~\cite{Khot2019learning} and our methods. Best view on screen.}\label{fig:startfig}
\vspace{0.5cm}
\end{@twocolumnfalse}
}]

\thispagestyle{empty}

\begin{abstract}
Recent supervised multi-view depth estimation networks have achieved promising results. Similar to all supervised approaches, these networks require ground-truth data during training. However, collecting a large amount of multi-view depth data is very challenging. 
Here, we propose a self-supervised learning framework for multi-view stereo that exploit pseudo labels from the input data. We start by learning to estimate depth maps as initial pseudo labels under an unsupervised learning framework relying on image reconstruction loss as supervision. We then refine the initial pseudo labels using a carefully designed pipeline leveraging depth information inferred from higher resolution images and neighboring views. We use these high-quality pseudo labels as the supervision signal to train the network and improve, iteratively, its performance by self-training. Extensive experiments on the DTU dataset show that our proposed self-supervised learning framework outperforms existing unsupervised multi-view stereo networks by a large margin and performs on par compared to the supervised counterpart. Code is available at \url{https://github.com/JiayuYANG/Self-supervised-CVP-MVSNet}.
\end{abstract}

\section{Introduction}
The goal of Multi-view Stereo (MVS) is to reconstruct the 3D model of a scene from a set of images captured at multiple viewpoints. While this problem has been studied for decades~\cite{seitz2006comparison}, the current best performance is
achieved by cost-volume based supervised deep neural networks for MVS~\cite{yao2018mvsnet,yao2019recurrent,Yang2020CVP,gu2019cas,cheng2020ucsnet,xu2020pvsnet}. The success of these networks mainly relies on large amount of ground truth depth as training data, which is generally captured by expensive and multiple synchronized image and depth sensors.

The use of synthetic data is considered a good alternative to handle the main challenges in collecting training data for MVS~\cite{yao2020blendedmvs}. Given a set of 3D scene models with a proper setting of lighting conditions, we can obtain a large number of synthetic multiple view images with ground truth depths~\cite{yao2020blendedmvs}. While it is possible to train the network using this synthetic data, for successfully deploying the model in~\emph{real scenes}, we still require to fine-tune the model using data from the target domain~\cite{mallick2020adapt}. Another alternative is adopting an unsupervised learning strategy~\cite{Dai2019mvs2,Khot2019learning}. In this case, the few existing unsupervised MVS approaches use an image reconstruction loss to supervise the training process. This training strategy heavily relies on image colors' photometric consistency for multiple views images, which is sensitive to illumination changes. While both alternatives remove the dependency on depth labels, their performance is far inferior compared to their corresponding supervised counterparts on the target domain.

In this paper, we propose a self-supervised learning framework for depth inference from multi-view images. Our goal is to generate high-quality depth maps as pseudo labels for training the network only from multiple view images. To this end, we first rely on an image reconstruction loss to supervise the training of a cost-volume based depth inference network. We then use this unsupervised network to infer depth maps as pseudo labels for self-supervision~\cite{Lee2013pseudolabel}. While our unsupervised network can estimate accurate depth for pixels with rich textures and satisfying color consistency across views, these pseudo depth labels still contain a large amount of noise.

To refine the pseudo labels, we first propose to infer depth from a higher resolution image than the required training image to obtain depth estimates of higher accuracy for trustful pixels. Then, we filter depth with large errors by leveraging depth information from neighboring views and, finally, use multi-view depth fusion, mesh generation, and depth rendering to fill in the incomplete pseudo depth labels. With our carefully designed pipeline, we improve the pseudo labels' quality; and use them for training the network, improving its performance within a few iterations. 

Our contributions can be summarized as follows:
\begin{itemize}
    \item We propose a self-supervised learning framework for multi-view depth estimation.
    \item We generate an initial set of pseudo depth labels from an unsupervised learning network and then improve their quality with a carefully designed pipeline to use them to supervise the network yielding performance improvements.
\end{itemize}
Our extensive set of experiments demonstrate that the proposed self-supervised framework outperforms existing unsupervised MVS networks by a large margin and performs on par compared to the supervised counterpart. 

\section{Related Works}

\noindent\textbf{Supervised Multi-view Stereo.} Recent supervised learning-based multi-view depth estimation networks have shown great potential to replace traditional optimization-based MVS pipelines~\cite{furu2010,tola2012,comp2008,schoenberger2016mvs}. In particular, cost volume-based networks have achieved impressive results for depth inference from multi-view images. For instance, Yao~\etal in~\cite{yao2018mvsnet} propose MVSNet to learn the depth map for each view by constructing a cost volume followed by 3D CNN regularization. While effective in inferring depth for low-resolution images, their framework cannot scale to handle high-resolution images. Follow-up works have focused on reducing the memory requirements of cost volume-based methods. Yao~\etal use a recurrent network to regularize the cost volume in a sequential manner~\cite{yao2019recurrent}, and a few other approaches integrate a coarse-to-fine strategy to construct partial cost volumes~\cite{gu2019cas, Yang2020CVP, cheng2020ucsnet} resulting not only in memory reductions but also achieving higher resolution estimation. All these works, however, focus on designing effective backbones. 

Another line of research~\cite{zhang2020vismvsnet, yi2020pvamvsnet,chen2020vapointmvsnet,xu2020pvsnet} explore multi-view aggregation to further leverage information from multiple view images and improve the performance of the network. Unlike these works mainly focusing on the backbone design or improving the view-aggregation strategy, we focus on self-supervised learning for depth inference. We adopt the backbone network in CVP-MVSNet~\cite{Yang2020CVP}, which is compact and flexible in handling high-resolution images, and leave as future work the introduction of view-aggregation into our framework.

\noindent\textbf{Synthetic Datasets for Multi-view Stereo.} Existing supervised methods rely on ground-truth depth maps for supervision. However, collecting a large amount of high-quality multi-view ground-truth depth data is very challenging. One solution is to use synthetic data for training. For instance, Yao~\etal created BlendedMVS~\cite{yao2020blendedmvs}, a synthetic dataset based on the rendered depth maps and blended images of meshes generated by existing MVS algorithms. This synthetic data is potentially enough for training MVS algorithms; however, algorithms trained on synthetic data inherently suffer from domain differences with real data. To bridge this domain gap, Mallick~\etal~\cite{mallick2020adapt} introduce a self-supervised domain adaptation method for multi-view stereo. This approach improves the model's performance over the model without using domain adaptation; however, its performance on the target domain is still far inferior to its supervised counterpart.

\noindent\textbf{Unsupervised Multi-view Stereo Networks.} Unsupervised learning-based methods have emerged as an alternative to reduce the requirement of ground-truth data~\cite{Dai2019mvs2, Khot2019learning, huang2020m3vsnet}. For instance, Dai~\etal propose the first unsupervised MVS network with a symmetric unsupervised network that enforces cross-view consistency of multi-view depth maps during both training and testing~\cite{Dai2019mvs2}. They use a view synthesis loss and a cross-view consistency loss to minimize the discrepancy between the source image and the reconstructed image and encourage cross-view consistency. Concurrently, Khot \etal propose to utilize photometric consistency for unsupervised training~\cite{Khot2019learning}. They also adopt a similar loss function including a $L_1$ loss between image intensity, a structure similarity (SSIM) loss, and a depth smoothness loss. Very recently, Huang \etal propose the M$^3$VSNet consisting of a multi-metric unsupervised network and a multi-metric loss function that provides comparable performance with the original supervised MVSNet~\cite{yao2018mvsnet}.

While these unsupervised MVS methods do not require ground-truth depth training data, their training strategy relies heavily on the color consistency across multiple views, which is sensitive to environmental lighting changes. As a result, their performance is still compromised compared to their supervised counterpart~\cite{yao2018mvsnet}. By contrast, we focus on self-supervised learning and generating pseudo depth labels from input image data to supervise the network's training.

 \begin{figure*}
	\begin{center}
    \includegraphics[width=0.96\linewidth]{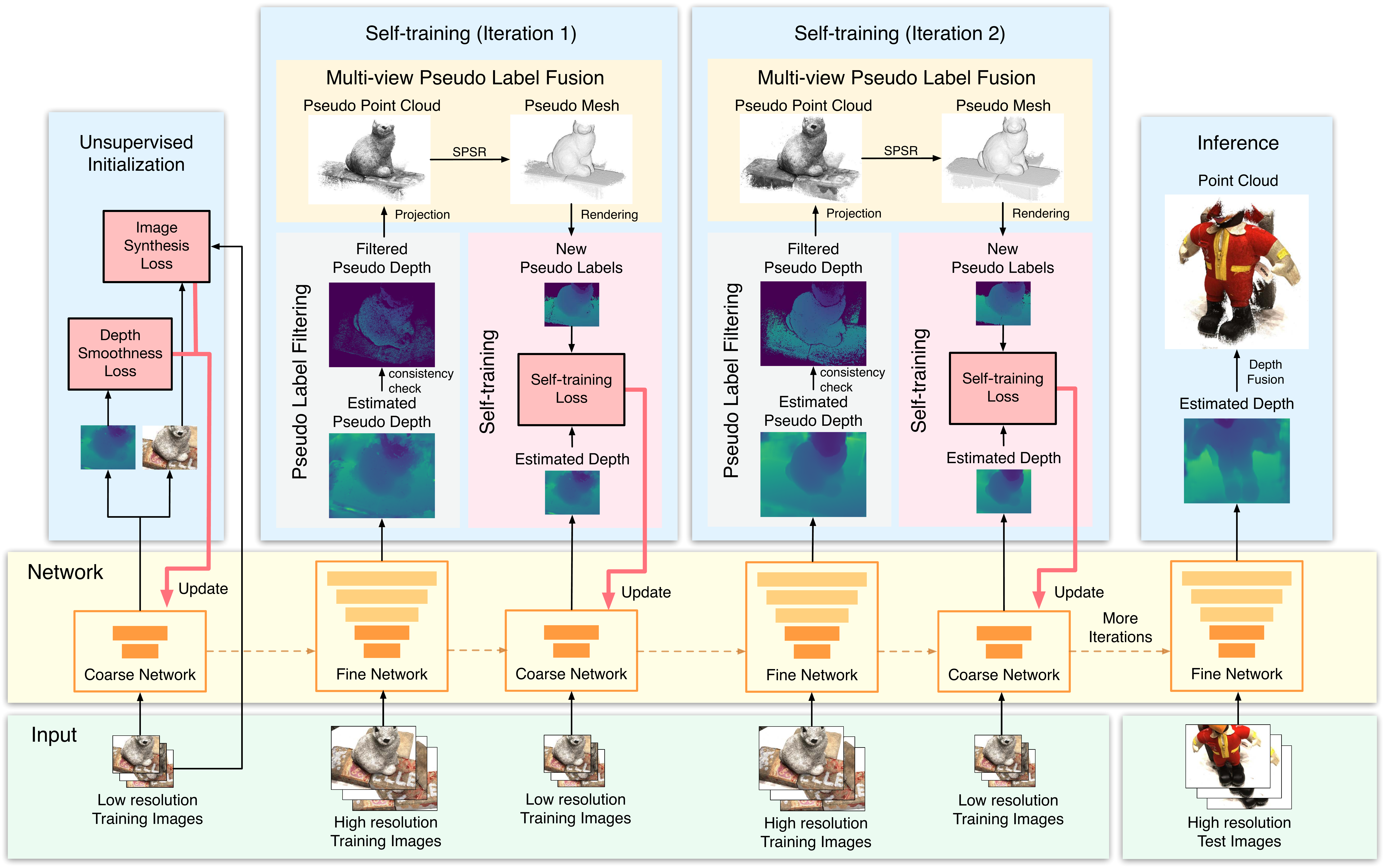}
	\caption{Self-supervised learning framework. We generate the initial pseudo labels by unsupervised learning. We then refine pseudo depth labels from the initial flawed ones and use them to supervise the network iteratively to improve the performance.}
	\label{fig:framework}
	\end{center}
	\vspace{-0.77cm}
\end{figure*}

\section{Method}
We aim to generate high-quality pseudo depth labels from multi-view images for self-supervised learning of depth inference.~To this end, we design a framework consisting of two-stages:~unsupervised learning for initial pseudo label estimation and iterative pseudo label refinement for self-training.  
Below, we first introduce the overall network structure in Section~\ref{sec:networkStr}, and then, in Sections~\ref{section:unsup} and~\ref{section:filtering}, the stages of our framework. The overall framework is depicted in Fig \ref{fig:framework} where we ignore camera parameters for simplicity.

\subsection{Network Structure}~\label{sec:networkStr}
 We apply the recent CVP-MVSNet~\cite{Yang2020CVP} as the backbone network in our framework.~Specifically, CVP-MVSNet takes as input a reference image ${\bf I}_0\in \mathbb{R}^{h\times w}$, source images $\{{\bf I}_i\}_{i=1}^N$ and the corresponding camera intrinsics and extrinsics parameters for all views $\{{\bf K}_i, {\bf R}_i, {\bf t}_i\}_{i=1}^N$ and infers the depth map ${D_0}$ for ${\bf I}_0$. Unlike other cost-volume-based MVS networks, CVP-MVSNet adopts a cost-volume pyramid structure with weight sharing across levels, which can be trained with low-resolution images and still handle any high-resolution image during inference. We follow the same network design as in~\cite{Yang2020CVP} and build a cost-volume pyramid of ($L+1$) levels. 

We formulate our self-training loss as
\begin{equation}
l_{pseudo}=\sum_{l=0}^{L}\sum_{{\bf p}\in \Omega}\|{D}_{\text pseudo}^l({\bf p})-{ D}^{l}({\bf p})\|_1,
\end{equation}
where $\Omega$ is the set of valid pixels associated to pseudo depth labels, ${ D}^l_{\text pseudo}$ and ${D}^{l}$ denote the pseudo depth label and depth estimate at the $l^{th}$ level of the cost volume pyramid, respectively. The quality of the pseudo depth label is crucial for achieving good performance. Next, we introduce the proposed unsupervised learning method to generate initial pseudo-labels, then the pseudo-label refinement process, and the overall self-supervised learning pipeline.

\subsection{Unsupervised Learning for Pseudo Depth Label}\label{section:unsup}
\begin{figure}
	\begin{center}
    \includegraphics[width=0.96\linewidth]{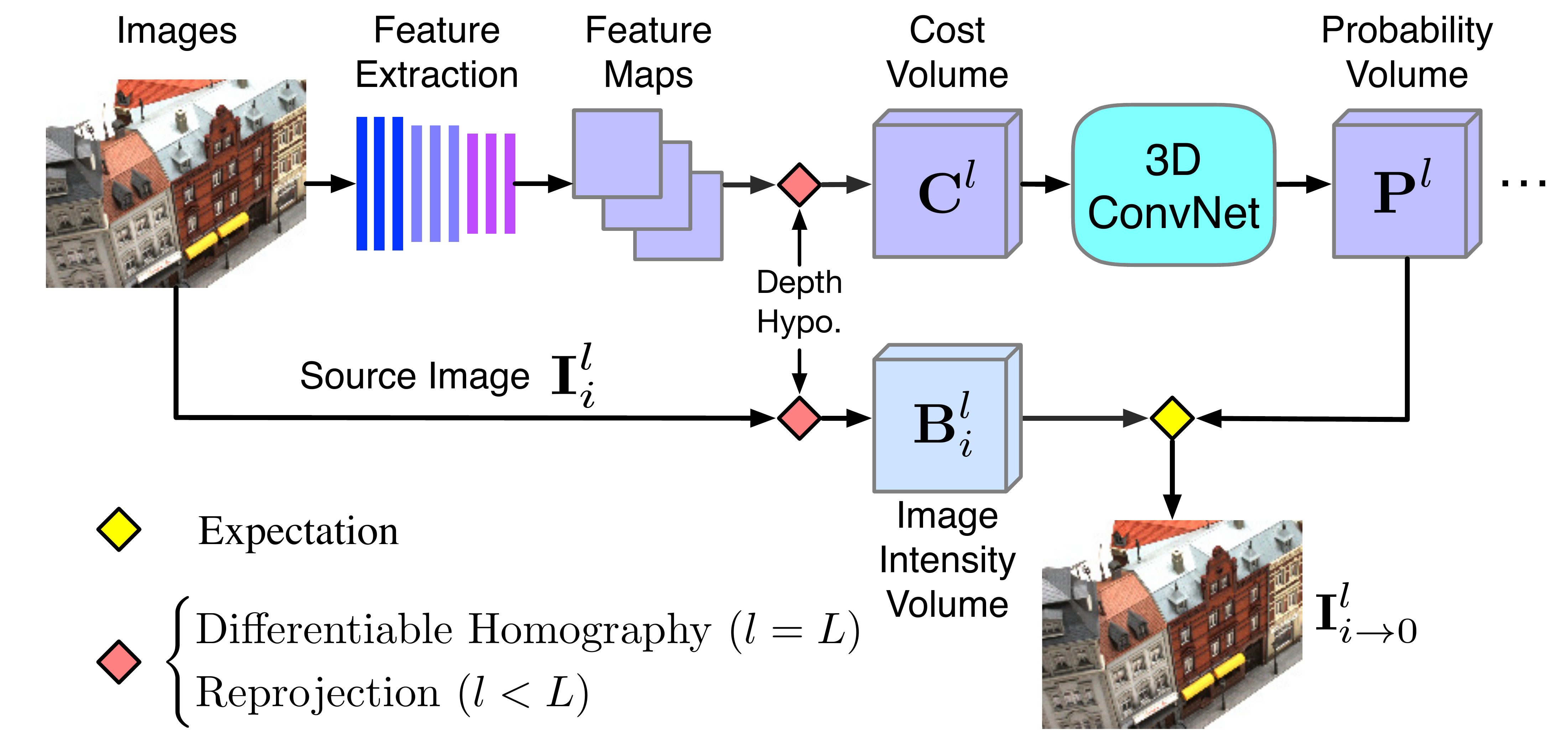}
	\caption{Probability based image synthesis applied on the backbone network~\cite{Yang2020CVP}. We directly synthesize image from probability volume $\textbf{P}^l$ using the image intensity volume $\textbf{B}^l$. }
	\label{fig:prob_syn}
	\end{center}
	\vspace{-0.77cm}
\end{figure}

In the first stage, we learn to estimate depth based on photometric consistency from multi-view images~(see Fig. \ref{fig:framework}). We adopt the CVP-MVSNet as the backbone network and use an image reconstruction loss as supervision signal to train the network. 
Unlike recent unsupervised MVS method~\cite{Dai2019mvs2} that uses the estimated depth map for image synthesis, we, inspired by networks designed for view synthesis~\cite{zhou2018stereomag}, directly synthesize image from the probability distribution of depth hypothesis. To leverage the~\emph{cost-volume pyramid} network structure in~\cite{Yang2020CVP}, we build \emph{image intensity volume pyramid}, $\{{\bf B}_i^l\}_{l=0}^{L}$ based on the warped pixel intensity of each depth hypothesis at each level $l$.
See Fig.~\ref{fig:prob_syn} for an illustration for one source view $i$ and pyramid level $l$.

Specifically, we adopt the differentiable homography defined in~\cite{Yang2020CVP} at each depth hypothesis for image warping at level $L$, and the perspective projection defined in~\cite{Yang2020CVP} for the other levels. Given the~\emph{image intensity volume} $\{{\bf B}_i^l\}_{l=0}^{L}$ and depth hypothesis probability volumes $\{{\bf P}^l\}_{l=0}^L$, we can obtain the synthesized image from source view $i$ as the expectation of warped image intensity based on all depth hypothesis,
\begin{equation}
    \textbf{I}^l_{i\rightarrow 0}(\textbf{x}) = \sum_{d}{\textbf{B}_{i,\textbf{x}}^{l}(d)\textbf{P}_{\textbf{x}}^{l}(d)}
\end{equation}
\vspace{-0.28cm}

\noindent{}where $\textbf{x}=(u,v)$ is a pixel in the reference view, 
$\textbf{B}_{i,\textbf{x}}^{l}(d)\in \mathbb{R}^{3}$ is the intensity of the warped image at pixel $\textbf{x}$ with depth $d$, and $\textbf{P}_{\textbf{x}}^{l}(d)\in[0,1]$ is the probability of pixel $\textbf{x}$ with depth $d$ predicted by the model. 

We explore a view synthesis loss functions similar to~\cite{Dai2019mvs2} to encourage depth smoothness and enforce consistency between the synthesized image and the reference image. We also adopt the perceptual loss proposed in~\cite{johnson2016perceptual} to enforce high-level contextual similarity between the synthesized images and the reference image. Specifically, we use a weighted combination of four loss functions: 

\begin{equation}
    l_{syn} = \alpha_1 l_{g} + \alpha_2 l_{ssim} + \alpha_3 l_{p} + \alpha_4 l_{s},
\end{equation}
\noindent{}where $l_{g}$ is the image gradient loss, $l_{ssim}$ is the structure similarity loss, $l_{p}$ is the perceptual loss, $l_{s}$ is the depth smoothness loss, and $\alpha_i$ sets the influence of each loss - see supplemental material for details of each loss function.

\subsection{Iterative Self-training}
Given the network initially trained in an unsupervised manner, we create an initial set of pseudo depth labels by inferring depth maps for the images in the training set~(see Fig. \ref{fig:framework}). Specifically, the network takes a reference image ${\bf I}_0$, and the source images $\{{\bf I}_i, |\;{\bf I}_i\in \mathbb{R}^{h\times w \times 3}\}_{i=1}^{N}$ as input to learn from the cost volume pyramid and estimate the depth map 
${D}_0 \in \mathbb{R}^{h\times w}$. We obtain the initial pseudo depth labels for the training set as $\{{D}_m|{D}_m\in\mathbb{R}^{h\times w}\}_{m=1}^M$.

As unsupervised learning relies on the image reconstruction loss, which is sensitive to illumination changes, the initial pseudo depth label is subject to a certain noise level.  Next, we describe three stages, refinement from a high-resolution image, pseudo depth filtering by consistency check, and multiple view fusion to improve the quality of the initial set of pseudo depth labels. 

\vspace{1mm}
\noindent{\bf Pseudo Label Refinement from High Resolution Image.}\label{section:ap}
Recall that CVP-MVSNet is a coarse-to-fine depth estimation network with parameter sharing across pyramid levels. Thus, we can evaluate a model trained on low-resolution images and depth pairs on higher resolution ones. 

To improve the quality of pseudo labels, we propose to refine the initial pseudo depth label by using information from a higher resolution training image ${ \{{\bf I}'_m\}_{m=1}^M}\in \mathbb{R}^{H\times W\times 3}$. As evidenced in CVP-MVSNet~\cite{Yang2020CVP}, a higher resolution image carries more discriminative features.~Therefore we can build a cost volume with a smaller depth search interval to further refine the depth map. As shown in Fig.\ref{fig:framework}, we extend the \textit{coarse network} (2 levels) to a \textit{fine network} (5 levels) to further refine the pseudo label and use the refined pseudo label to supervise the original \textit{coarse network} itself as self-training. As we will show, this process improves the accuracy of depth estimate for pixels with rich features and satisfying photometric consistency across views.

However, depth estimates for pixels sensitive to illumination changes or in textureless regions still have large errors. In the following, we detail our approach to filter noise and improve performance.

\vspace{1mm}
\noindent{\bf Pseudo Label Filtering.}\label{section:filtering} Assume the refined pseudo depth label obtained from high-resolution images is $\{{D}'_m|{D}'_m\in\mathbb{R}^{H\times W}\}_{m=1}^M$.
To select reliable depth labels for self-training, we apply a cross-view depth consistency check utilizing depth re-projection error to measure the pseudo depth labels' consistency. To refine the pseudo depth label for each view $D'_i$, we form pairs of views between the reference view $i$ and any other view for the same scene to calculate depth re-projection errors.

\begin{figure}[!t]
	\begin{center}
    \includegraphics[width=0.5\linewidth]{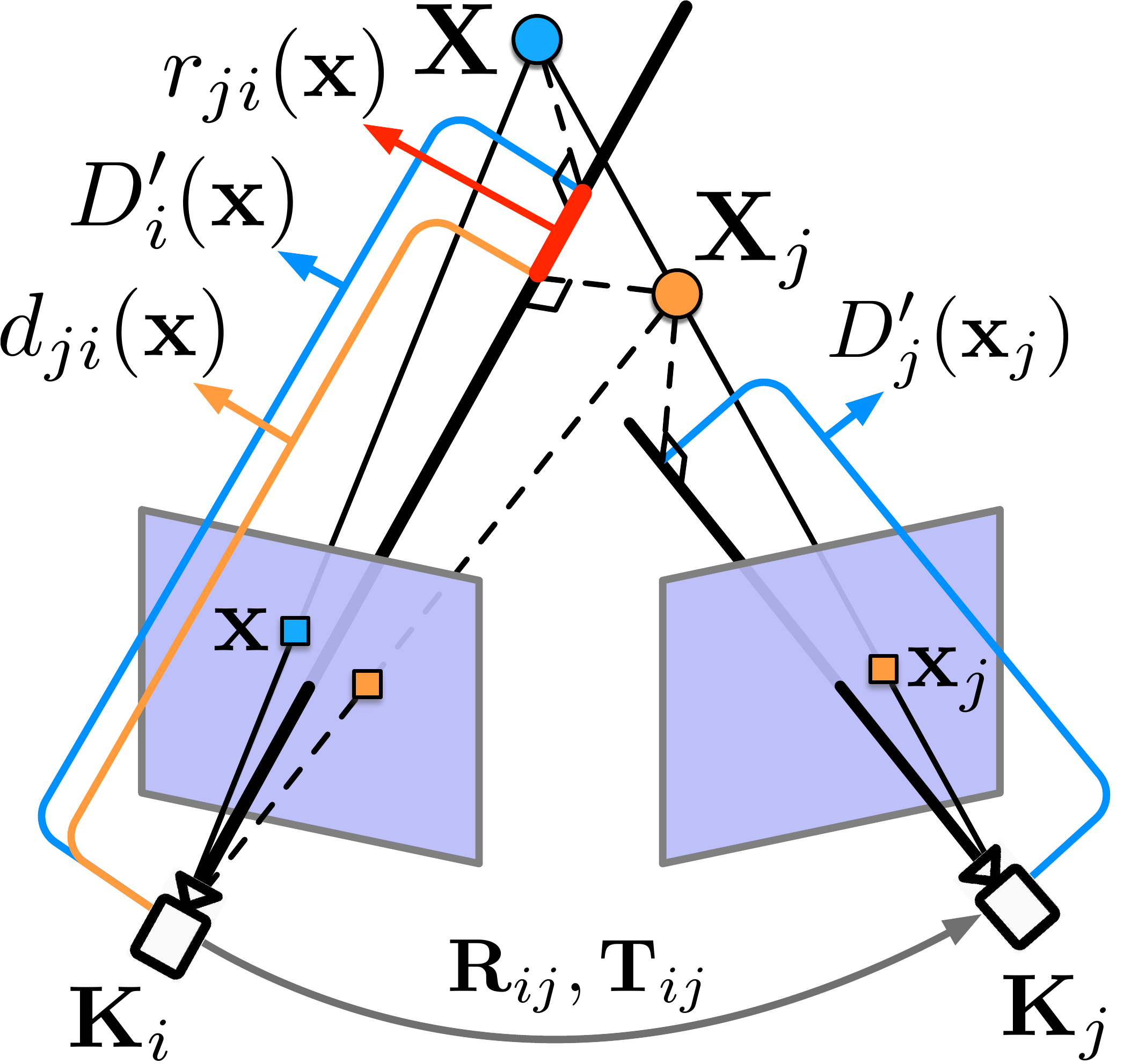}
	\caption{Calculation of depth re-projection error $r_{ji}(\textbf{x})$ for pixel $\textbf{x}$ on reference view $i$ given source view $j$.}
	\label{fig:reproj}
	\end{center}
	\vspace{-0.77cm}
\end{figure}
Here, we provide the calculation of the depth re-projection error between a reference view ${ D}'_i$ and a source view ${D}'_j$.~\figref{fig:reproj} shows the visualization of the depth re-projection error. Assume the camera calibration matrices for view $i$ and $j$ are ${\bf K}_i$ and ${\bf K}_j$, the relative rotation matrix and the translation vector between this pair of views are defined as ${\bf R}_{ij}$ and ${\bf T}_{ij}$, respectively. For each pixel ${\bf x}=(u,v)^{T}$ in the $i$-th view, its corresponding 3D point defined in the camera coordinate system for that view is defined as ${\bf X}={ D}'_i({\bf x}){\bf K}_i^{-1}\bar{{\bf x}}$, where ${\bar {\bf x}}$ is the homogeneous coordinate of ${\bf x}$. Its projection to the $j$-th view is defined as $\lambda_{{\bf x}_j}\bar{{\bf x}}_j={\bf K}_j({\bf R}_{ij}{\bf X}+{\bf T}_{ij})$, where $\bar{{\bf x}}_j$ is the homogeneous coordinate of ${\bf x}_j$. As ${\bf x}_{j}$ might not be integer, we obtain the depth for pixel ${\bf x}_j$, namely ${D}'_j({\bf x}_j)$, by bilinear interpolation. We then obtain the 3D point in the $j$-th view based on ${D}'_j({\bf x}_j)$ as ${\bf X}_j = {D}'_j({\bf x}_j){\bf K}_j^{-1}{\bar{\bf x}}_j$. Therefore, by re-projecting this point back to view $i$, we can obtain ${\bf X}_{ji}={\bf R}_{ij}^{-1}({\bf X}_j-{\bf T}_{ij})$. Its depth in the $i$-th view is defined as $d_{ji}({\bf x})={\bf X}^z_{ji}$, where the superscript $z$ means the $z$-th coordinate of ${\bf X}_{ji}$. Finally, the depth reprojection error for pixel ${\bf x}$ computed from the $j$-th view is defined as $r_{ji}({\bf x})=|D_i'({\bf x}) - d_{ji}({\bf x})|$. 

Assume there are ($M-1$) source views for the current reference view $i$. We compute the set of depth reprojection errors $\{r_{ji}({\bf x})\}_{j=1}^{M-1}$ from ${M-1}$ source views and then define a criterion to filter noisy ones and obtain a refined pseudo depth map $\{D''_j\}_{j=1}^M$ with sparse but accurate depth values. More precisely, $D''_j({\bf x})=D'_i({\bf x})$ iff $\sum_{j=1}^{M-1}q_{ji}({\bf x})>n_{min}$, where $n_{min}$ is the minimum number views of depth consistency, and $q_{ji}({\bf x})$ is the filtering criterion defined as:
\[
    q_{ji}({\bf x})= 
\begin{cases}
    1,& \text{if } r_{ji}({\bf x})\leq r_{max}\\
    0,              & \text{otherwise.}
\end{cases}
\]

\vspace{1mm}
\noindent{\bf Multi-view Pseudo Label Fusion.}\label{section:ca}
We now focus on completing the sparse pseudo depth map resulting from the filtering process. Each view provides a different set of sparse points; therefore, we can combine them to generate a more complete point cloud. To this end, we first project points defined by depth maps from multiple views into the world coordinate system to form a point cloud~(see Fig. \ref{fig:framework}). Specifically, given $M$ filtered high-quality pseudo depth map ${\{{ D}''_m\}_{m=1}^{M}}\in \mathbb{R}^{H\times W}$ of the same scene, we project them into 3D space using their corresponding camera parameters $\{{\bf K}_m, {\bf R}_m, {\bf t}_m\}_{m=1}^M$ to form a pseudo point cloud ${\mathcal X} \in \mathbb{R}^{\Phi\times 3}$, where $\Phi$ is the number of pseudo-3D points corresponding to the aggregation of valid pseudo labels from all views.

Formally, we define the pseudo point cloud from fusing multiple views as 
\begin{equation}
    {\mathcal X} = \{\mathcal{P}_m^{\bf x}|m\in \{1,2,\cdots,M\},{\bf x}\in \Omega_m\},
\end{equation}
\noindent where $\Omega_m$ is the set of pixel coordinates with high quality depth values per image, and ${\mathcal{P}}_m^{{\bf x}} ={\bf R}_m^{-1} (D''_m({\bf x}){\bf K}_m^{-1}{\bar{\bf x}}-{\bf t}_m)$ is the pseudo-3D point for each pixel ${\bf x}$ in the $m$-th view. We then use the Screened Poisson Surface Reconstruction method~\cite{kazhdan2013screened} denote as \textbf{SPSR} to filter out noisy pseudo labels, improve the completeness of {$\mathcal X$}, and generate a mesh $S=\textbf{SPSR}(\mathcal{X})$.

Finally, we render this mesh $S$ into image coordinate of each view as a complete pseudo depth map ${ \{{\bf D}'''_m\}_{m=1}^{M}}\in \mathbb{R}^{h\times w}$ and use them as supervision signal for each view. 

\noindent{\bf Overall self-supervised learning pipeline.}\label{section:pipeline}
Our self-supervised learning framework can be summarised in~\AlgRef{algorithm:selfsup} where we ignore camera parameters for simplicity. Note that we trained the model on low resolution images and depth map pairs. In particular, we render the 3D model to small resolution depth map after each iteration. Such process guarantees the supervision signal for low resolution depth training is of high quality and the performance will not deteriorate rapidly after iterative self-training.

  \begin{algorithm}[!t]
    \caption{Self-supervised Learning Framework}
    \label{algorithm:selfsup}
    \textbf{Input:} $\{{\bf I}_m\}_{m=1}^M\in \mathbb{R}^{h\times w\times 3}$, ${ \{{\bf I}'_m\}_{m=1}^M}\in \mathbb{R}^{H\times W\times 3}$\\
    \textbf{Output:}  Trained model parameters $\varepsilon_{T}$\\
    \textbf{Unsupervised Initialization:}
    
    \begin{algorithmic}[1]
    
      \STATE Train $\varepsilon_0$ using $ \{{\bf I}_m\}_{m=1}^M$ and $l_{syn}$
      
      \hspace{-0.54cm}\textbf{Iterative Self-training:}\\
      \FOR{$t = 1$ to $T$}
      \STATE Inference $\{D^t_m\}_{m=1}^M$ from ${\{{\bf I}_m\}_{m=1}^M}$ using $\varepsilon_{t-1}$
      \STATE Refine $\{D^t_m\}_{m=1}^M$ to $\{D'^t_m\}_{m=1}^M$ \\ ~~using ${\{{\bf I'}_m\}_{m=1}^M}$ and ~$\varepsilon_{t-1}$
      \STATE Filter $\{D'^t_m\}_{m=1}^M$ to $\{D''^t_m\}_{m=1}^N$ by $r_{max}$
      \STATE Project $\{D''^t_m\}_{m=1}^M$ to ${\mathcal{X}}^t$
      \STATE Interpolate ${\mathcal{X}}^t$ to $S^t$ using SPSR.
      \STATE Render $S^t$ to $\{D'''^t_m\}_{m=1}^M\in\mathbb{R}^{h\times w}$
      \STATE Train $\varepsilon_{t}$ using $\{{\bf I}_m\}_{m=1}^M$, $\{D'''^t_m\}_{m=1}^M$ and $l_{pseudo}$
      \ENDFOR
      \RETURN $\varepsilon_{T}$
    \end{algorithmic}
  \end{algorithm}

\section{Experiments}
In this section, we demonstrate the performance of our proposed self-supervised learning framework with a comprehensive set of experiments in standard benchmarks. Below, we first describe the datasets and benchmarks and then analyze our results.

\begin{figure*}[!ht]
    \begin{center}
    \setlength\tabcolsep{0pt}
    \begin{tabular}{rrrrr}
      \includegraphics[width=0.2\linewidth]{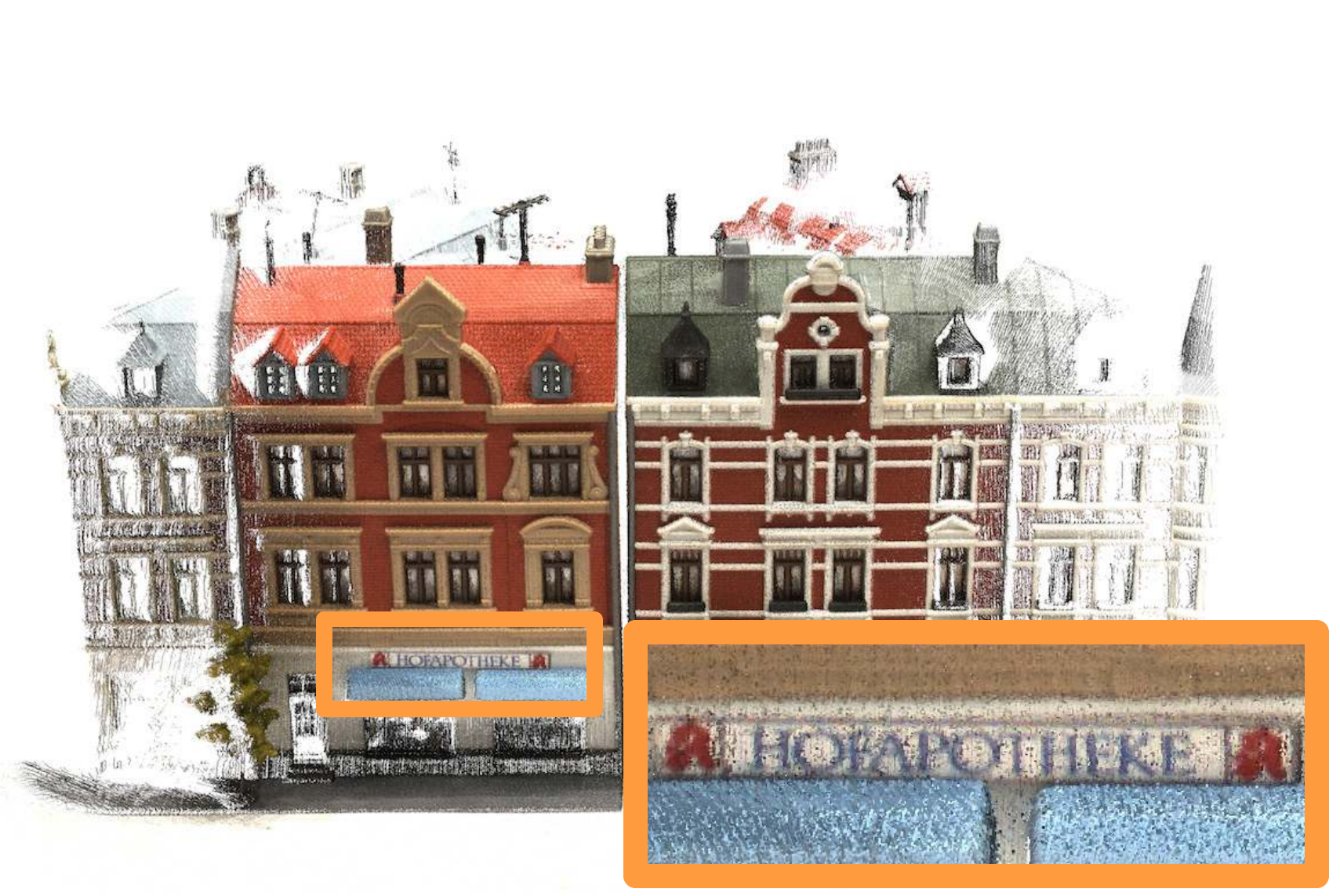}
      & \includegraphics[width=0.2\linewidth]{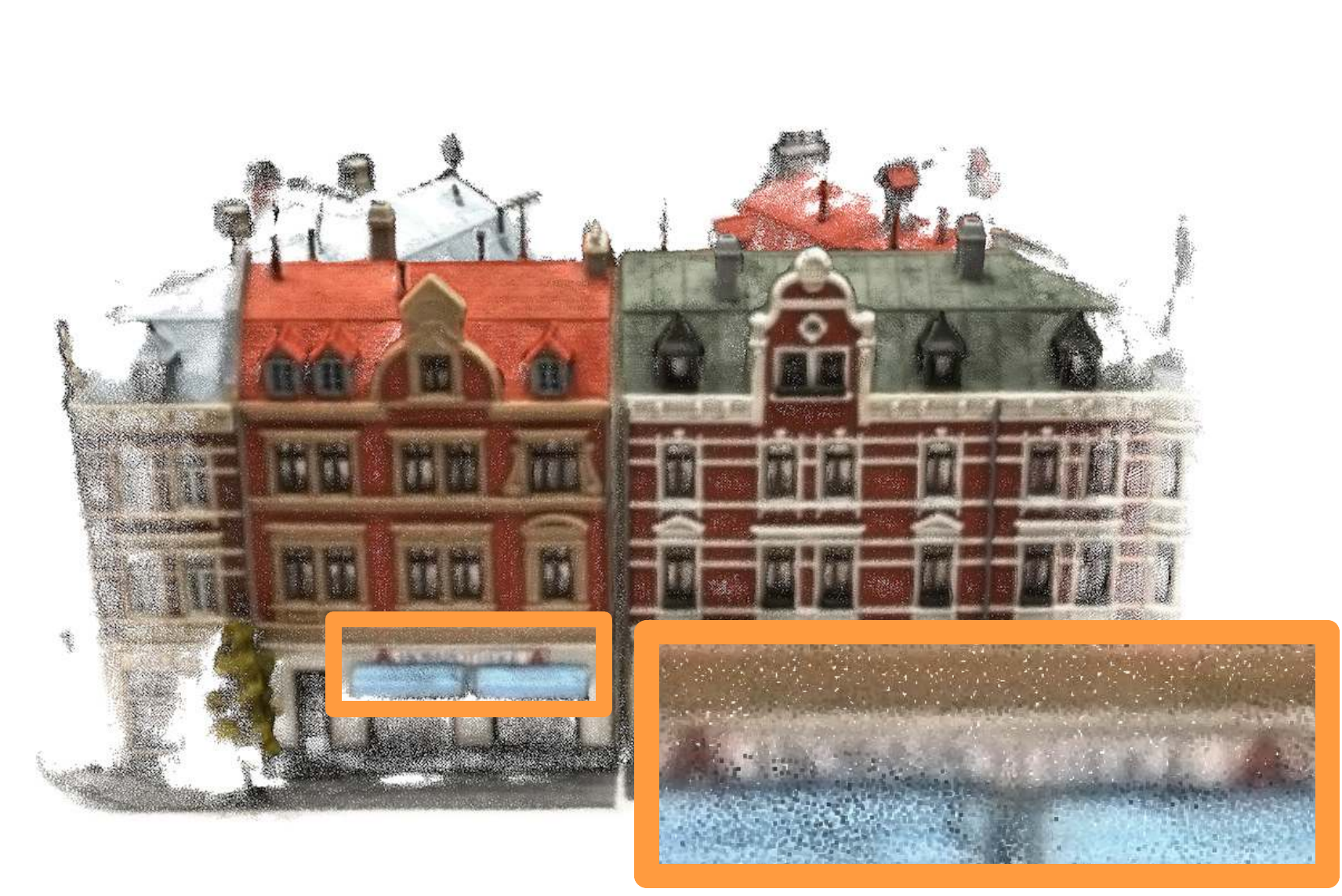}
      & \includegraphics[width=0.2\linewidth]{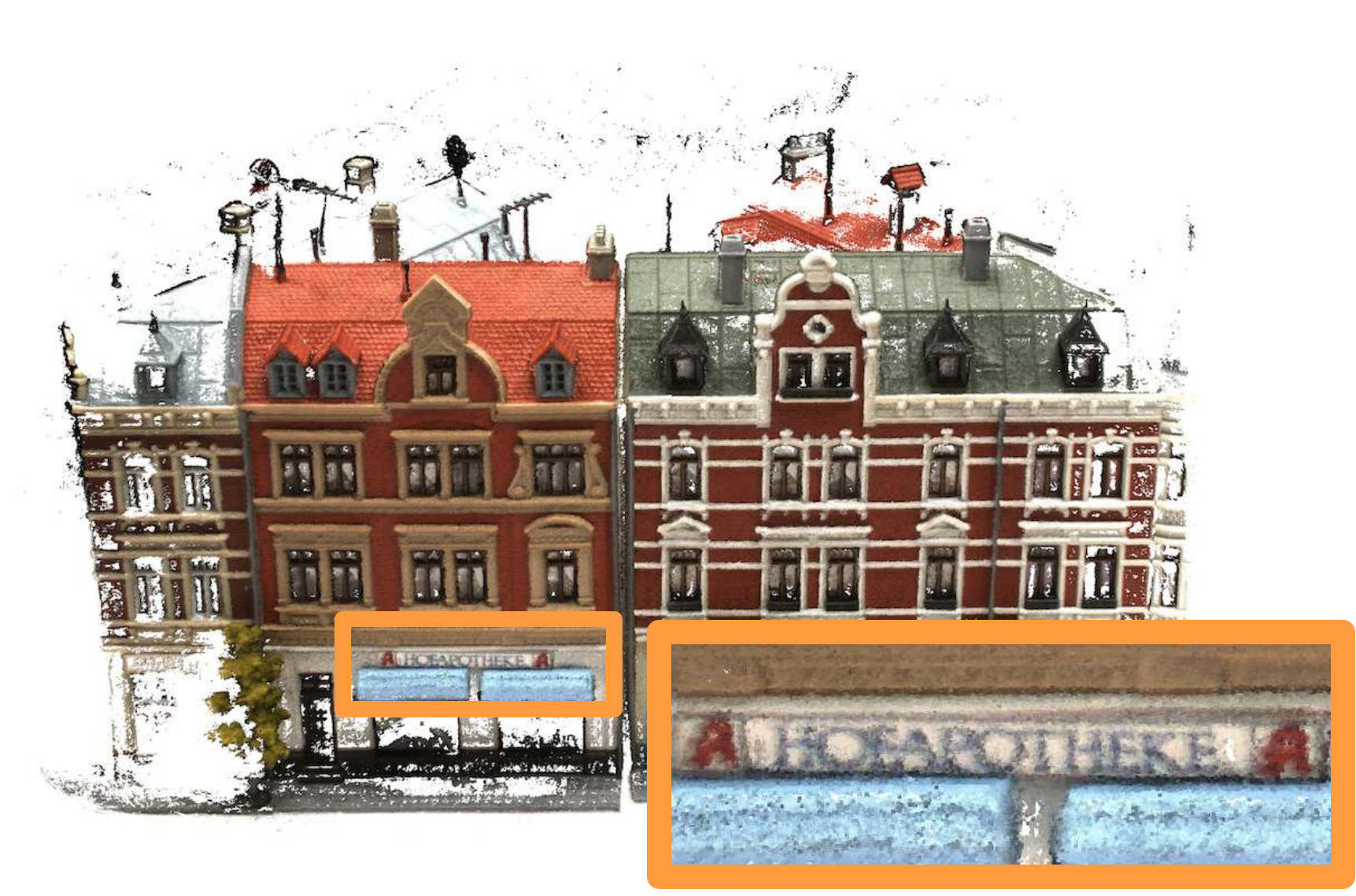}
      & \includegraphics[width=0.2\linewidth]{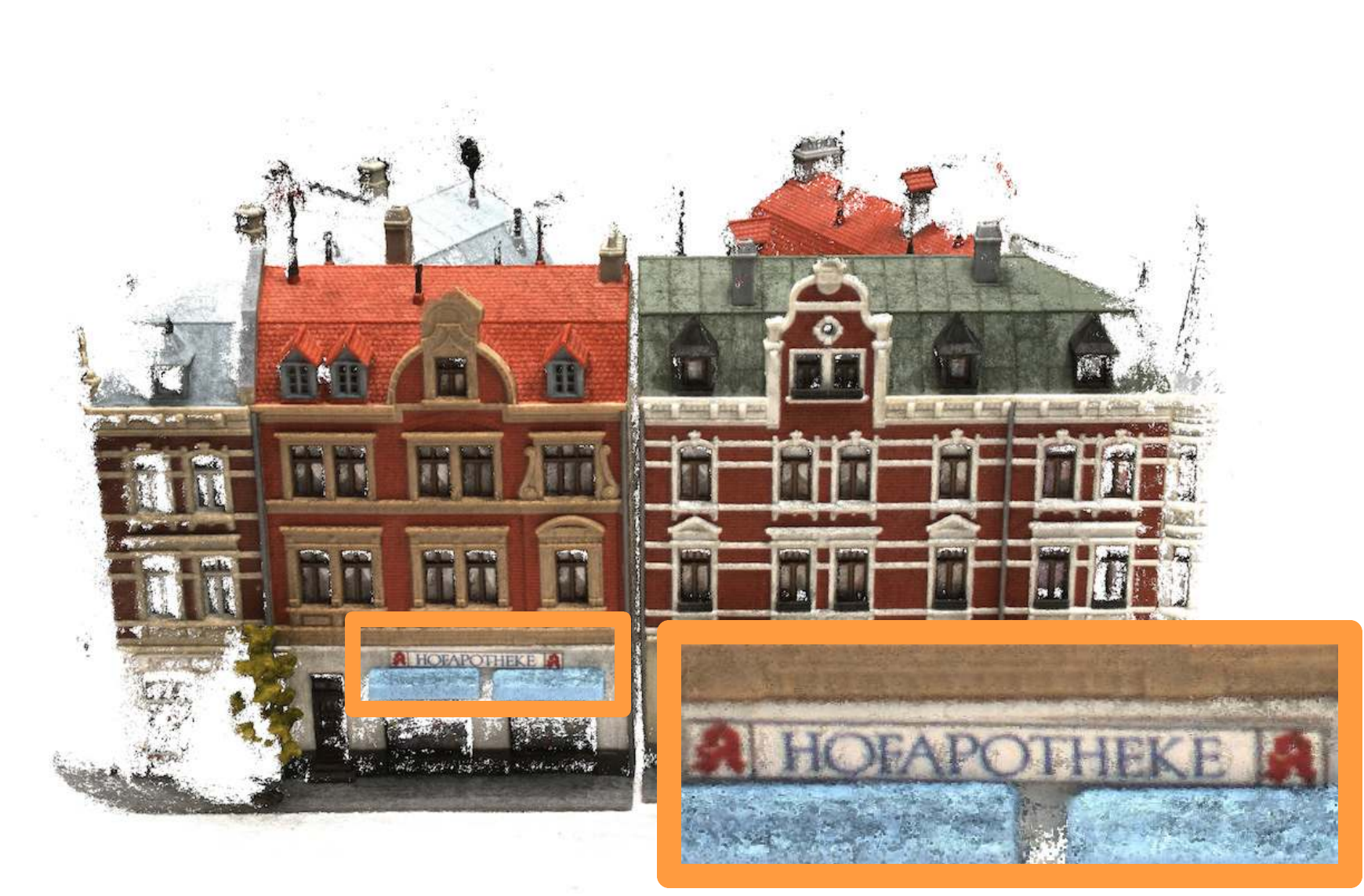}
      & \includegraphics[width=0.2\linewidth]{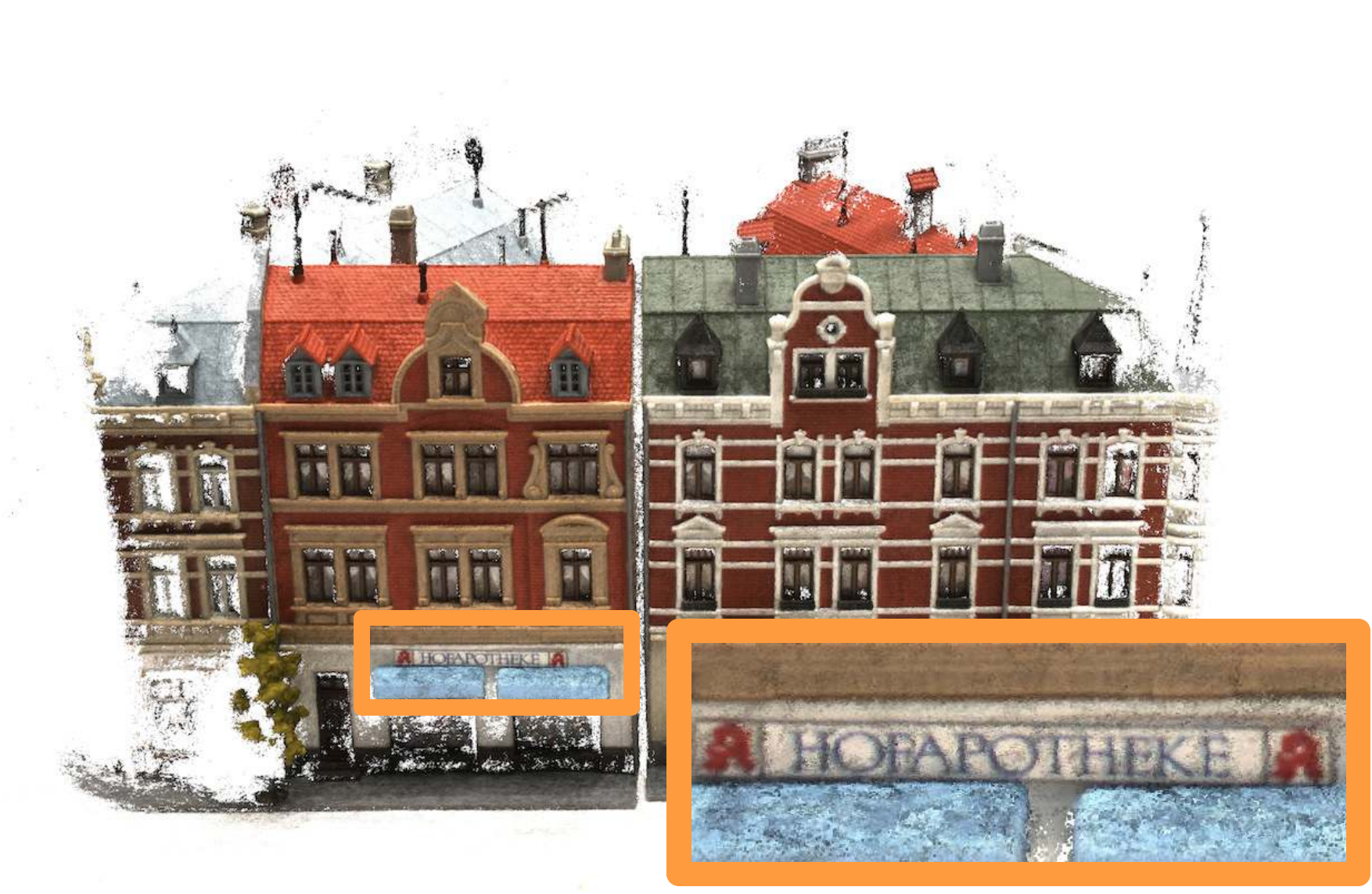}\\
      \includegraphics[width=0.13\linewidth]{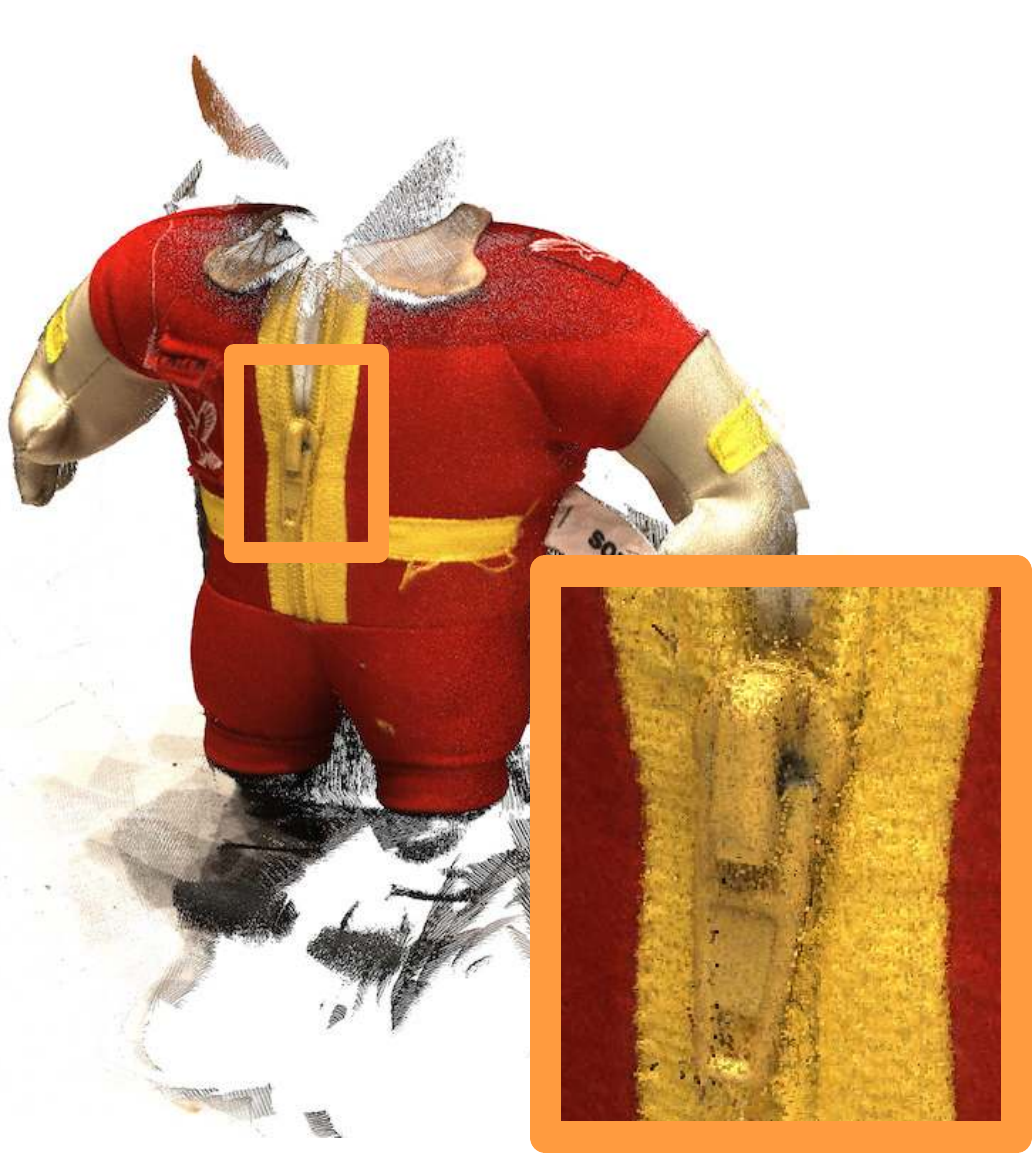}
      & \includegraphics[width=0.13\linewidth]{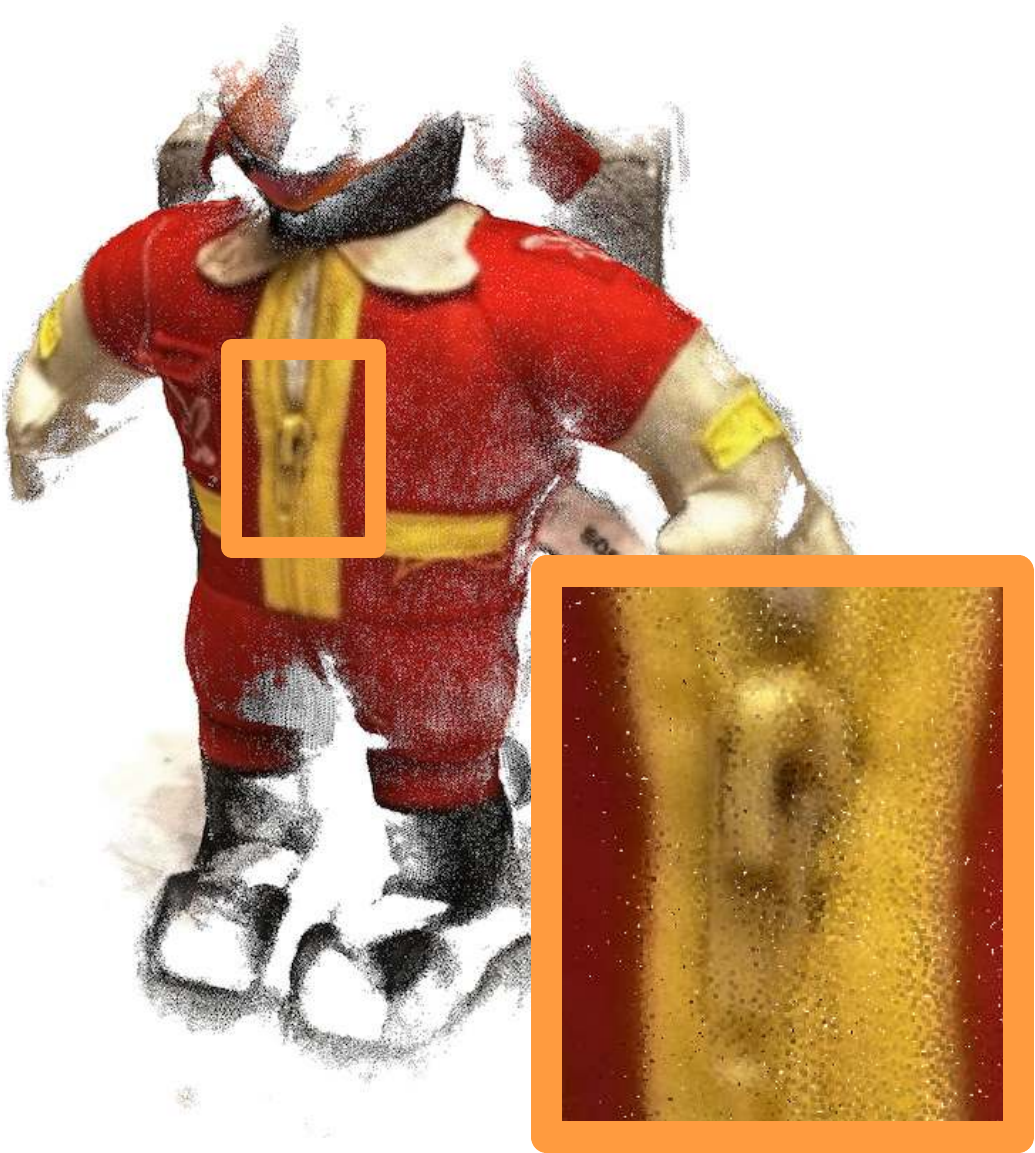}
      & \includegraphics[width=0.13\linewidth]{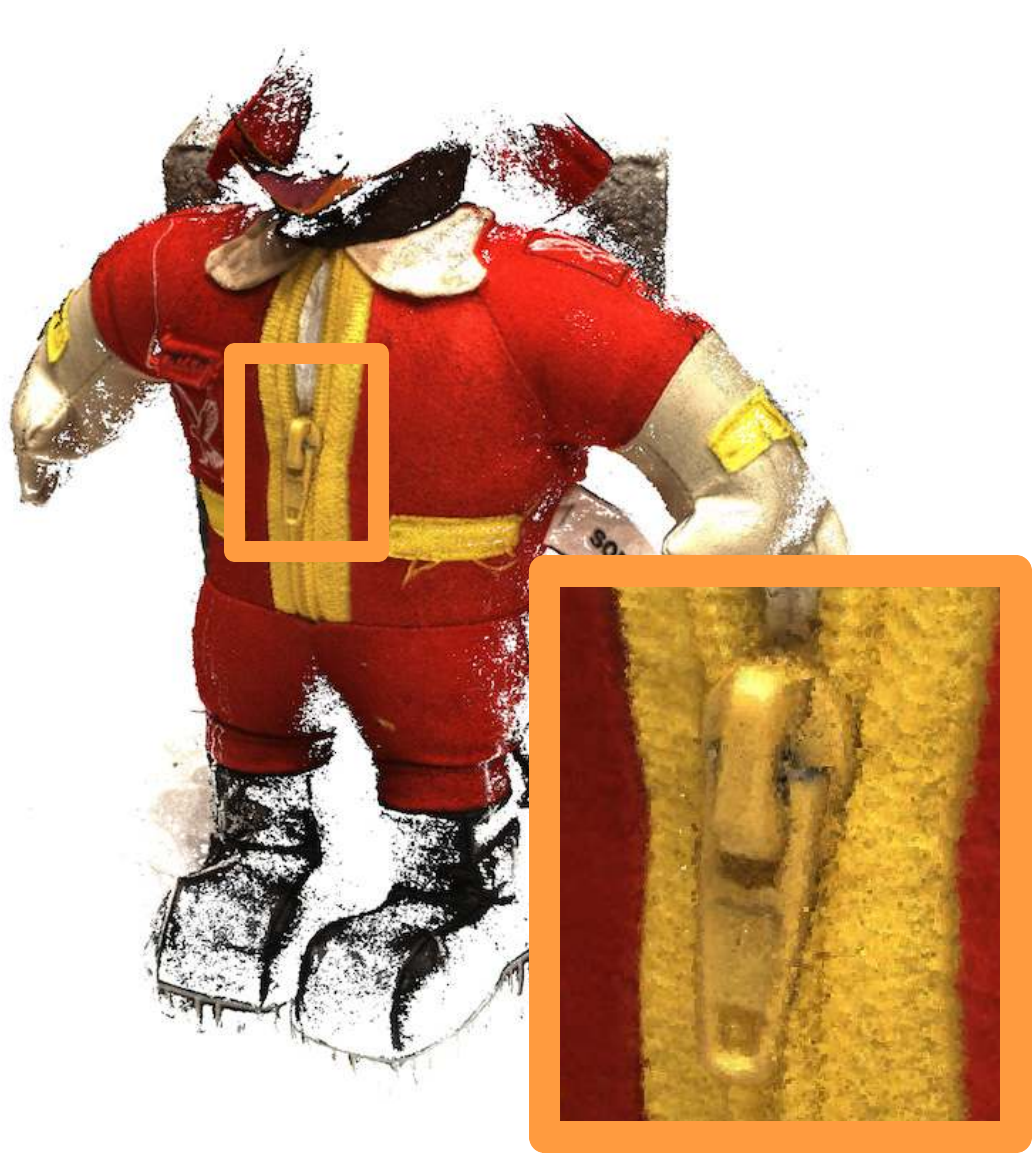}
      & \includegraphics[width=0.13\linewidth]{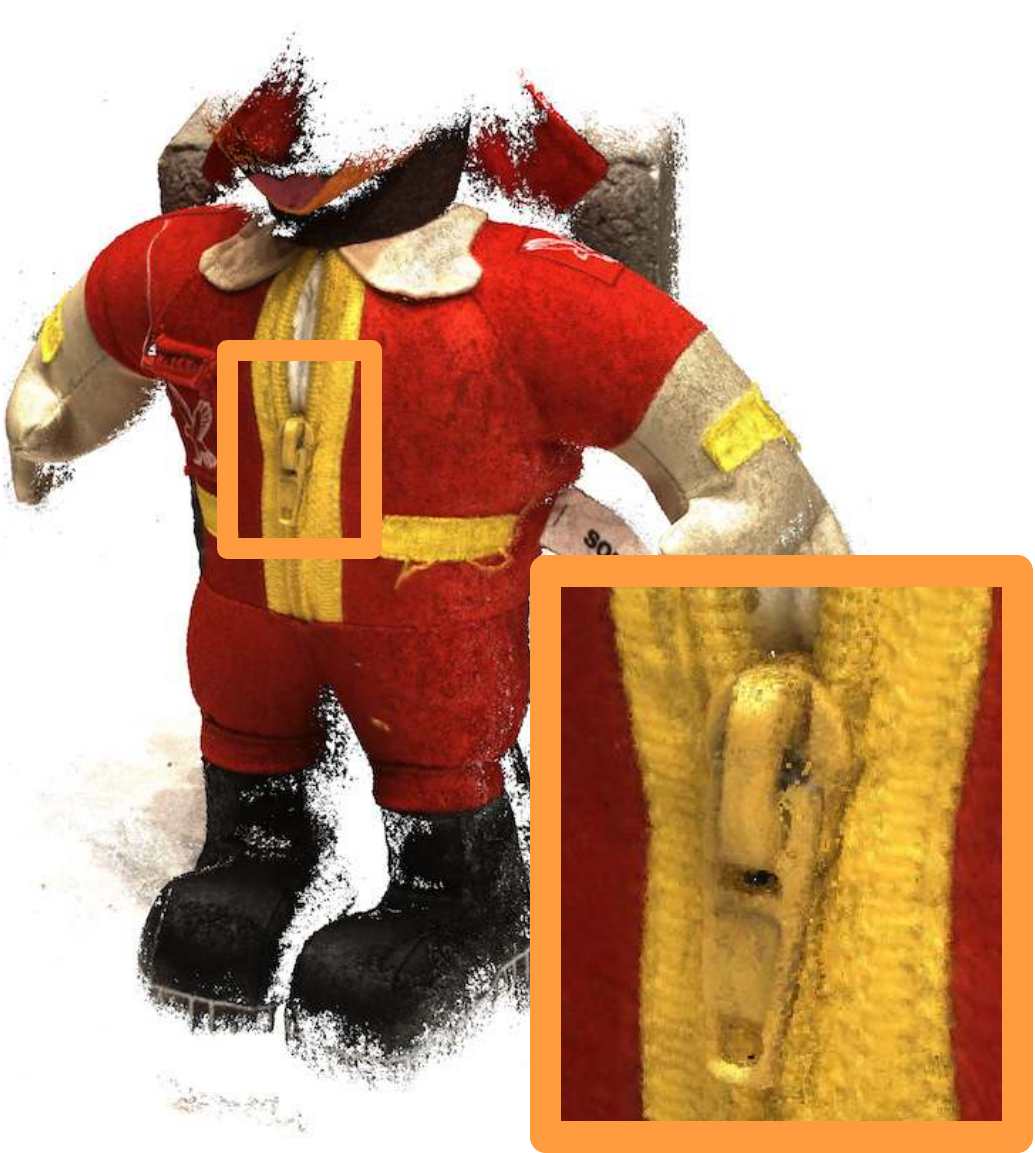}
      & \includegraphics[width=0.13\linewidth]{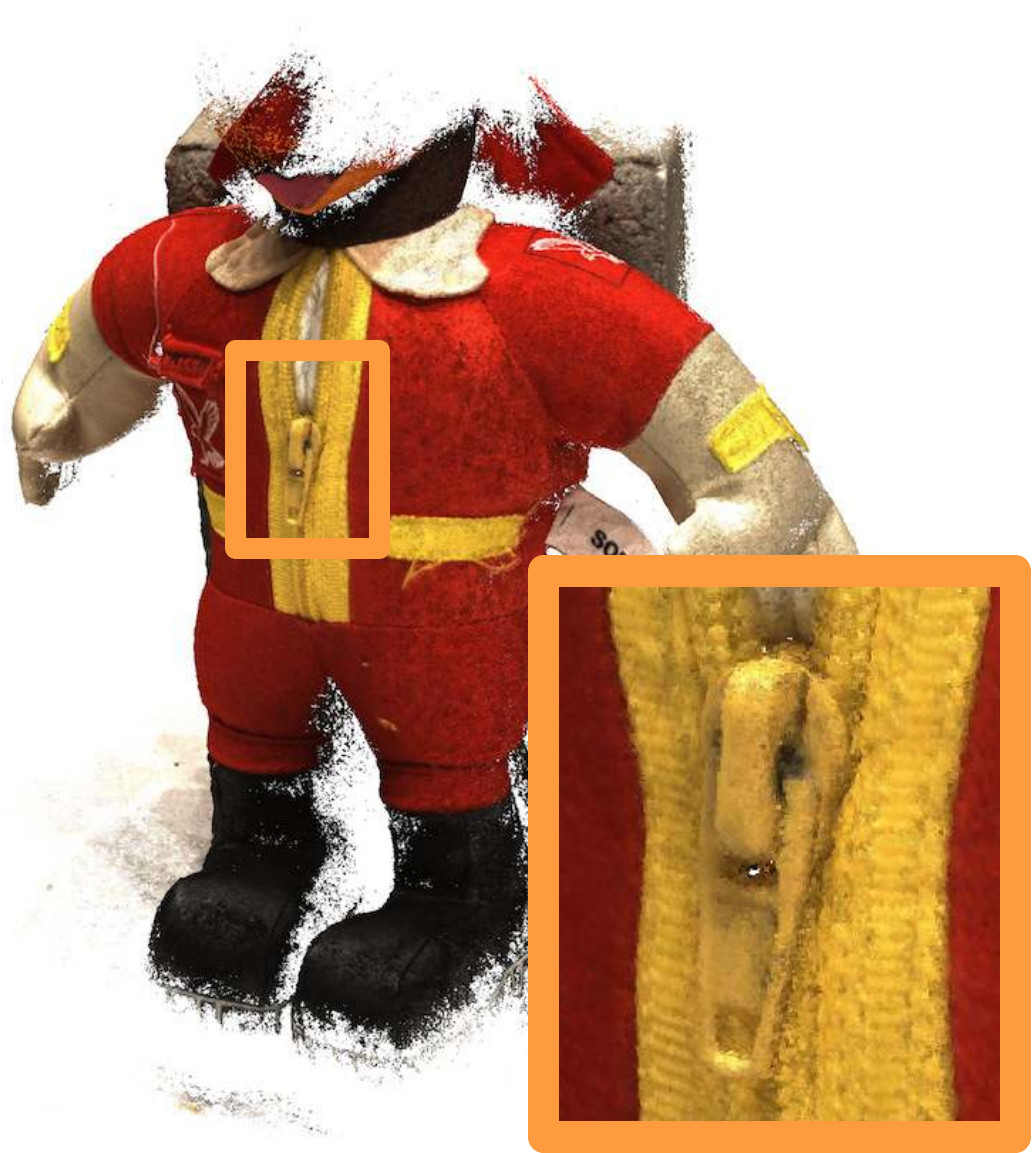}\\
      \includegraphics[width=0.16\linewidth]{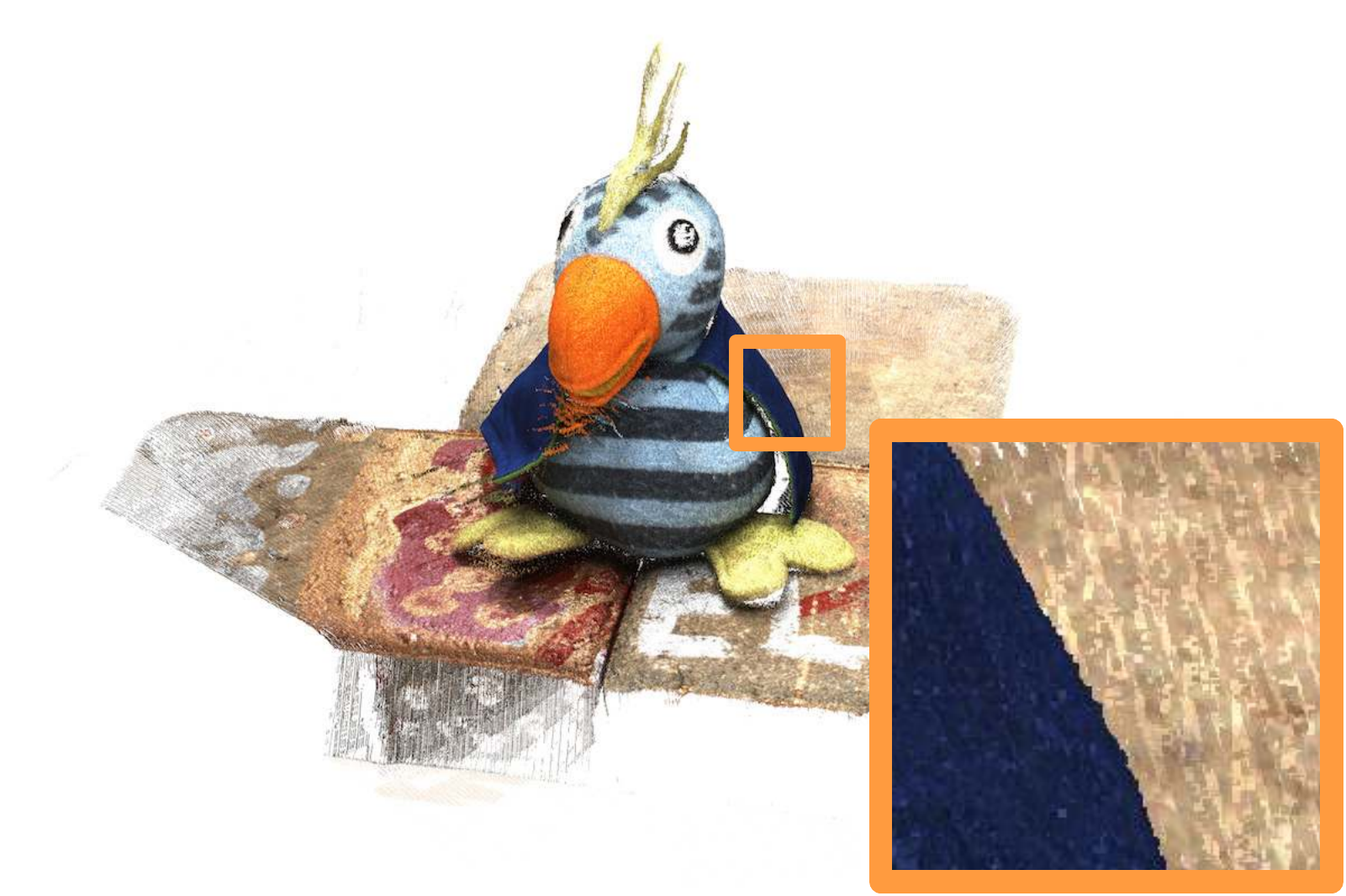}
      & \includegraphics[width=0.16\linewidth]{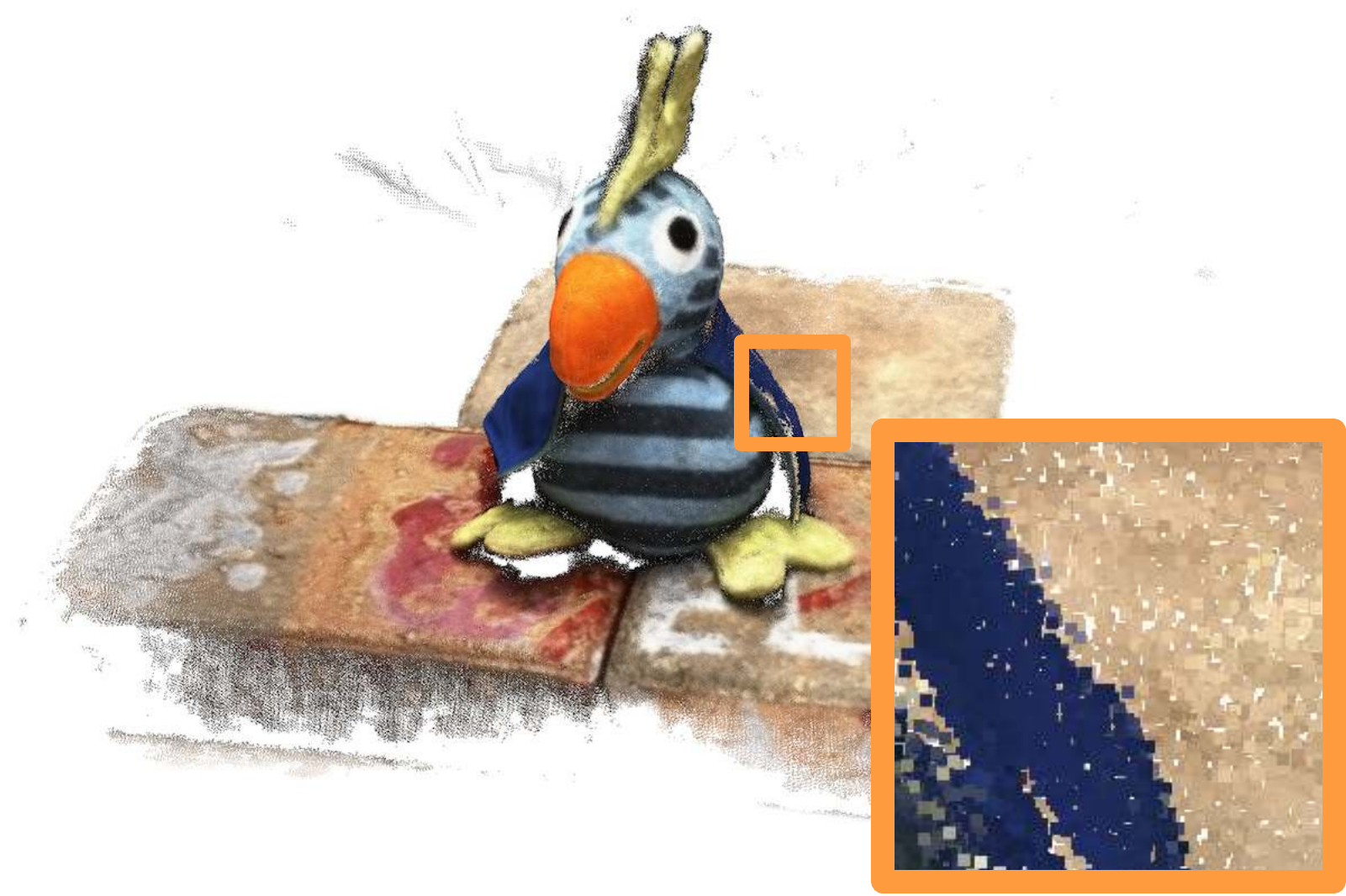}
      & \includegraphics[width=0.16\linewidth]{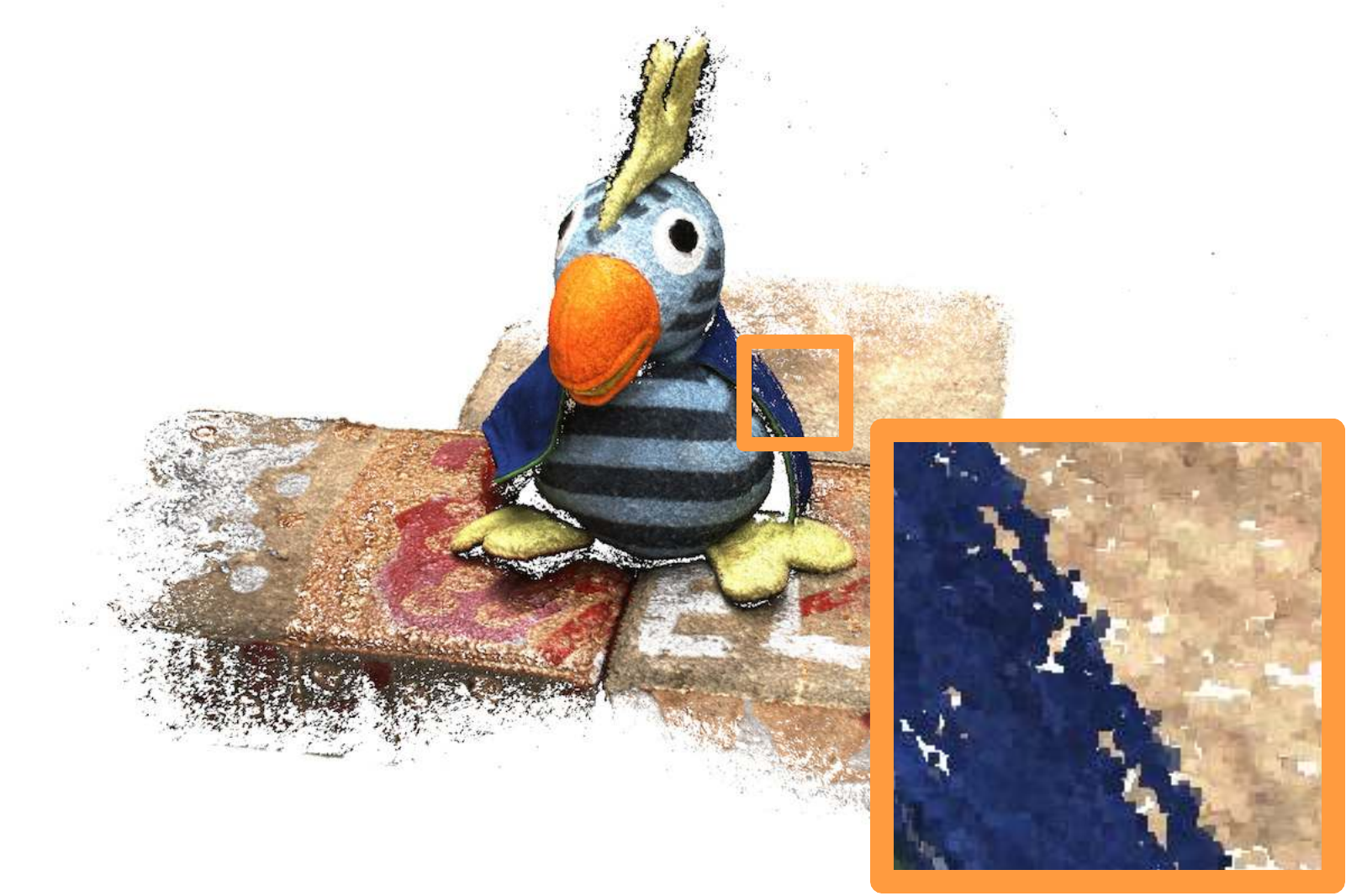}
      & \includegraphics[width=0.16\linewidth]{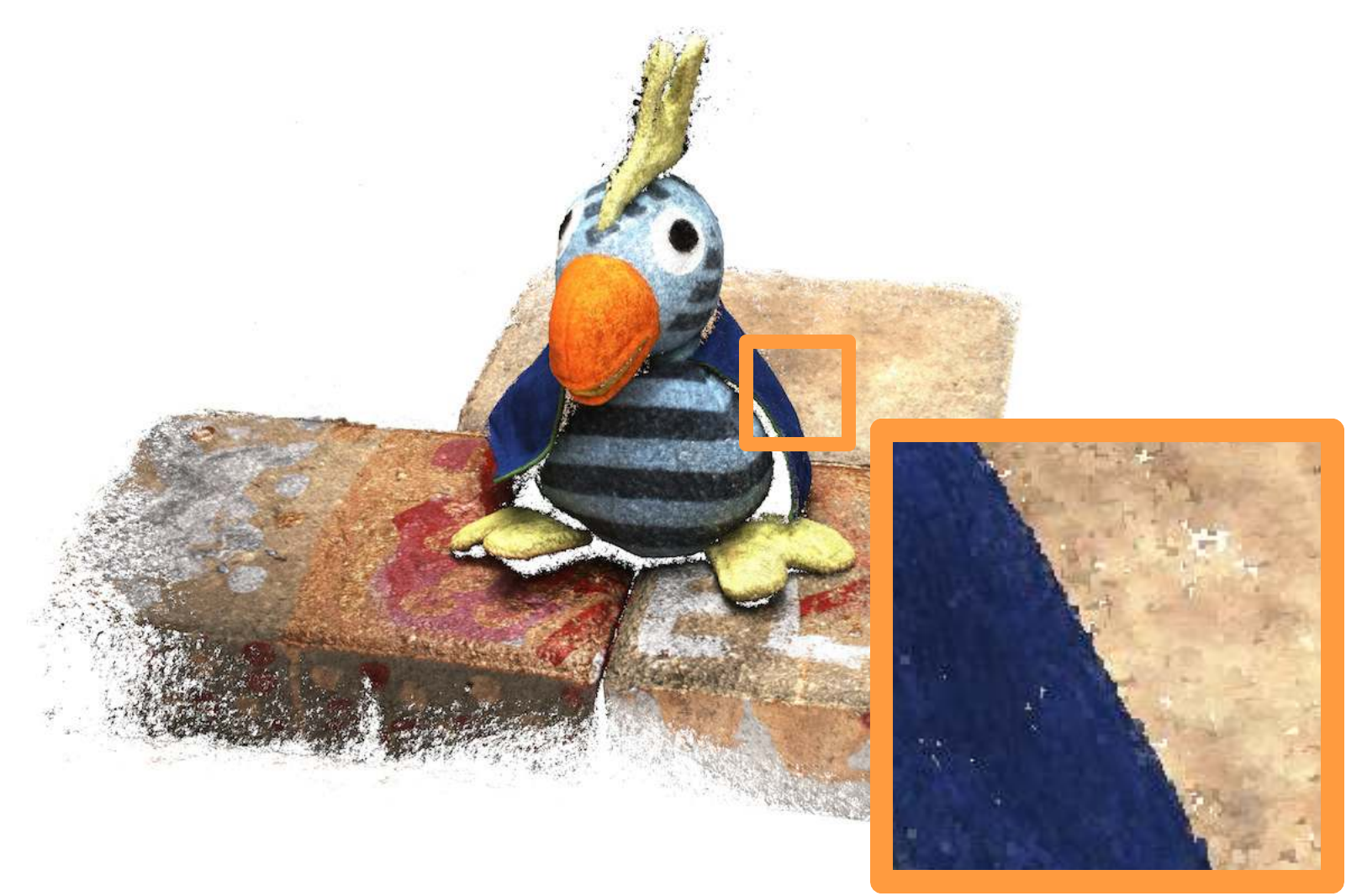}
      & \includegraphics[width=0.16\linewidth]{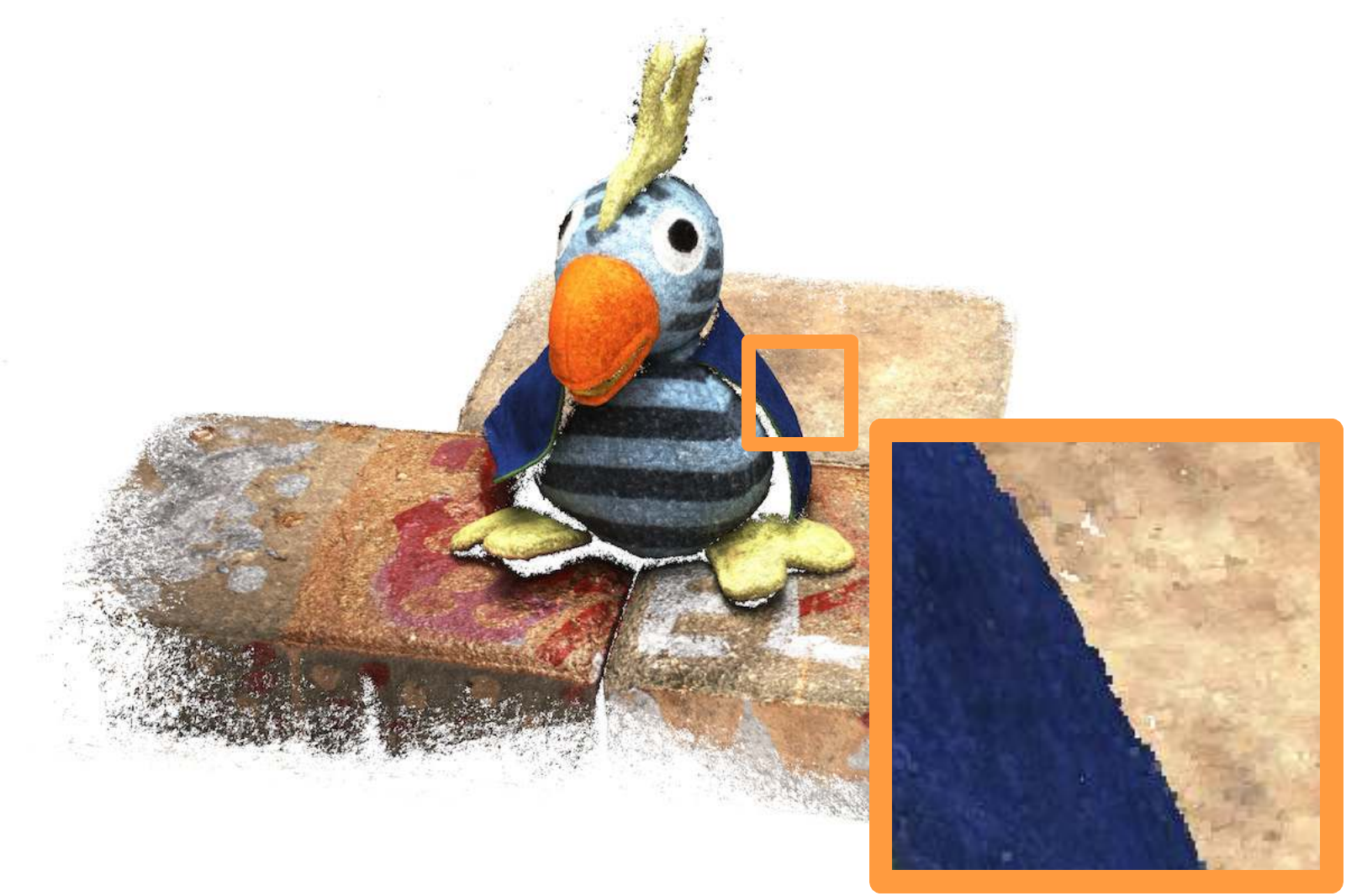}\\
      
      \multicolumn{1}{c}{Ground-truth} & \multicolumn{1}{c}{Khot \etal~\cite{Khot2019learning}} & \multicolumn{1}{c}{Ours} & \multicolumn{1}{c}{Ours} & \multicolumn{1}{c}{CVP-MVSNet~\cite{Yang2020CVP}}\\
      \multicolumn{1}{c}{Point Cloud} & \multicolumn{1}{c}{(Unsupervised)} & \multicolumn{1}{c}{(Unsupervised)} & \multicolumn{1}{c}{(Self-supervised)} & \multicolumn{1}{c}{(Supervised)}
\end{tabular}
    \end{center}
    \vspace{-0.5cm}
    \caption{\textbf{DTU Dataset.} Representative point cloud results. Best viewed on screen.}
    \label{fig:dtu}
    \vspace{0.1cm}
\end{figure*}

\begin{figure*}[!ht]
    \begin{center}
    \setlength\tabcolsep{3pt}
    \includegraphics[width=0.9\linewidth]{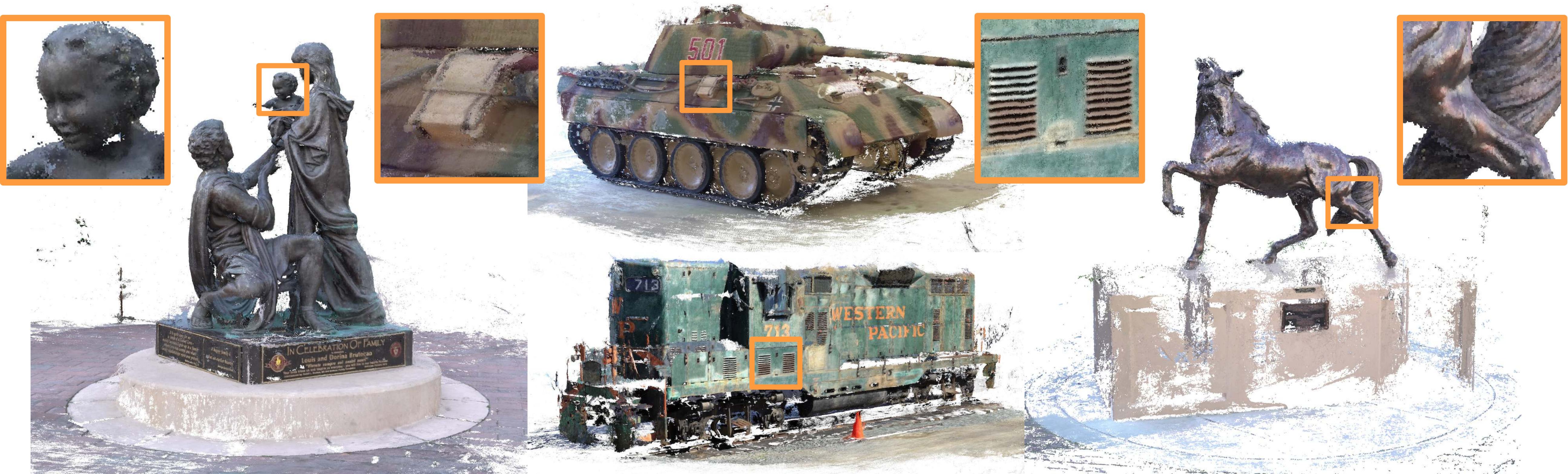}
    \end{center}
    \vspace{-0.5cm}
    \caption{\textbf{Tanks and Temples.} Representative point cloud results. Best viewed on screen.}
    \label{fig:tanks}
    \vspace{-0.3cm}
\end{figure*}

\begin{figure*}[!ht]
    \begin{center}
    \setlength\tabcolsep{3pt}
    \includegraphics[width=0.96\linewidth]{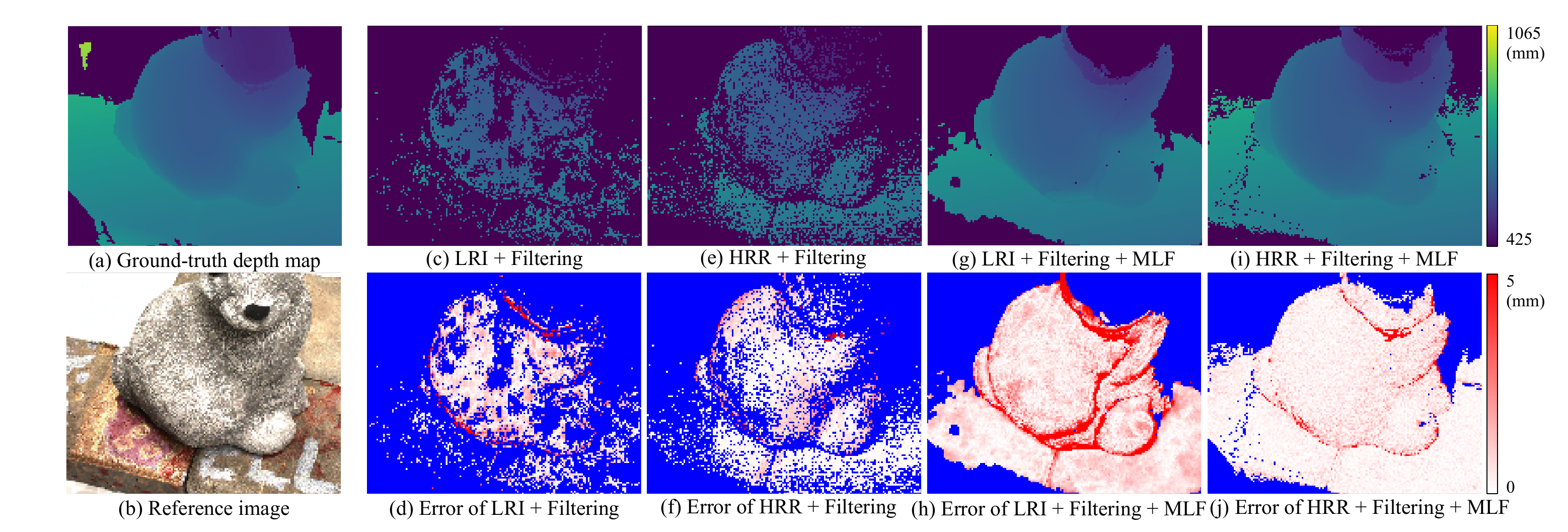}
    \end{center}
    \vspace{-0.5cm}
    \caption{Pseudo depth labels generated using different methods. (a) Ground-truth depth map. (b) Reference image. Top row, Columns 2-5: Pseudo depth label generated from different combination of following methods: Low Resolution Inference (LRI), High Resolution pseudo label Refinement (HRR), Pseudo label filtering (Filtering) and Multi-view pseudo Label Fusion (MLF). Bottom row is the error visualization of corresponding pseudo depth label. Areas with no pseudo depth labels are marked as blue in the error visualization. Best viewed on screen.}
    \label{fig:pseudo_depth_maps}
    \vspace{-0.4cm}
\end{figure*}

\subsection{Dataset}
\noindent\textbf{DTU Dataset}~\cite{aanaes2016large} is a large-scale MVS dataset with 124 scenes scanned from 49 or 64 views under 7 different lighting conditions. DTU provides 3D point clouds acquired using structured-light sensors.~Each view consists of an image and the calibrated camera parameters. We only use the provided images and camera parameters for the proposed self-supervised learning framework. For the unsupervised initialization, we downsample the images in the training set into $512\times 640$. For generating pseudo depth labels in the iterative self-training, we use the original $1600\times 1200$ training images. For iterative self-training, we downsample the images in the training set into $160\times 128$. We use the same training, validation and evaluation sets as defined in~\cite{yao2018mvsnet,yao2019recurrent}. We report the mean accuracy~\cite{aanaes2016large}, mean completeness\cite{aanaes2016large} and overall score~\cite{yao2018mvsnet}. For ablation experiments on this dataset, we also report $0.5mm$ \textit{f-score}. 

\noindent\textbf{Tanks and Temples}~\cite{knapitsch2017tanks} contains both indoor and outdoor scenes under realistic lighting conditions with large scale variations. We evaluate the generalization ability of our proposed self-supervised learning framework on the \emph{intermediate set}. We report the mean \textit{f-score} and the \textit{f-score} for each scene in that set.

\subsection{Self-supervised Learning}
To demonstrate the performance of the incremental self-supervised learning framework, we use our approach to train a CVP-MVSNet~\cite{Yang2020CVP} network on the DTU training dataset. No ground-truth depth training data is used.~\tabref{table:selfsup} shows the summary of performance of our results and existing unsupervised MVS networks. As shown, our method outperforms existing unsupervised MVS networks by a large margin.~We perform qualitative comparison with supervised and unsupervised methods in~\figref{fig:dtu} to further demonstrate the performance of our approach.

We also compare our self-supervised results obtained without any ground-truth training data to traditional geometric-based MVS frameworks and supervised MVS networks, including recent methods with learning based view aggregation ~\cite{yi2020pvamvsnet,chen2020vapointmvsnet,zhang2020vismvsnet,xu2020pvsnet} that outperform our backbone network\cite{Yang2020CVP} in supervised scenario. \tabref{table:selfsup_supervised} summarizes this comparison. As shown, our approach provides competitive results compared to traditional and supervised networks. \tabref{table:cvp_compare} shows a more detailed comparison between our approach and its supervised counter part~\cite{Yang2020CVP}. Overall, the supervised approach achieves a slightly better \textit{f-score} (+0.21\%) and slight reduction in reconstruction error ($0.012 mm$) at the expense, however, of needing ground-truth data. 

\vspace{-1mm}
\subsection{Generalization Ability}
To evaluate the generalization ability of the proposed self-supervised learning framework, we firstly train the CVP-MVSNet with self-supervised learning on DTU training dataset and directly test the model on Tanks and Temples dataset without any fine tuning. Further more, since our method does not rely on ground truth labels, we can apply the method on the training images of Tanks and Temples dataset. Results are listed in Tab.~\ref{table:tanks} and ~\figref{fig:tanks}. Our results clearly outperform existing unsupervised MVS networks. Fine-tuning on Tanks and Temples training data can further boost performance~(See first row of Tab. \ref{table:tanks}).

\begin{table}[!t]
\begin{center}
\footnotesize
\begin{tabular}{l|ccc}
\hline
Method & Acc.$\downarrow$ & Comp.$\downarrow$ & Overall$\downarrow$ (\textit{mm}) \\
\hline\hline
Khot \etal~\cite{Khot2019learning} & 0.881 & 1.073 & 0.977 \\
MVS$^2$~\cite{Dai2019mvs2} & 0.760 & 0.515 & 0.637 \\
M$^3$VSNet~\cite{huang2020m3vsnet} & 0.636 & 0.531 & 0.583 \\
Ours (self-sup.) & \textbf{0.308}	& \textbf{0.418} & \textbf{0.363} \\
\hline
\end{tabular}
\end{center}
\vspace{-0.3cm}
\caption{\textbf{DTU Dataset.} Quantitative reconstruction results of unsupervised and self-supervised MVS networks}
\label{table:selfsup}
\vspace{-0.4cm}
\end{table}

\begin{table}[!t]
\begin{center}
\footnotesize
\resizebox{\linewidth}{!}{
\begin{tabular}{l|cccccc}
\hline
Method & Acc.$\downarrow$ & Comp.$\downarrow$ & Overall$\downarrow$ & Precision$\uparrow$ & Recall$\uparrow$ & \textit{f-score}$\uparrow$ \\
\hline\hline
Supervised & \textbf{0.296}	& \textbf{0.406} & \textbf{0.351}  & 88.99\%	& \textbf{88.39\%}	& \textbf{88.63\%}\\
Ours (self-sup.) & 0.308	& 0.418 & 0.363 & \textbf{89.21\%}	& 87.80\%	& 88.42\%\\
\hline
\end{tabular}
}
\end{center}
\vspace{-0.6cm}
\caption{\textbf{DTU Dataset.} Quantitative reconstruction results of proposed self-supervised model compared to its supervised counterpart.
}
\label{table:cvp_compare}
\vspace{-0.2cm}
\end{table}

\subsection{Ablation Study}
Hereby we provide ablation studies analysis by evaluating the contribution of each part of our self-supervised approach to the final reconstruction quality. We also evaluate the ability of self-improving on reconstruction quality and discuss about the limitation of proposed method on texture-less areas. More ablation experiments and discussions can be found in supplementary material.

\noindent\textbf{Probability based image synthesis.}
We first analyze the effect of probability based image synthesis by comparing our approach to directly warp image base on the estimated depth. \tabref{table:prob_syn} summarizes the results for this experiment. As shown, there is a significant performance improvement when using probability based image synthesis for unsupervised learning.

\begin{table}[!t]
\begin{center}
\footnotesize
\begin{tabular}{ll|ccc}
\hline
\multicolumn{2}{c|}{Method} & Acc.$\downarrow$ & Comp.$\downarrow$ & Overall$\downarrow$ (\textit{mm}) \\
\hline\hline
\parbox[t]{2mm}{\multirow{5}{*}{\rotatebox[origin=c]{90}{Traditional}}}
& Furu~\cite{furu2010} & 0.613 & 0.941 & 0.777 \\
&Tola~\cite{tola2012} & 0.342 & 1.190 & 0.766 \\
&Camp~\cite{comp2008} & 0.835 & 0.554 & 0.695 \\
&Gipuma~\cite{galliani2016gipuma} & \textbf{0.283} & 0.873 & 0.578 \\
&Colmap~\cite{schoenberger2016sfm,schoenberger2016mvs} & 0.400 & 0.664 & 0.532 \\
\hline
\parbox[t]{2mm}{\multirow{10}{*}{\rotatebox[origin=c]{90}{Supervised}}}
&MVSNet~\cite{yao2018mvsnet} & 0.396 & 0.527 & 0.462 \\
&Point-MVSNet~\cite{chen2019point} & 0.342 & 0.411 & 0.376 \\
&CasMVSNet~\cite{gu2019cas} & 0.325 & 0.385 & 0.355 \\
&CVP-MVSNet~\cite{Yang2020CVP} & 0.296 & 0.406 & 0.351 \\
&UCSNet~\cite{cheng2020ucsnet} & 0.338 & 0.349 & 0.344 \\
&PVA-MVSNet~\cite{yi2020pvamvsnet} & 0.379 & 0.336 & 0.357 \\
&VA-Point-MVSNet~\cite{chen2020vapointmvsnet} & 0.359 & 0.358 & 0.359 \\
&Vis-MVSNet~\cite{zhang2020vismvsnet} & 0.369 & 0.361 & 0.365 \\
&PVSNet~\cite{xu2020pvsnet} & 0.337 & \textbf{0.315} & \textbf{0.326} \\
\hline
\parbox[t]{2mm}{\multirow{2}{*}{\rotatebox[origin=c]{90}{}}}
&Ours~(self-supervised) & 0.308	& 0.418 & 0.363 \\
\hline
\end{tabular}
\end{center}
\vspace{-0.6cm}
\caption{\textbf{DTU dataset.} Quantitative reconstruction results of traditional, supervised MVS networks, and our self-supervised approach.
}
\label{table:selfsup_supervised}
\vspace{-0.6cm}
\end{table}

\begin{table*}[!t]
\begin{center}
\footnotesize
\begin{tabular}{l|cccccccccc}
\hline
Method       & Rank$\downarrow$       & Mean$\uparrow$  & Family$\uparrow$ & Francis$\uparrow$ & Horse$\uparrow$ & Lighthouse$\uparrow$ & M60$\uparrow$   & Panther$\uparrow$ & Playground$\uparrow$ & Train$\uparrow$ \\
\hline\hline
Ours (Self-sup. T\&T) &  \textbf{42.00} & \textbf{56.54} & \textbf{76.35} & \textbf{49.06} & \textbf{43.04} & \textbf{57.35} & \textbf{60.64} & \textbf{57.35} & \textbf{58.47} & \textbf{50.06} \\
Ours (Self-sup. DTU) & 70.62 & 46.71 & 64.95 & 38.79 & 24.98 & 49.73 & 52.57 & 51.53 & 50.66 & 40.45 \\
M$^3$VSNet~\cite{huang2020m3vsnet} & 100.38 & 37.67 & 47.74 & 24.38 & 18.74 & 44.42 & 43.45 & 44.95 & 47.39 & 30.31 \\
MVS$^2$~\cite{Dai2019mvs2} & 100.38 & 37.21 & 47.74 & 21.55 & 19.50 & 44.54 & 44.86 & 46.32 & 43.48 & 29.72 \\

\hline
\end{tabular}
\end{center}
\vspace{-0.6cm}
\caption{\textbf{Tanks and Temples.} Performance as November 16, 2020. Our results clearly outperform existing unsupervised MVS networks.}
\label{table:tanks}
\vspace{-0.6cm}
\end{table*}

\begin{table}[!t]
\footnotesize
\begin{center}
\begin{tabular}{c|cccc}
\hline
Synthesis method & Acc.$\downarrow$ & Comp.$\downarrow$ & Overall$\downarrow$ & \textit{f-score}$\uparrow$ \\
\hline\hline
Depth Warping   & 0.447    & 0.773    & 0.610    & 75.29\%   \\
Probability Based   & {\bf 0.415}    & {\bf 0.720}    & {\bf 0.567}    & {\bf 77.06}\%   \\
\hline
\end{tabular} 
\end{center}
\vspace{-0.5cm}
\caption{\textbf{DTU dataset.} Effect of the probability based image synthesis.}
\vspace{-0.1cm}
\label{table:prob_syn}
\end{table}

\noindent\textbf{Each part of the self-supervised learning framework.}
In this experiment, we analyze the contribution of each part of the proposed self-supervised learning framework to the model performance. For the contribution of high resolution pseudo label inference, we compare the model performance to a model supervised by the filtered pseudo labels directly generated from low resolution training image. For the Multi-view pseudo label fusion, we compare the model performance to a model supervised by filtered depth pseudo labels without the multi-view pseudo label fusion. Results are listed in~\tabref{table:selfsup_contribution}. As shown, using the proposed pseudo label refinement and multi-view pseudo label fusion to generate pseudo labels and train a model yields better reconstruction results. In~\figref{fig:pseudo_depth_maps}, we also show pseudo depth labels generated by each of the methods. As shown, pseudo labels generated with proposed approach results in the lowest error and the best completeness.

\begin{table}[!t]
\footnotesize
\begin{center}
\begin{tabular}{c|cccc}
\hline
Model & Acc.$\downarrow$ & Comp.$\downarrow$ & Overall$\downarrow$ & \textit{f-score}$\uparrow$ \\
\hline\hline
Unsupervised   & 0.415    & 0.720    & 0.567    & 77.06\%   \\
LRI + Filtering   & 0.322    & 0.434    & 0.378    & 87.75\%   \\
HRR + Filtering & 0.316    & 0.428    & 0.372    & 88.01\%   \\
LRI + MLF + Filtering & 0.325    & 0.429    & 0.376    & 87.91\%   \\
HRR + MLF + Filtering & \textbf{0.308}    & \textbf{0.418}    & \textbf{0.363}    & \textbf{88.42\%}   \\
\hline
\end{tabular} 
\end{center}
\vspace{-0.5cm}
\caption{\textbf{DTU dataset.} Contribution of each part of the proposed incremental self-supervision framework on final reconstruction quality. Filtering: Pseudo label filtering. LRI: Low Resolution Inference. HRR: High Resolution pseudo label Refinement. MLF: Multi-view pseudo Label Fusion}
\label{table:selfsup_contribution}
\vspace{-0.5cm}
\end{table}

\noindent\textbf{Iteration of incremental self-training.}
We now analyze the performance of the model at each self-training iteration. \tabref{table:selfsup_itr} summarizes the results for this experiment. As shown, there is an initial increment in the performance during the first two iterations to yield a $0.5mm$-\textit{f-score} only 0.12\% lower than the supervised counterpart. After that iteration, the performance becomes stable after the 3rd iteration. These results suggest our self-supervised approach does not lead to potential performance drops if applied continuously.

\begin{table}[!t]
\footnotesize
\begin{center}
\begin{tabular}{c|cccc}
\hline
Self-supervision Itr. & Acc.$\downarrow$ & Comp.$\downarrow$ & Overall$\downarrow$ & \textit{f-score}$\uparrow$ \\
\hline\hline
init   & 0.415    & 0.720    & 0.567    & 77.06\%   \\
1      & 0.306    & 0.431    & 0.368    & 88.16\%   \\
2      & 0.308    & 0.418    & 0.363    & 88.42\%   \\
3      & 0.309    & 0.420    & 0.364    & 88.49\%   \\
4      & 0.309    & 0.421    & 0.365    & 88.47\%   \\
\hline
supervised & \textbf{0.296} & \textbf{0.406} & \textbf{0.351} & \textbf{88.61\%} \\
\hline
\end{tabular} 
\end{center}
\vspace{-0.5cm}
\caption{\textbf{DTU dataset.} Performance at different self-supervised iterations.}
\label{table:selfsup_itr}
\vspace{-0.4cm}
\end{table}

\begin{figure}[!t]
    \begin{center}
    \setlength\tabcolsep{1pt}
    \begin{tabular}{ccc}
      \includegraphics[width=0.32\linewidth]{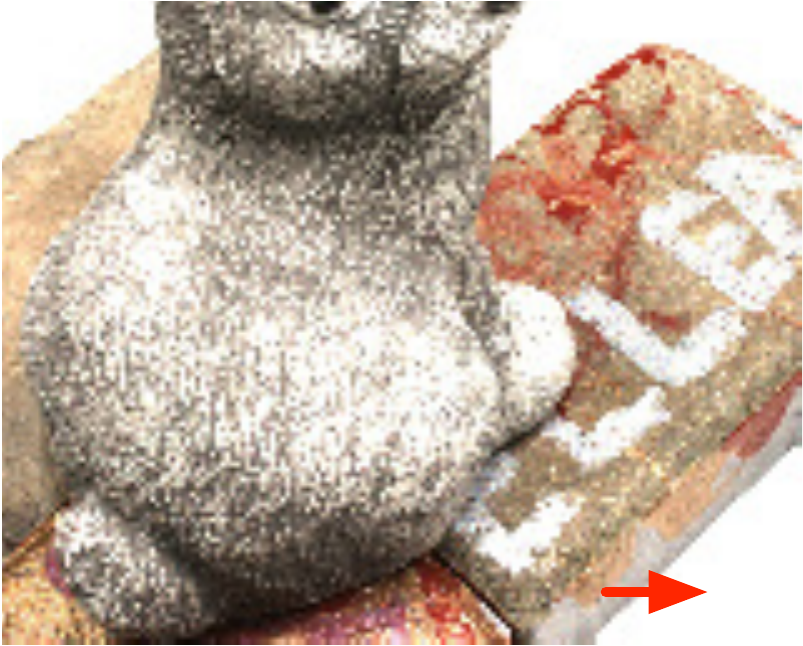}
      & \includegraphics[width=0.32\linewidth]{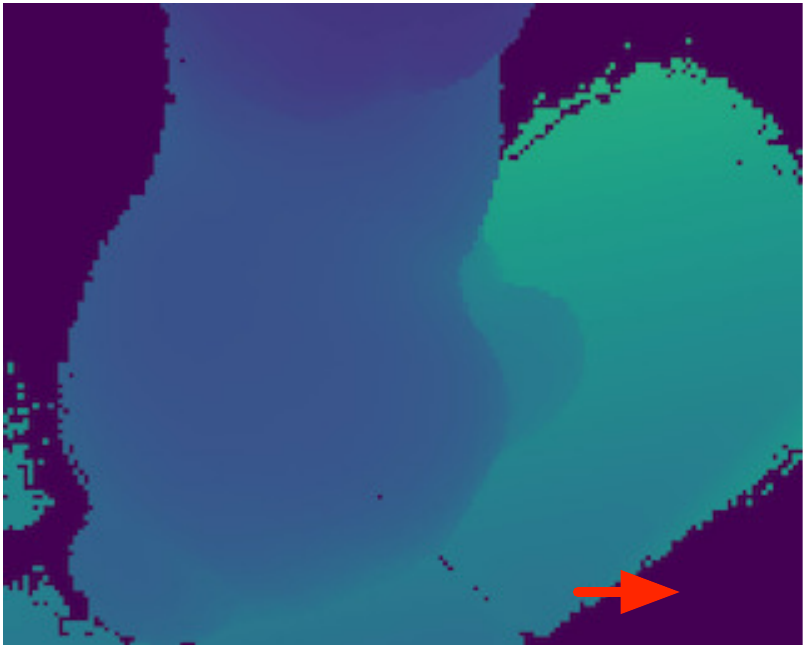}
      & \includegraphics[width=0.32\linewidth]{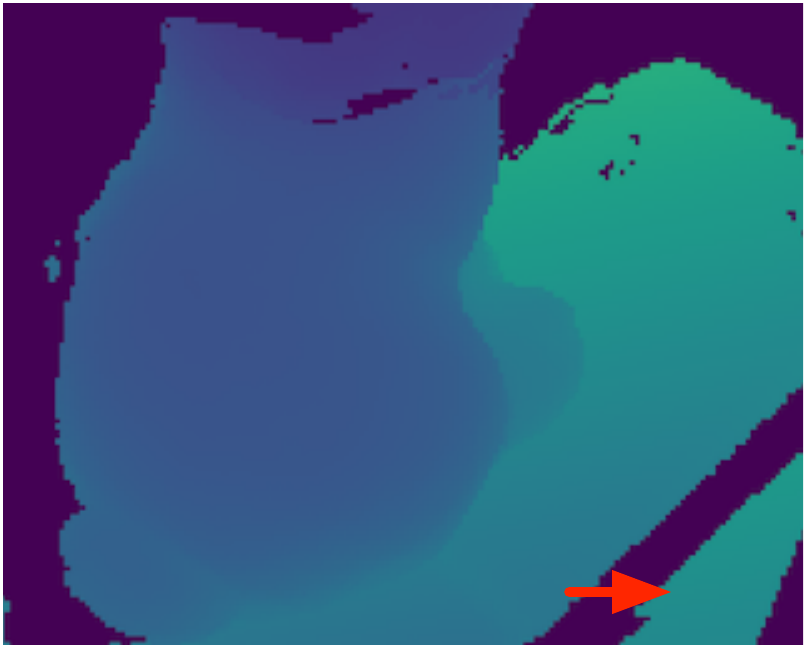}\\
      Image & Pseudo depth & Ground-truth
\end{tabular}
    \end{center}
    \vspace{-0.7cm}
    \caption{Generating pseudo labels for severely texture-less areas is the current main limitation of our method. Best viewed on screen.}
    \label{fig:texture_less}
    \vspace{-0.5cm}
\end{figure}

\noindent\textbf{Texture-less areas.} Despite the good performance achieved by our self-supervised learning method, one limitation appears on texture-less areas. As shown in Fig.~\ref{fig:texture_less}, pseudo depth labels generated by our approach do not contain any label on severe texture-less regions. This is mainly caused by the initial pseudo label generation. Recall that the initial pseudo labels are generated from an unsupervised learning framework based on photometric consistency across views. Thus, the algorithm can not find matches in texture-less areas and fails to provide the initial labels. Subsequent refinement processes will not be able to complete those regions. One possible solution would be to enforce long range smoothness to propagate depth from texture rich areas to texture-less ones during the initial pseudo label generation and label refinement process.

\section{Conclusion}

We proposed a self-supervised learning framework for depth inference from multiple view images. Given initial pseudo depth labels generated from a network under unsupervised learning process, we refine the flawed initial pseudo depth labels using a carefully designed pipeline. Using these refined pseudo depth labels as supervision signal, we achieve significantly better performance than state-of-the-art unsupervised networks and achieve similar performance compared to supervised learning frameworks. One current limitation of our approach is handling texture-less areas as the unsupervised stage fails to extract information to generate the initial pseudo labels. We will address this in our future work.

\section*{Acknowledgments}
This research is supported by Australian Research Council grants (DE180100628, DP200102274).

{\small
\bibliographystyle{ieee_fullname}
\bibliography{egbib}
}

\newpage
\appendix
\noindent{\LARGE{\textbf{Supplementary Material}}}
\vspace{0.4cm}

In this supplementary material, we first provide details of the image synthesis loss functions in section \ref{sec:syn_loss}. In section~\ref{sec:dtu_results}, we show additional qualitative reconstruction results by our method for different scans on the DTU dataset. In section~\ref{sec:pseudo_many}, we provide visualization of the pseudo depth labels generated by our self-supervised learning framework. In section~\ref{sec:pseudo_itr}, we show the pseudo depth labels generated by each iteration of our self-supervised learning framework. In section~\ref{sec:ablations}, we provide more ablation experiments. In section~\ref{sec:discussion}, we provide more discussions regarding the proposed method.

\section{Image synthesis loss}\label{sec:syn_loss}
In our approach, as introduced in section 3.2, we use a weighted combination of four loss functions, 
\begin{equation}
    l_{syn} = \alpha_1 l_{g} + \alpha_2 l_{ssim} + \alpha_3 l_{p} + \alpha_4 l_{s},
\end{equation}
\noindent where $l_{g}$ is the image gradient loss, $l_{ssim}$ is the structure similarity loss, $l_{p}$ is the perceptual loss, $l_{s}$ is the depth smoothness loss, and $\alpha_i$ sets the influence of each loss. 

\noindent{\bf Image gradient loss} is defined as the L1 distance between the gradient of input reference image $\nabla \textbf{I}_{0}^{l}({\bf x})$ and the synthesized image $\nabla \textbf{I}_{i\rightarrow 0}^{l}({\bf x})$ for each source view $i$ and each pyramid level $l$,
\begin{equation}
    l_g = \sum_{l=0}^{L}\frac{1}{N}\sum_{i=1}^{N}\sum_{{\bf x}\in \Omega}||\nabla \textbf{I}_{i\rightarrow 0}^{l}({\bf x}) - \nabla \textbf{I}_{0}^{l}({\bf x})||_1.
\end{equation}
where $\Omega$ is the set of valid pixels of the synthesized image.

\noindent{\bf Structure similarity loss} enforces the contextual similarity between a synthesized image and the input reference image. Specifically, we use the Structure Similarity Index~\cite{Wang2004ssim} to measure the contextual similarity. This index increases as the structure similarity between the images increases, with a range $[-1,1]$. We formulate the loss as the negative of $SSIM$ between each synthesized image and input reference image, 
\begin{equation}
    l_{ssim} = \sum_{l=0}^{L}\frac{1}{N}\sum_{i=1}^{N} 1-SSIM(\textbf{I}_{i\rightarrow 0}^{l},\textbf{I}_{0}^{l}).
\end{equation}

\noindent{\bf Perceptual loss} also encourages high-level contextual similarity between images~\cite{johnson2016perceptual}. This loss is defined as the L1 distance in the feature space of a shared weight perceptual network taking each image as input~\cite{johnson2016perceptual}. In our experiments, we use a VGG model~\cite{simonyan15vgg} and extract features from $3_{th}$,$8_{th}$,$15_{th}$ and $22_{th}$ layers. Therefore, we formulate the loss as follows,
\begin{equation}
    l_{p} = \sum_{l=0}^{L}\frac{1}{N}\sum_{i=1}^{N}\sum_{j\in[3,8,15,22]}||VGG(\textbf{I}_{i\rightarrow 0}^l,j)-VGG(\textbf{I}^l_{0},j)||_1.
\end{equation}

\noindent{\bf Depth smoothness loss} encourages local depth smoothness. This term encourages depth smoothness with respect to the alignment of image and depth discontinuities, which is measured by the gradient of color intensity of input reference image. We define this loss as follows,
\begin{equation}
    l_{sm} = \sum_{l=0}^{L}\sum_{{\bf x}\in \Omega}|\nabla_u \tilde D^l({\bf x})|e^{-|\nabla_u\textbf{I}_{0}^{l}({\bf x})|}+|\nabla_v \tilde D^l({\bf x})|e^{-|\nabla_v\textbf{I}_{0}^{l}({\bf x})|}
\end{equation}
\noindent where $\nabla_u$ and $\nabla_v$ refer to the gradient on $x$ and $y$ direction, and $\tilde D = D/\bar D$ is the mean-normalized inverse depth~\cite{wang2018cvpr}. 

\section{Qualitative results on DTU dataset}\label{sec:dtu_results}
\begin{figure*}[htbp]
    \vspace{1cm}
    \begin{center}
    \setlength\tabcolsep{0pt}
    \begin{tabular}{rrrrr}
    
      \includegraphics[width=0.2\linewidth]{Figs/1_1.pdf}
      & \includegraphics[width=0.2\linewidth]{Figs/1_2.pdf}
      & \includegraphics[width=0.2\linewidth]{Figs/1_3.pdf}
      & \includegraphics[width=0.2\linewidth]{Figs/1_4.pdf}
      & \includegraphics[width=0.2\linewidth]{Figs/1_5.pdf}\\
      \includegraphics[width=0.16\linewidth]{Figs/21.pdf}
      & \includegraphics[width=0.16\linewidth]{Figs/22.pdf}
      & \includegraphics[width=0.16\linewidth]{Figs/23.pdf}
      & \includegraphics[width=0.16\linewidth]{Figs/24.pdf}
      & \includegraphics[width=0.16\linewidth]{Figs/25.pdf}\\
      \includegraphics[width=0.15\linewidth]{Figs/31.pdf}
      & \includegraphics[width=0.15\linewidth]{Figs/32.pdf}
      & \includegraphics[width=0.15\linewidth]{Figs/33.pdf}
      & \includegraphics[width=0.15\linewidth]{Figs/34.pdf}
      & \includegraphics[width=0.15\linewidth]{Figs/35.pdf}\\
      \includegraphics[width=0.2\linewidth]{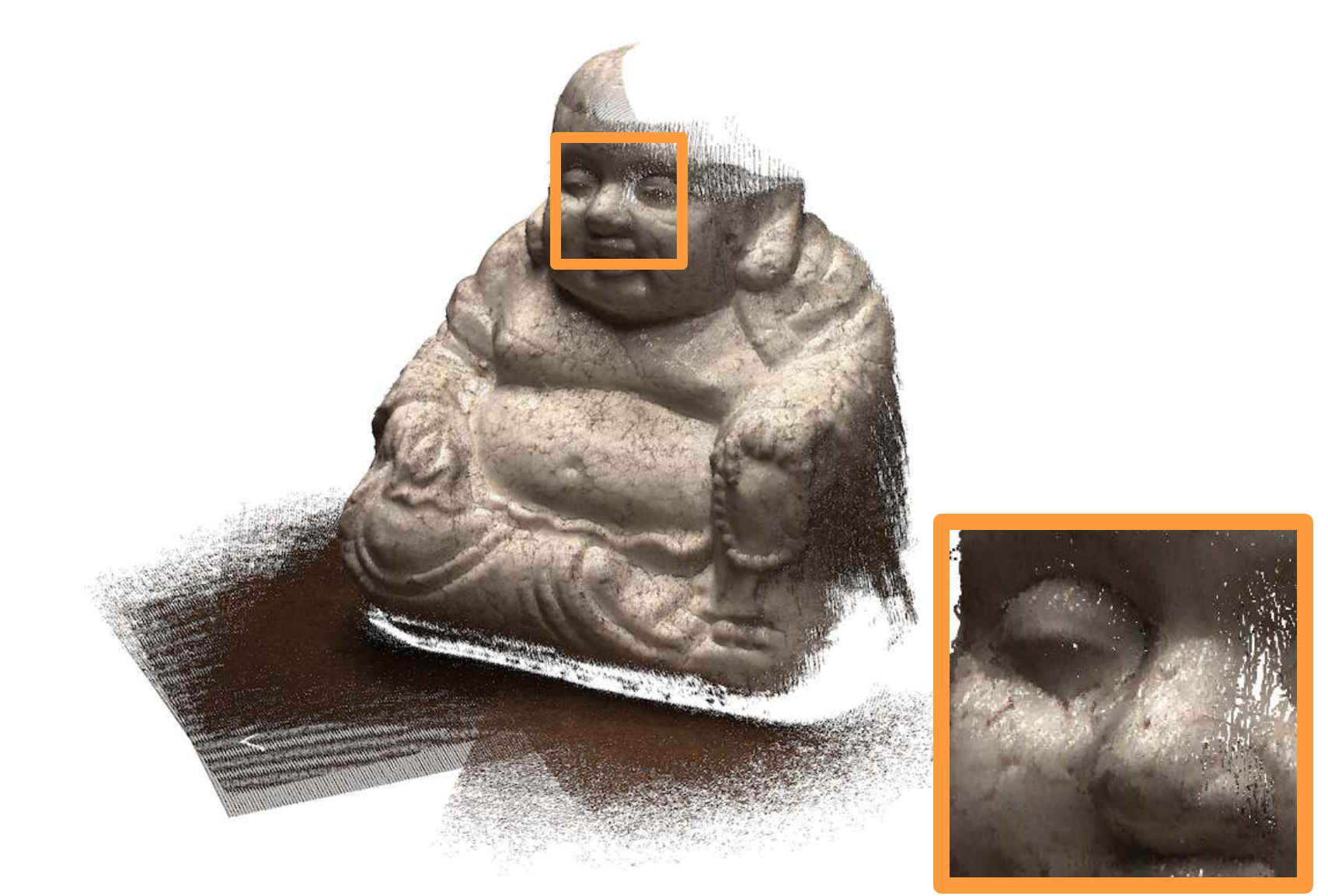}
      & \includegraphics[width=0.2\linewidth]{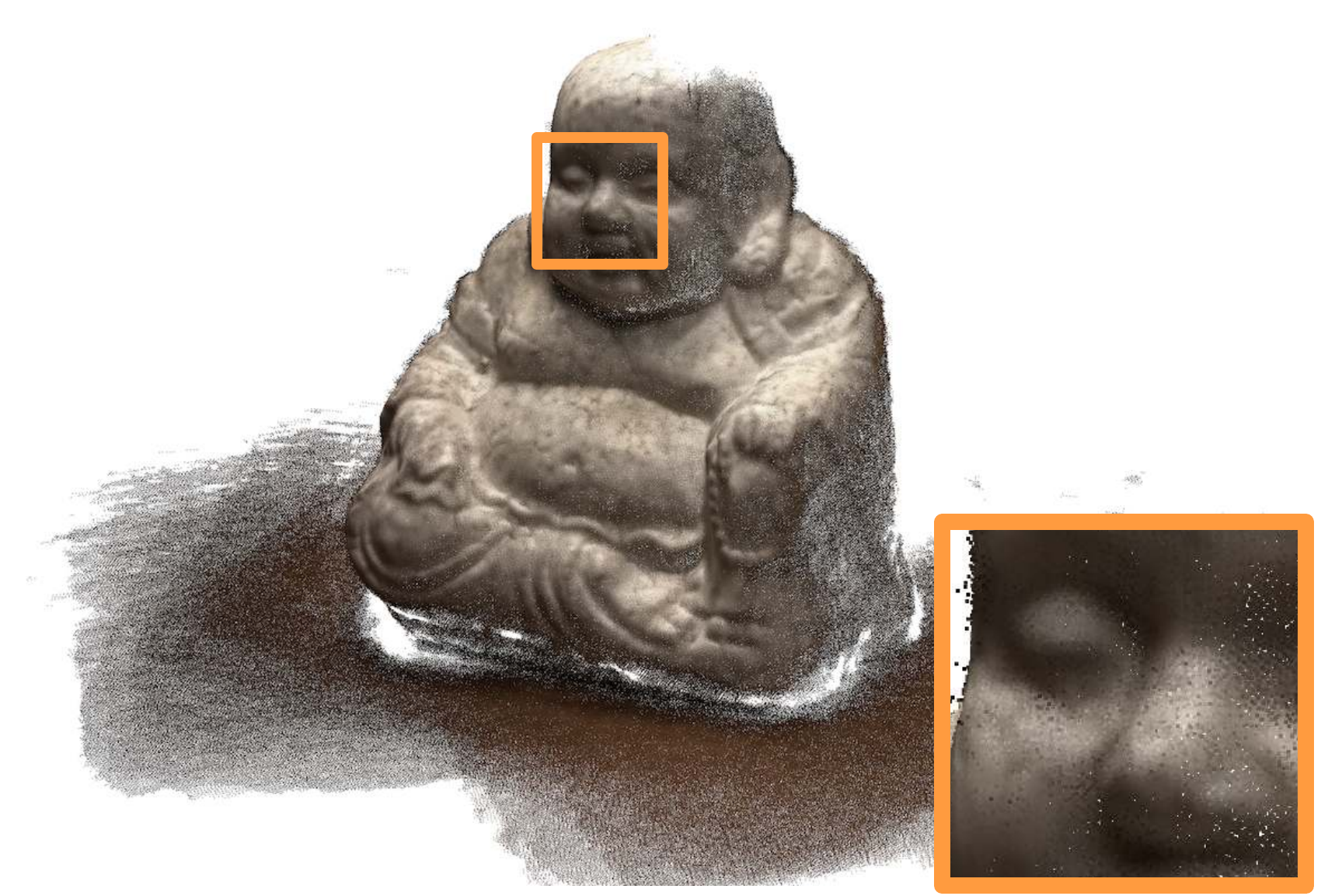}
      & \includegraphics[width=0.2\linewidth]{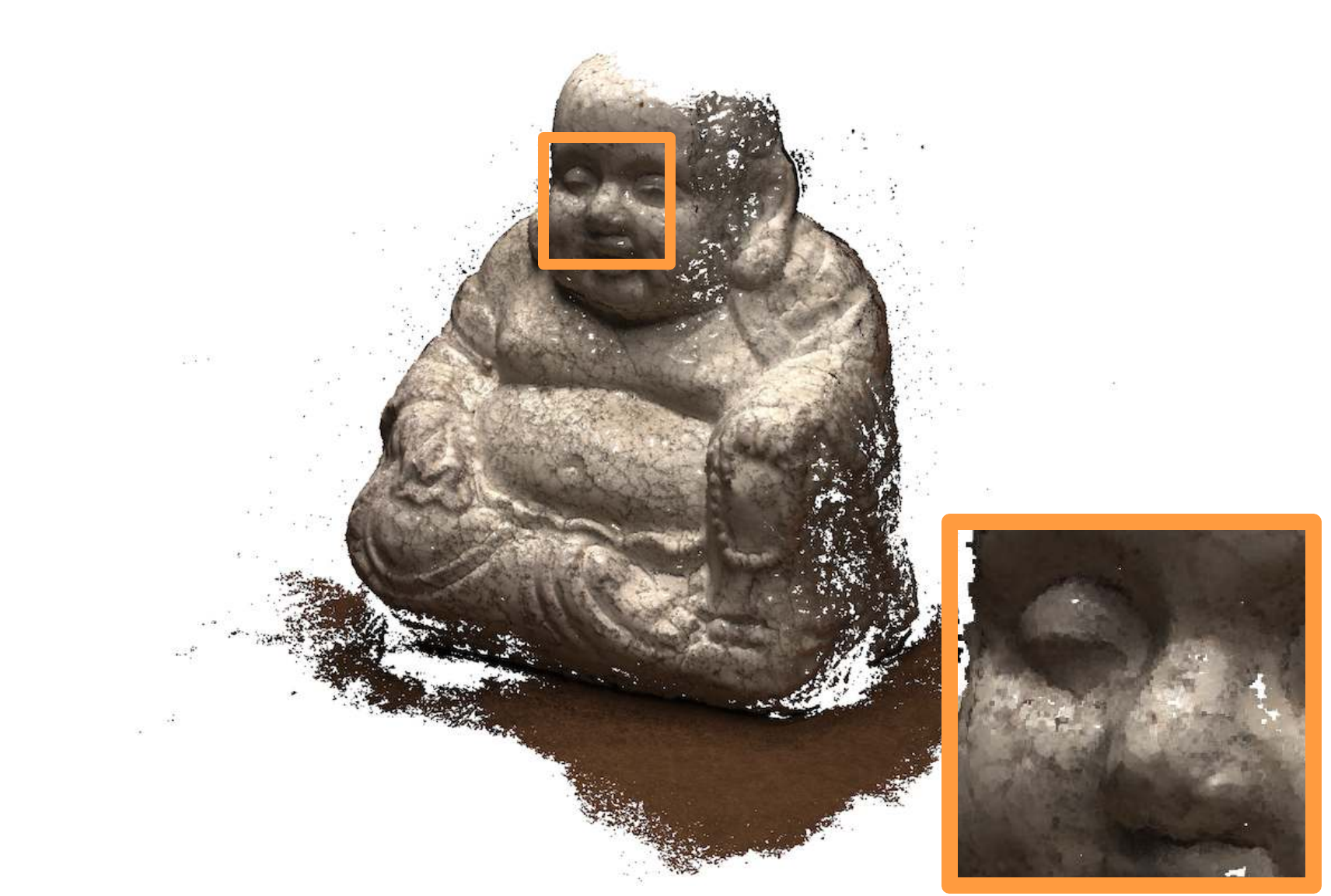}
      & \includegraphics[width=0.2\linewidth]{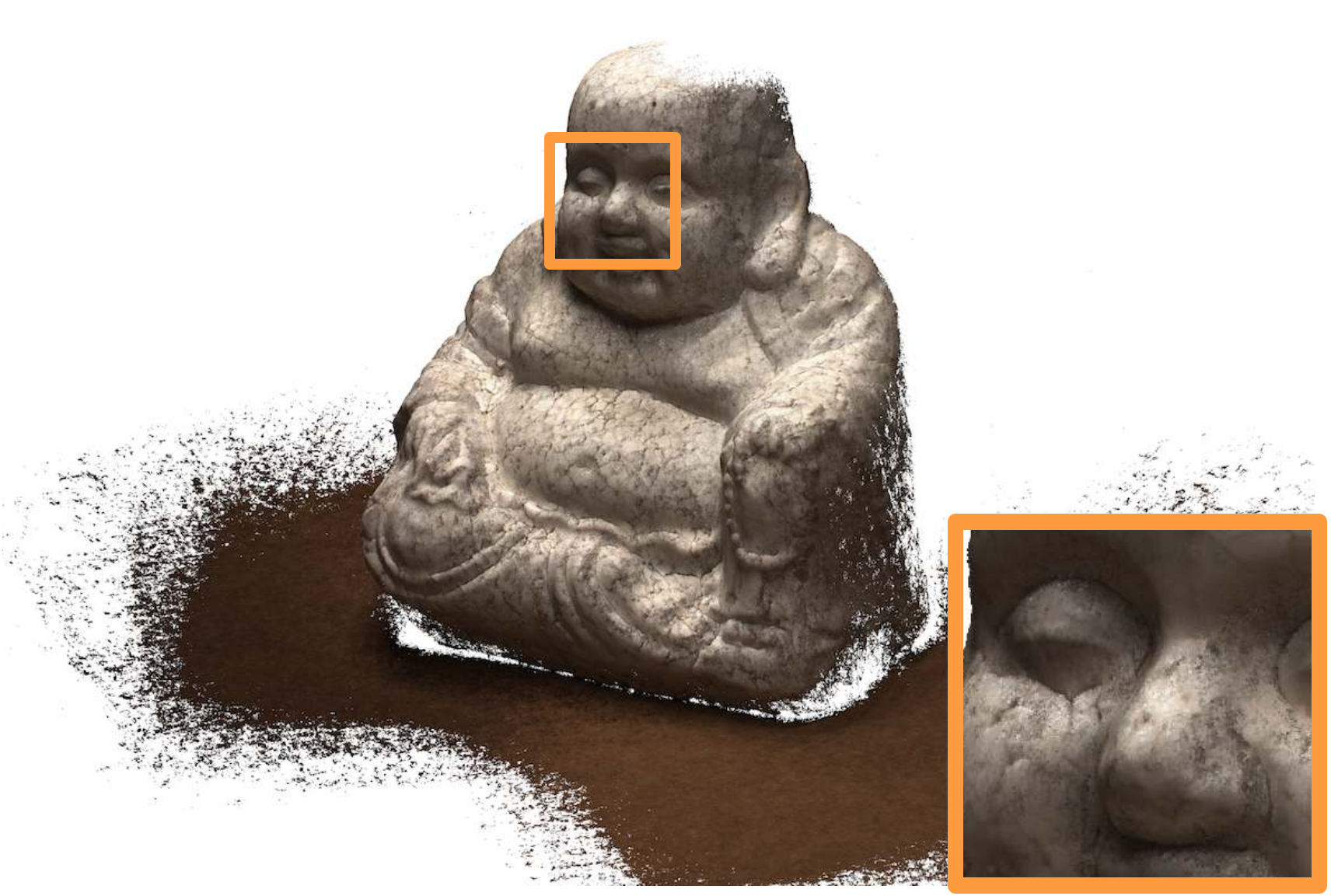}
      & \includegraphics[width=0.2\linewidth]{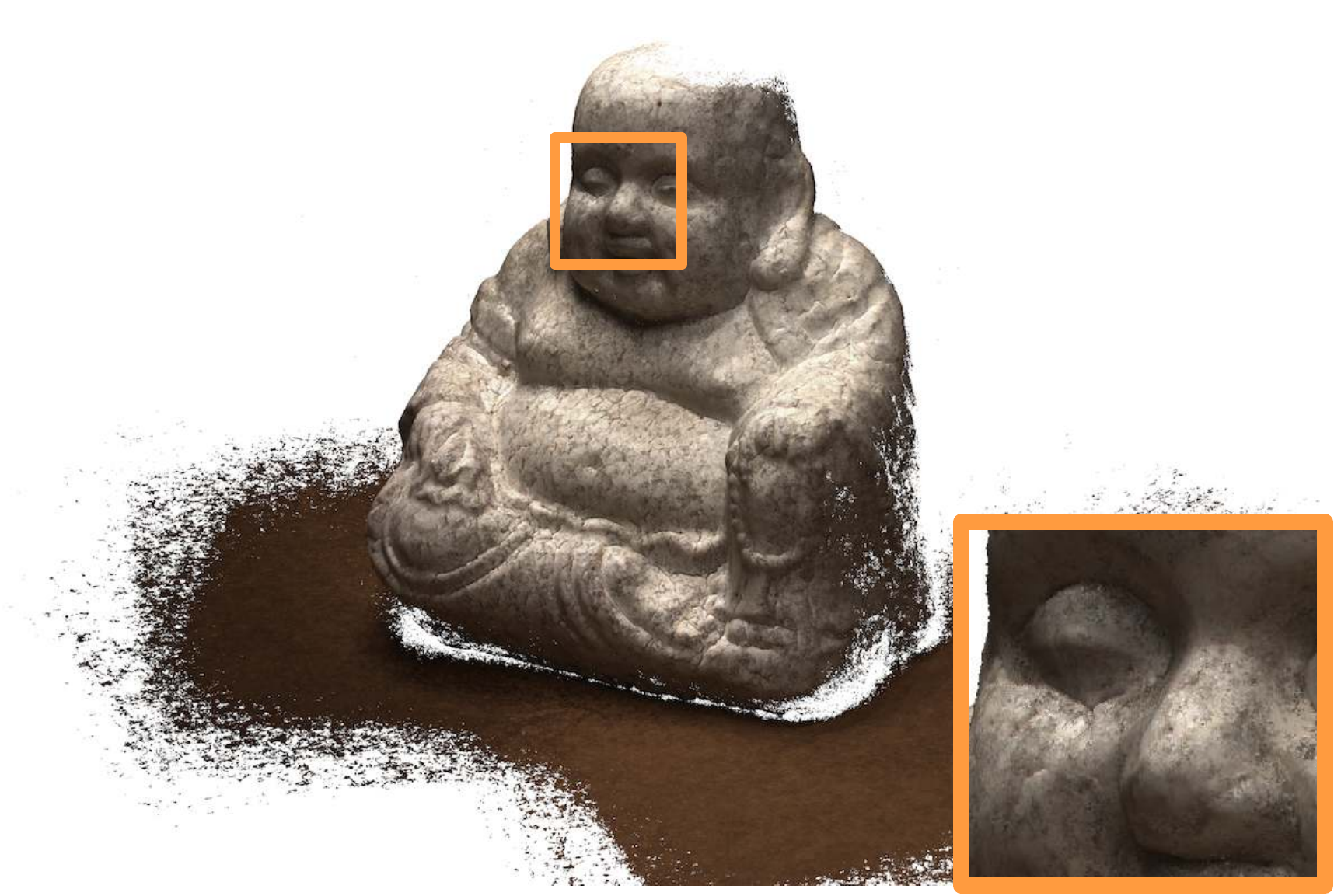}\\
      \includegraphics[width=0.2\linewidth]{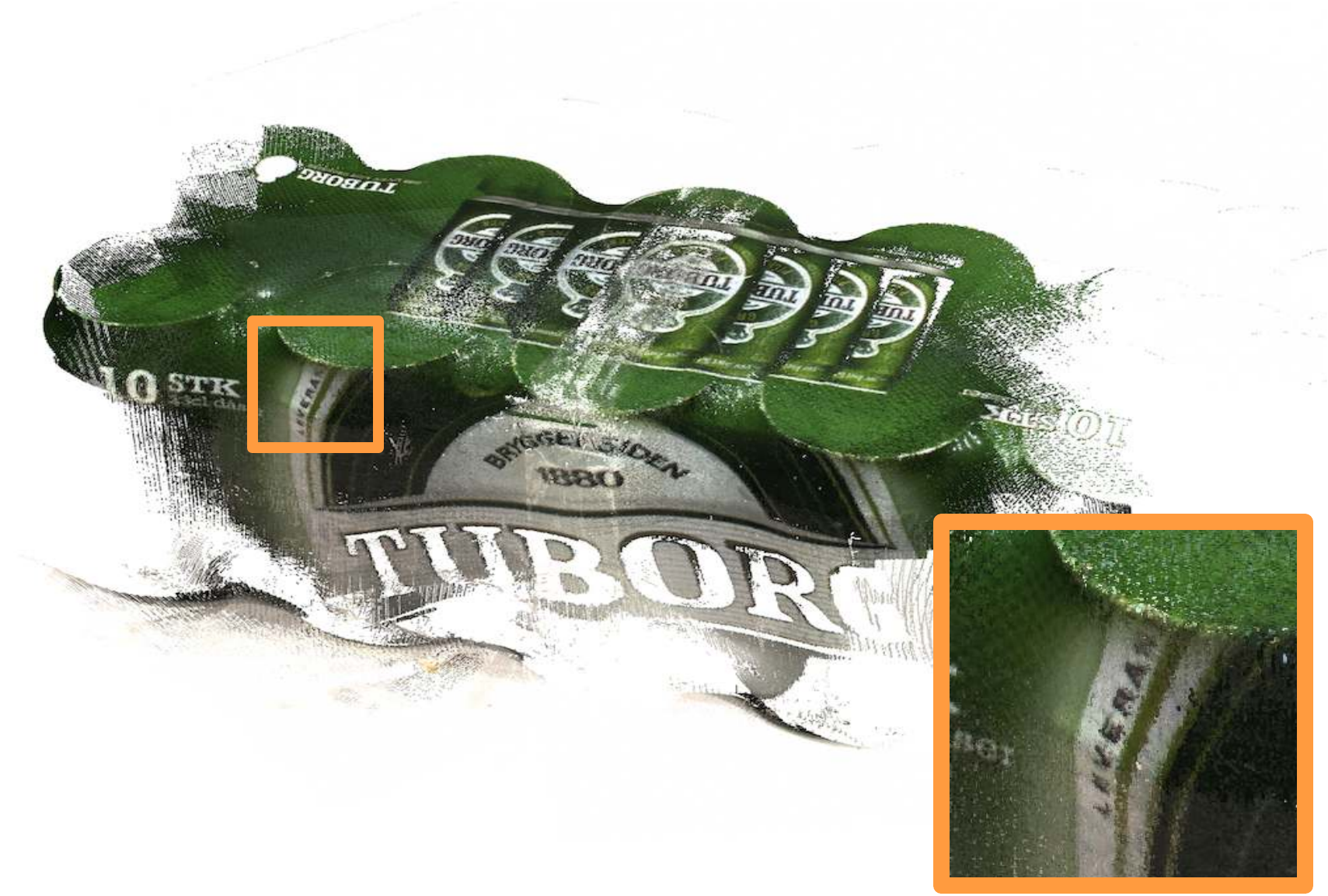}
      & \includegraphics[width=0.2\linewidth]{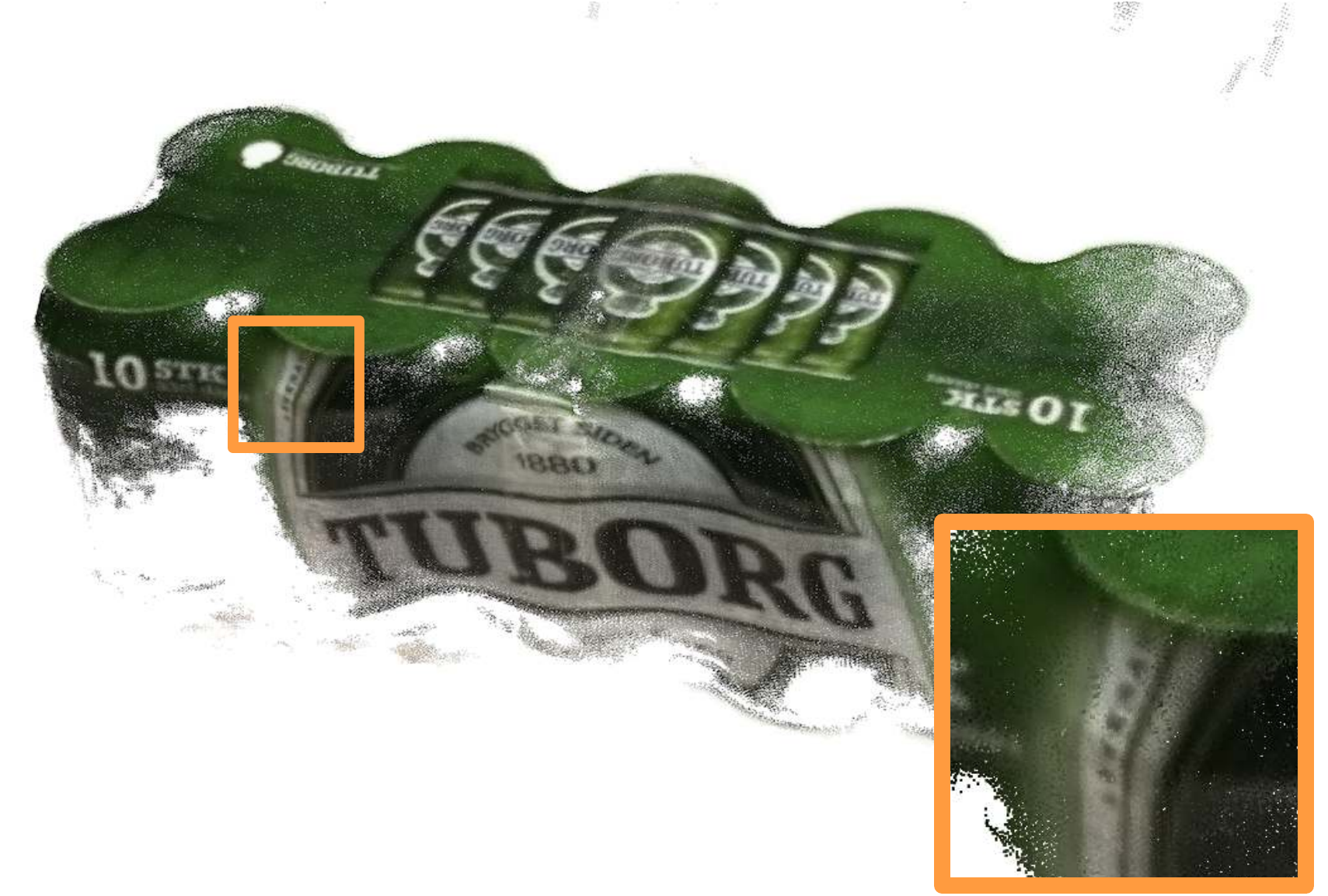}
      & \includegraphics[width=0.2\linewidth]{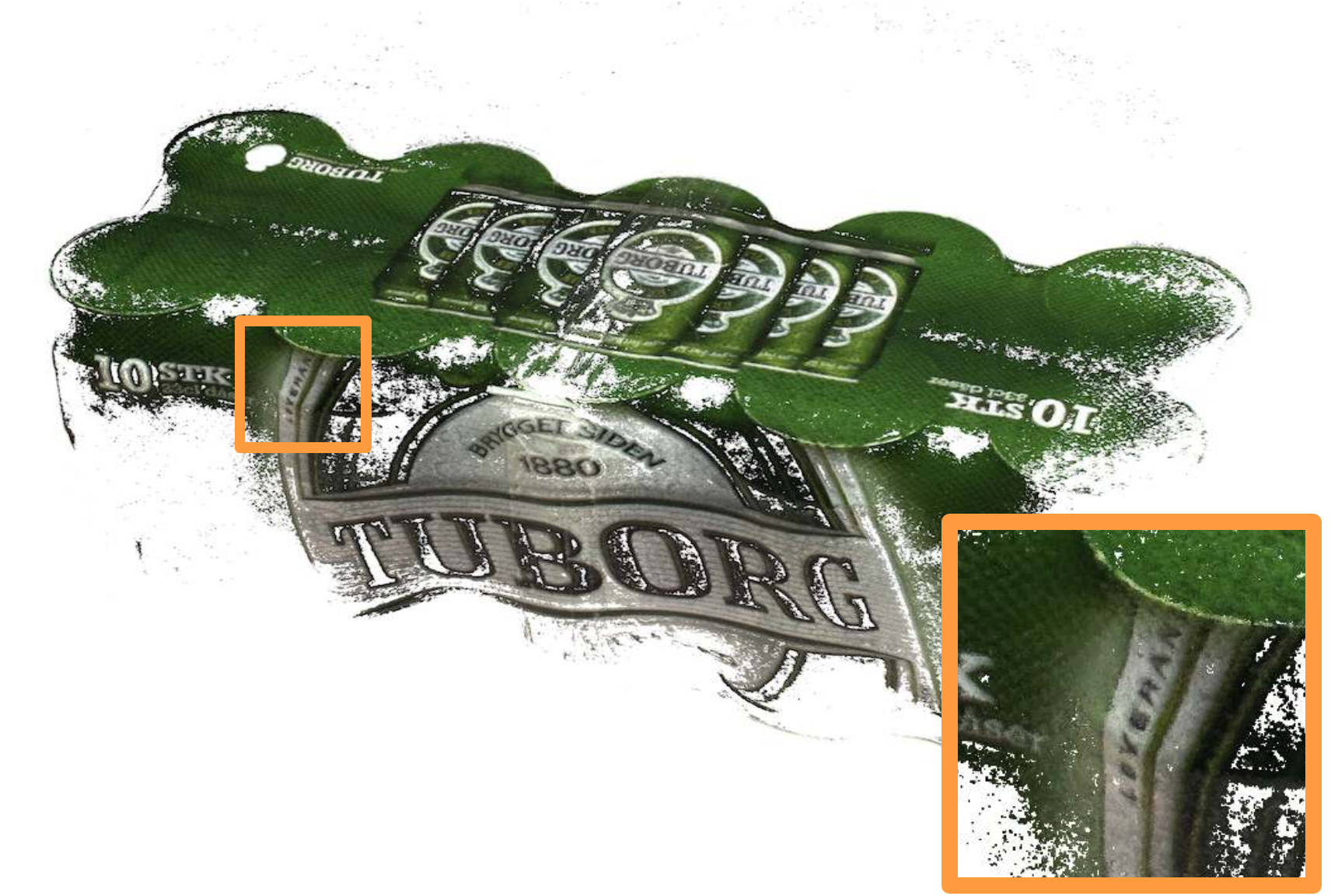}
      & \includegraphics[width=0.2\linewidth]{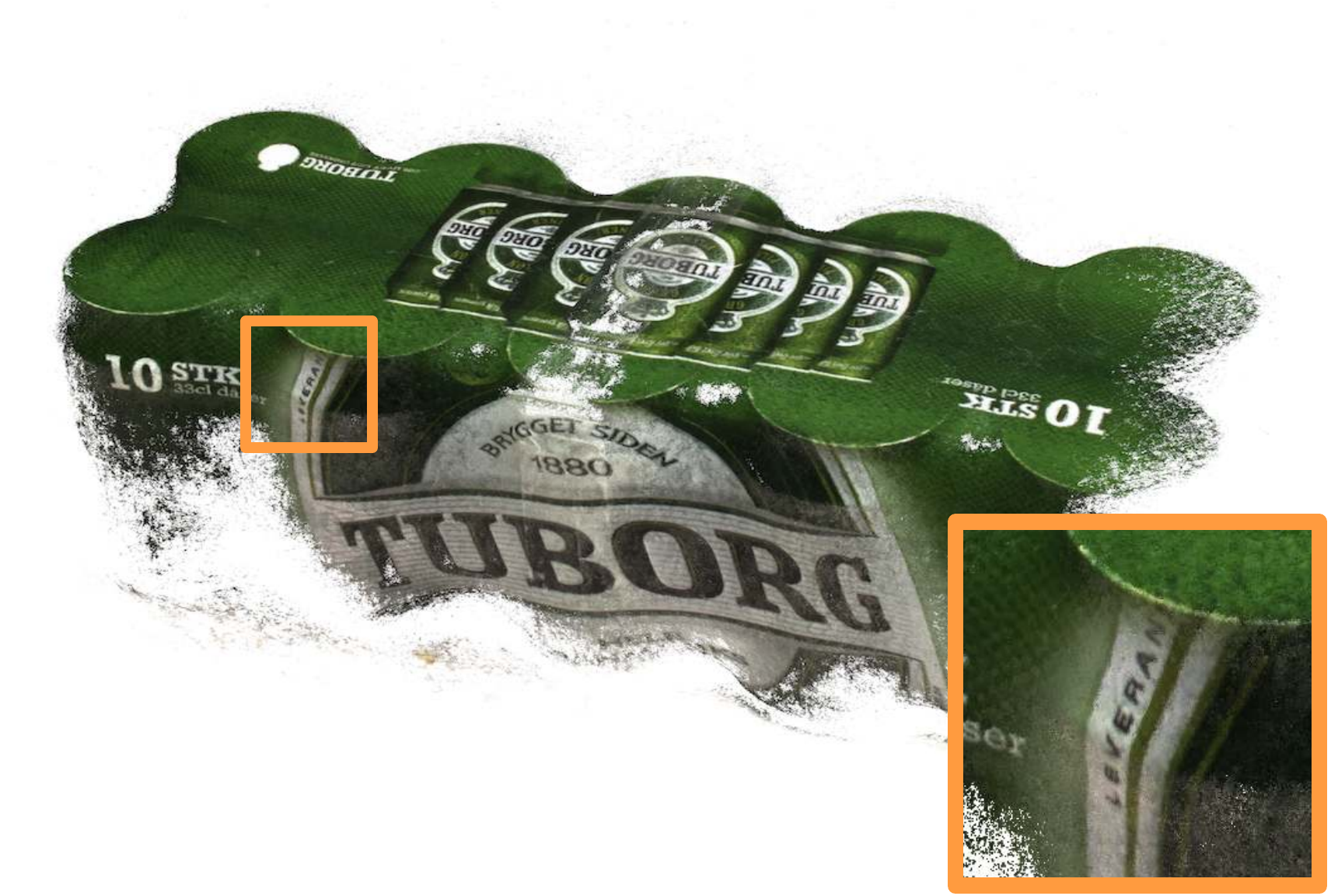}
      & \includegraphics[width=0.2\linewidth]{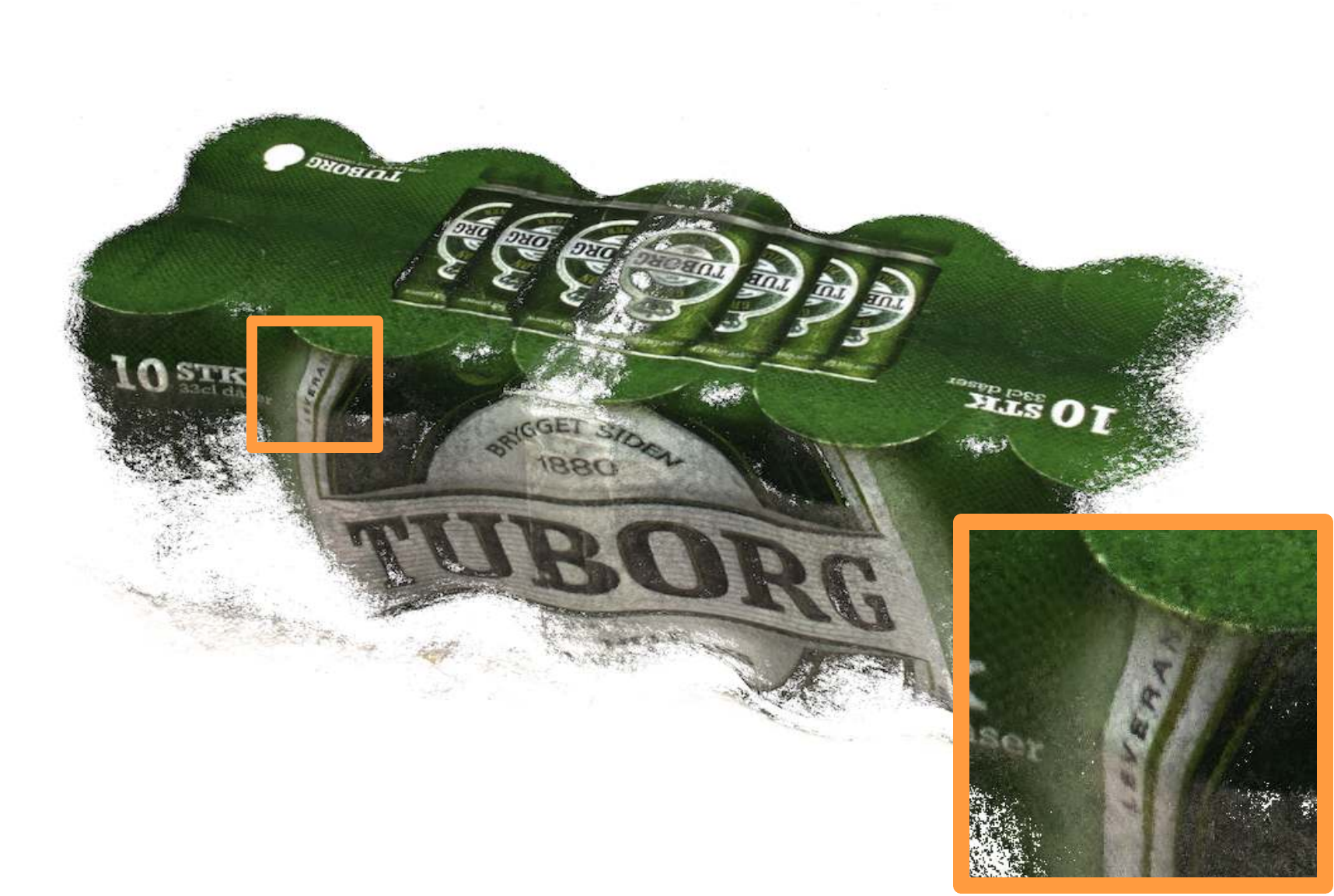}\\
      \includegraphics[width=0.2\linewidth]{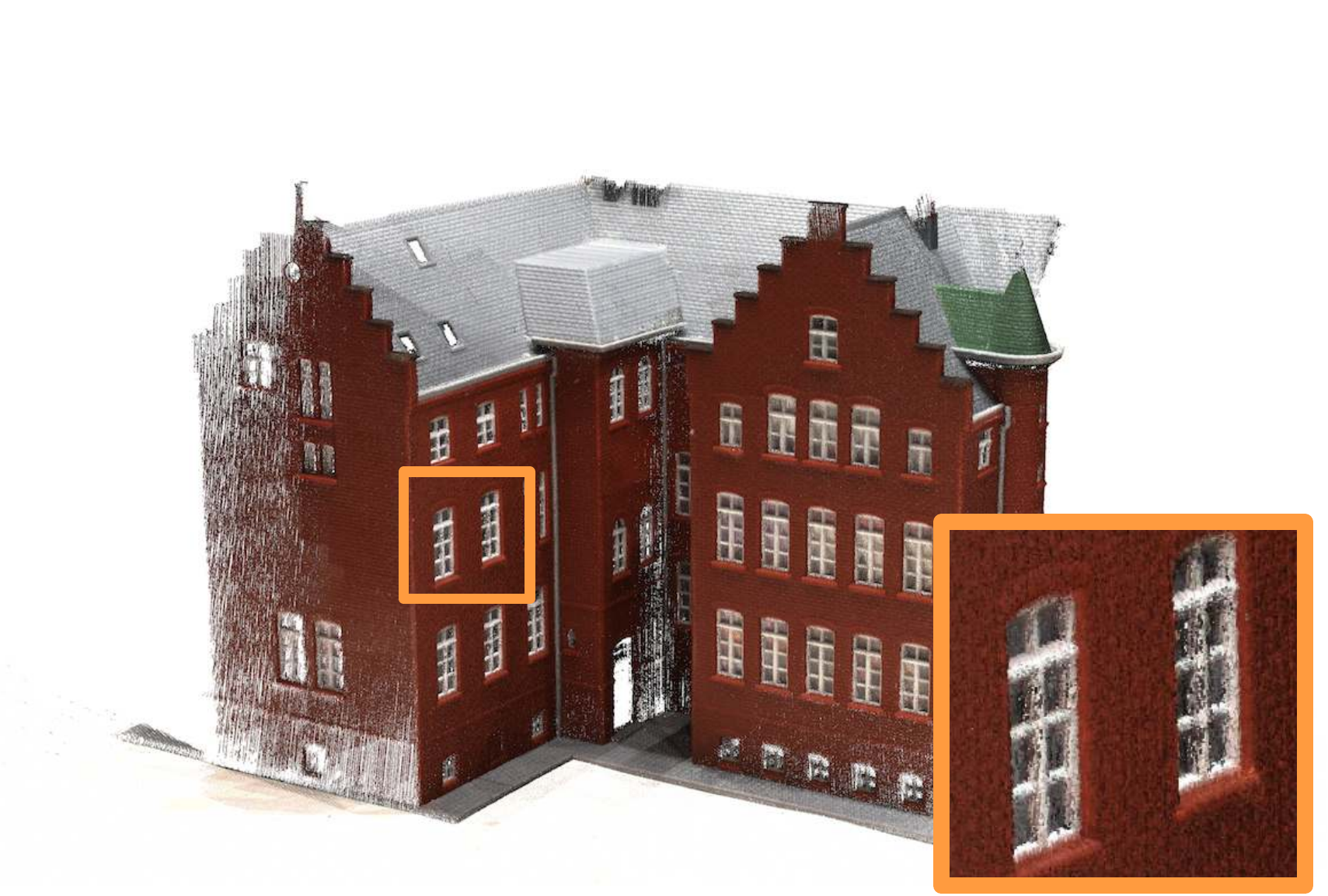}
      & \includegraphics[width=0.2\linewidth]{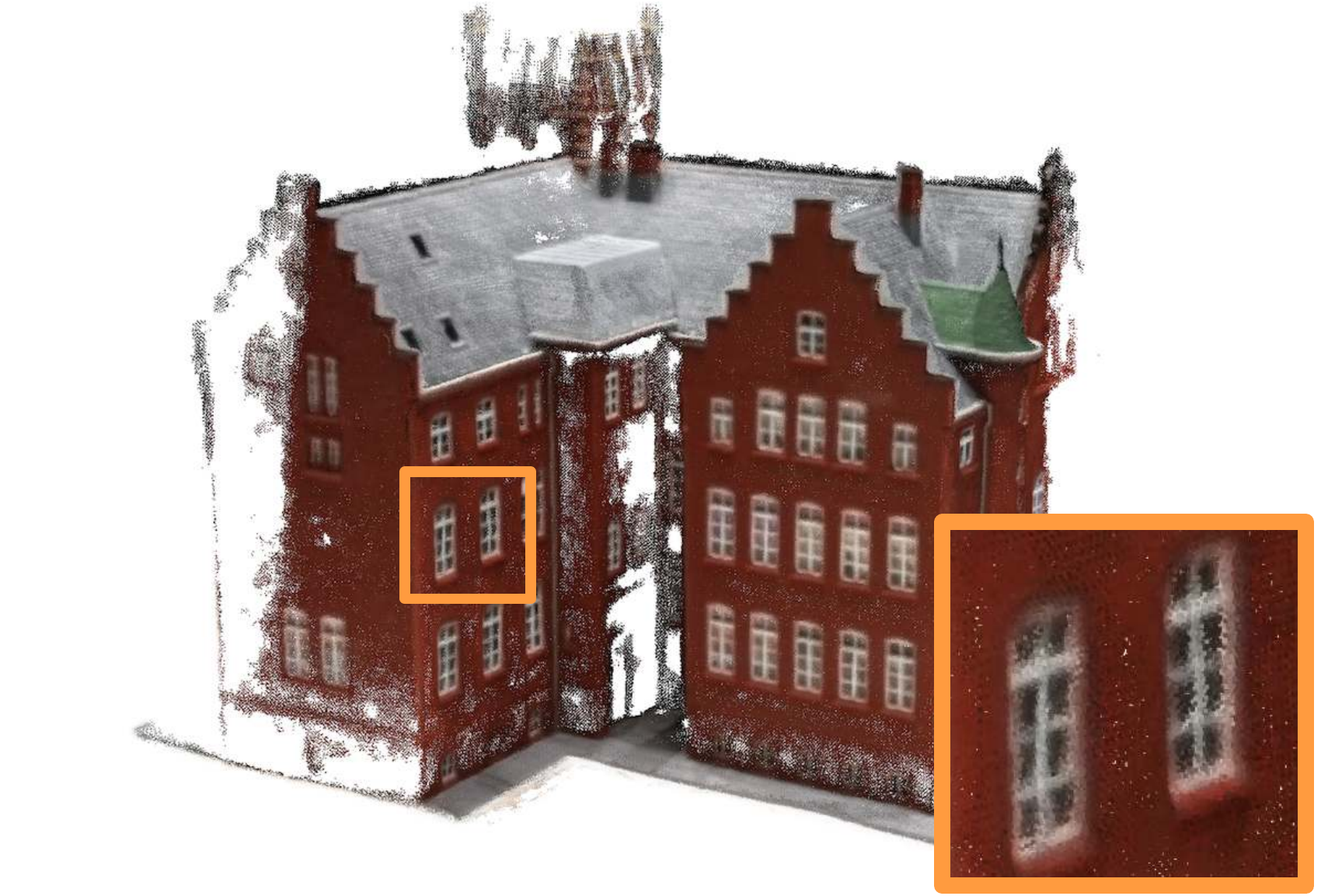}
      & \includegraphics[width=0.2\linewidth]{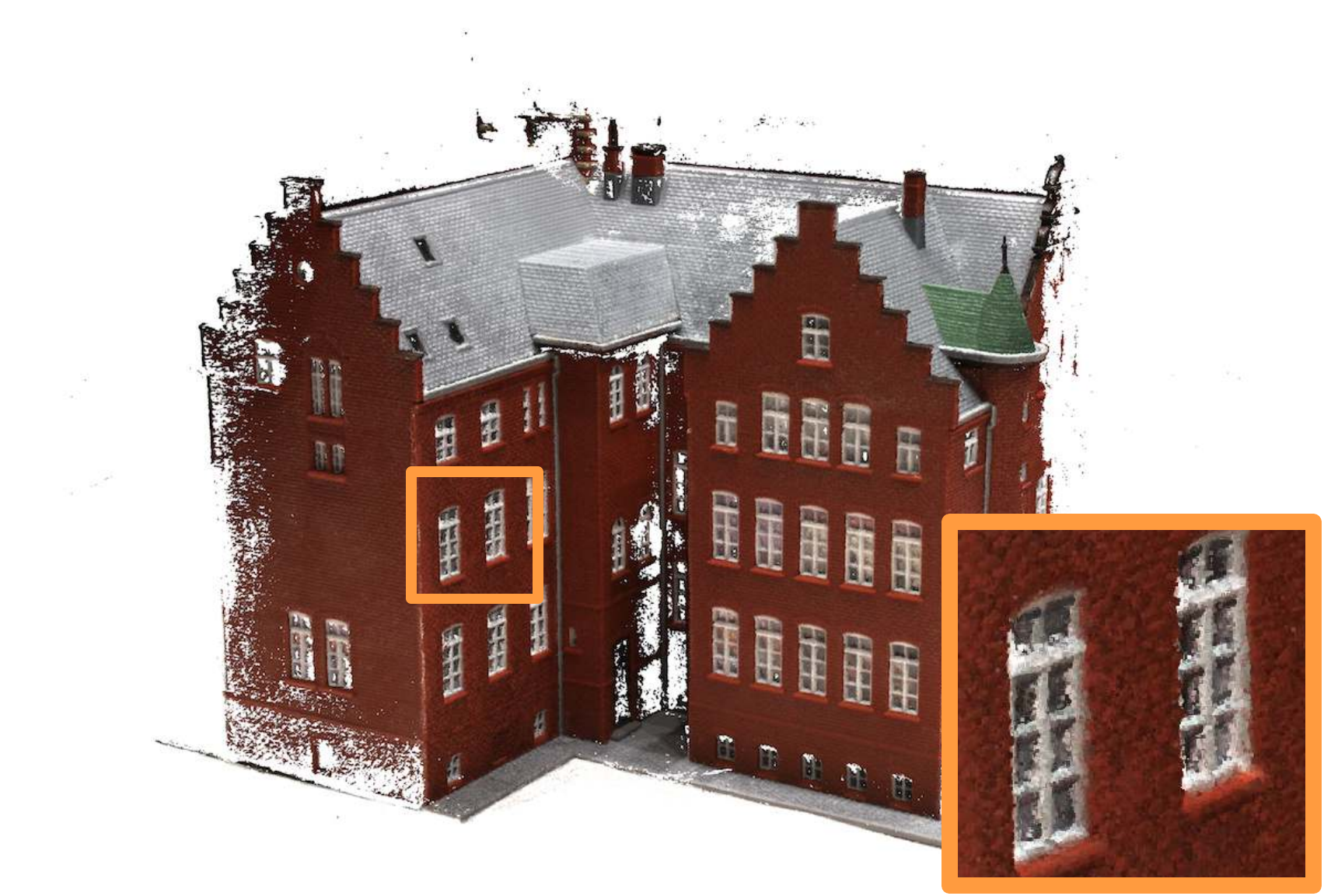}
      & \includegraphics[width=0.2\linewidth]{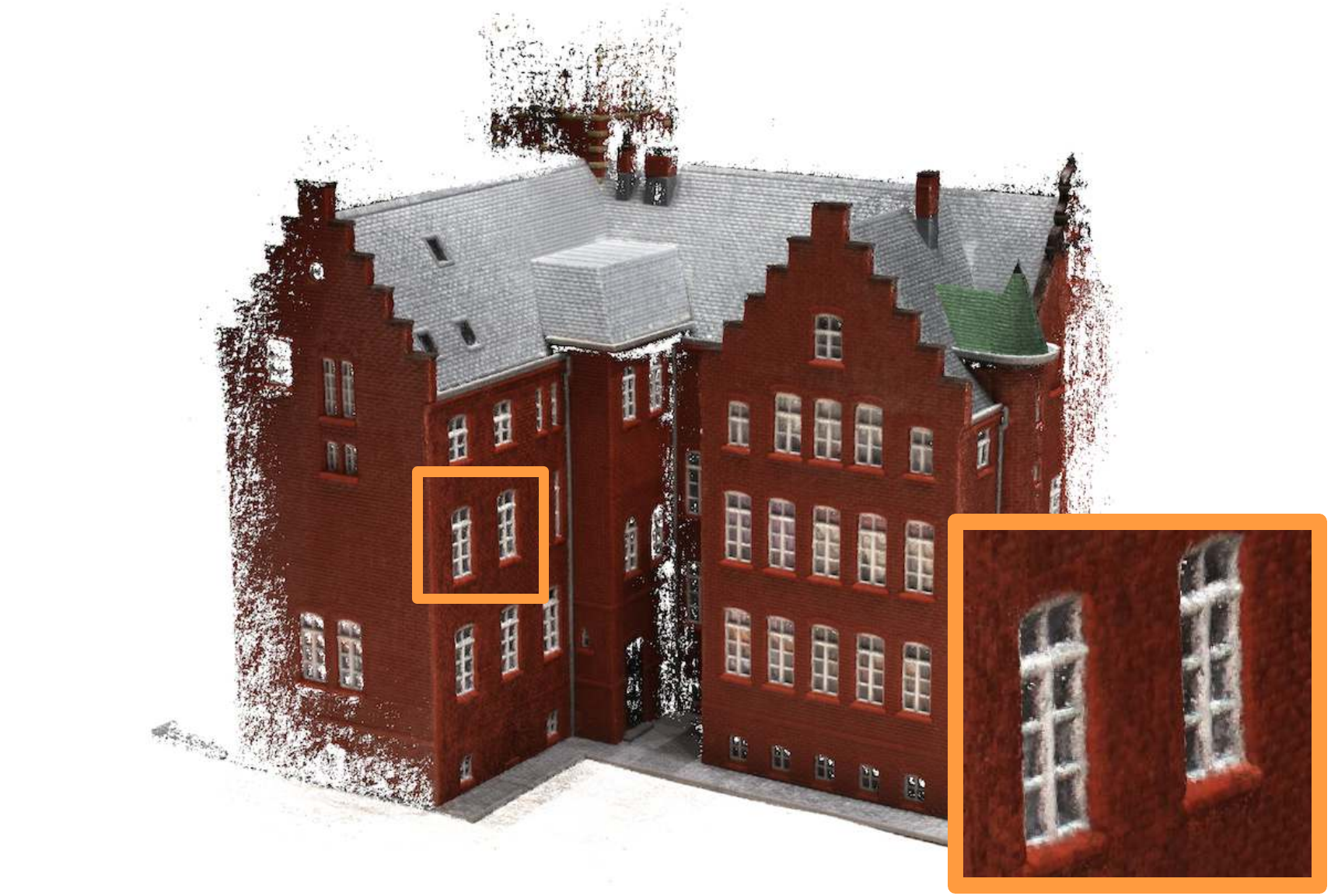}
      & \includegraphics[width=0.2\linewidth]{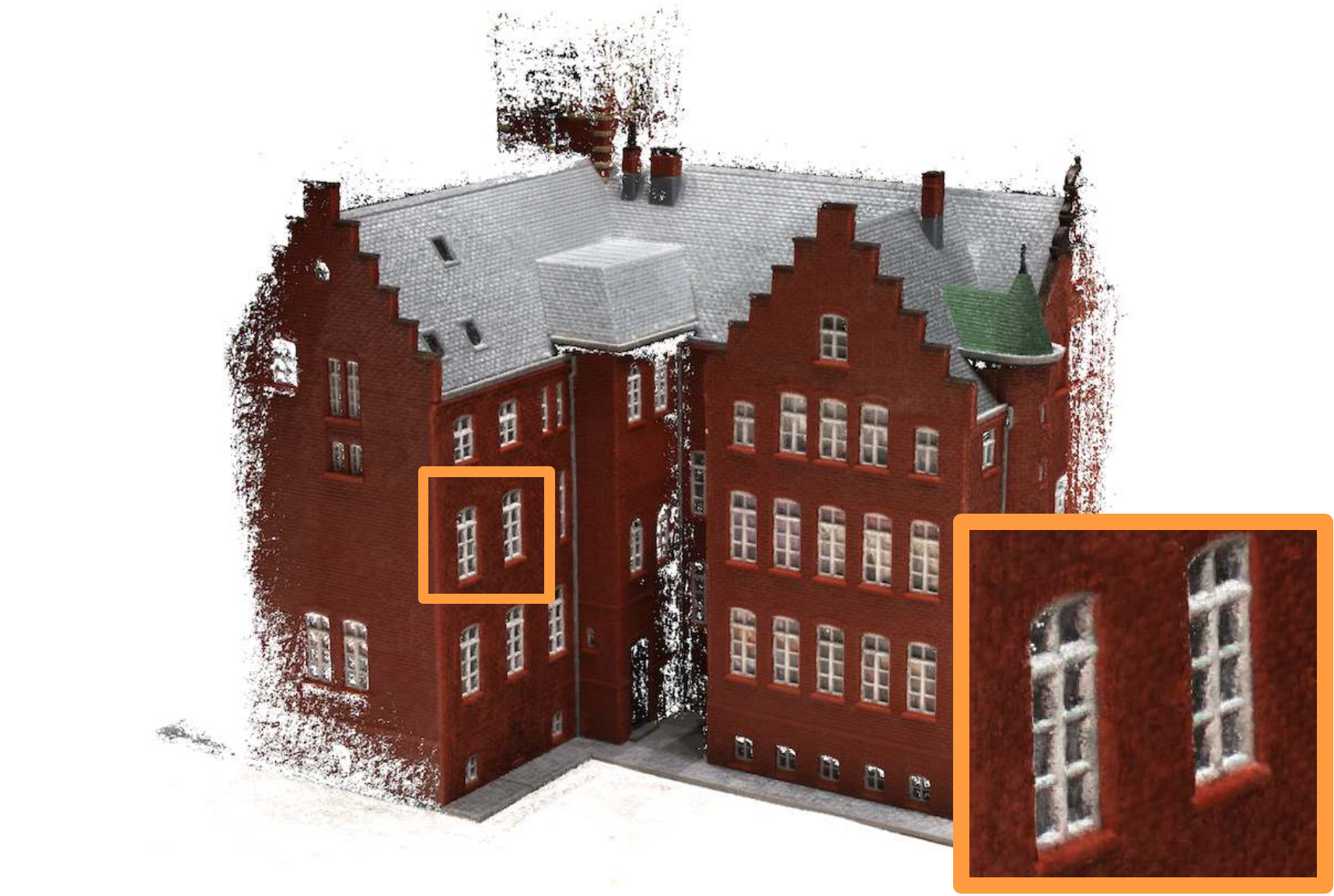}\\
      \includegraphics[width=0.2\linewidth]{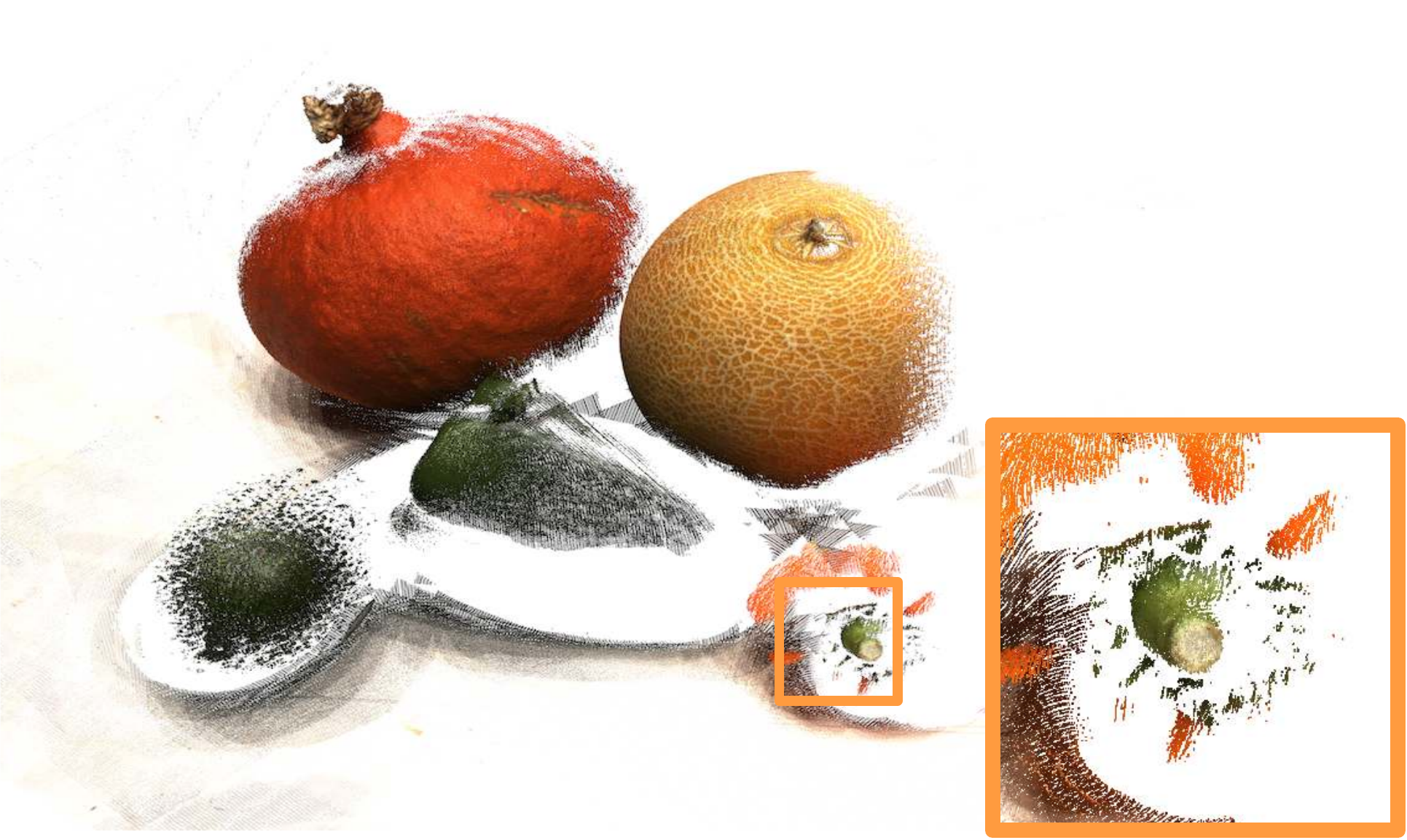}
      & \includegraphics[width=0.2\linewidth]{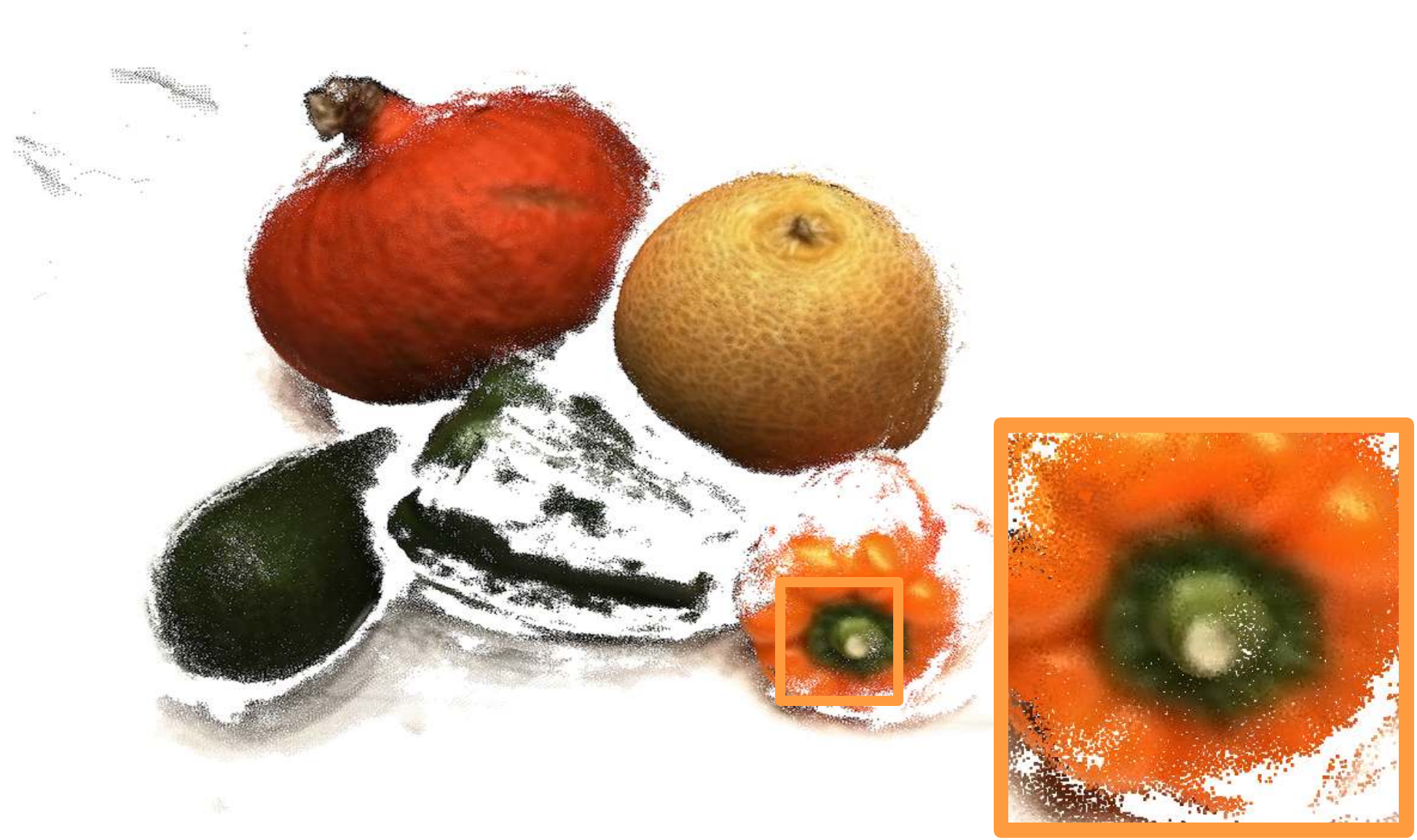}
      & \includegraphics[width=0.2\linewidth]{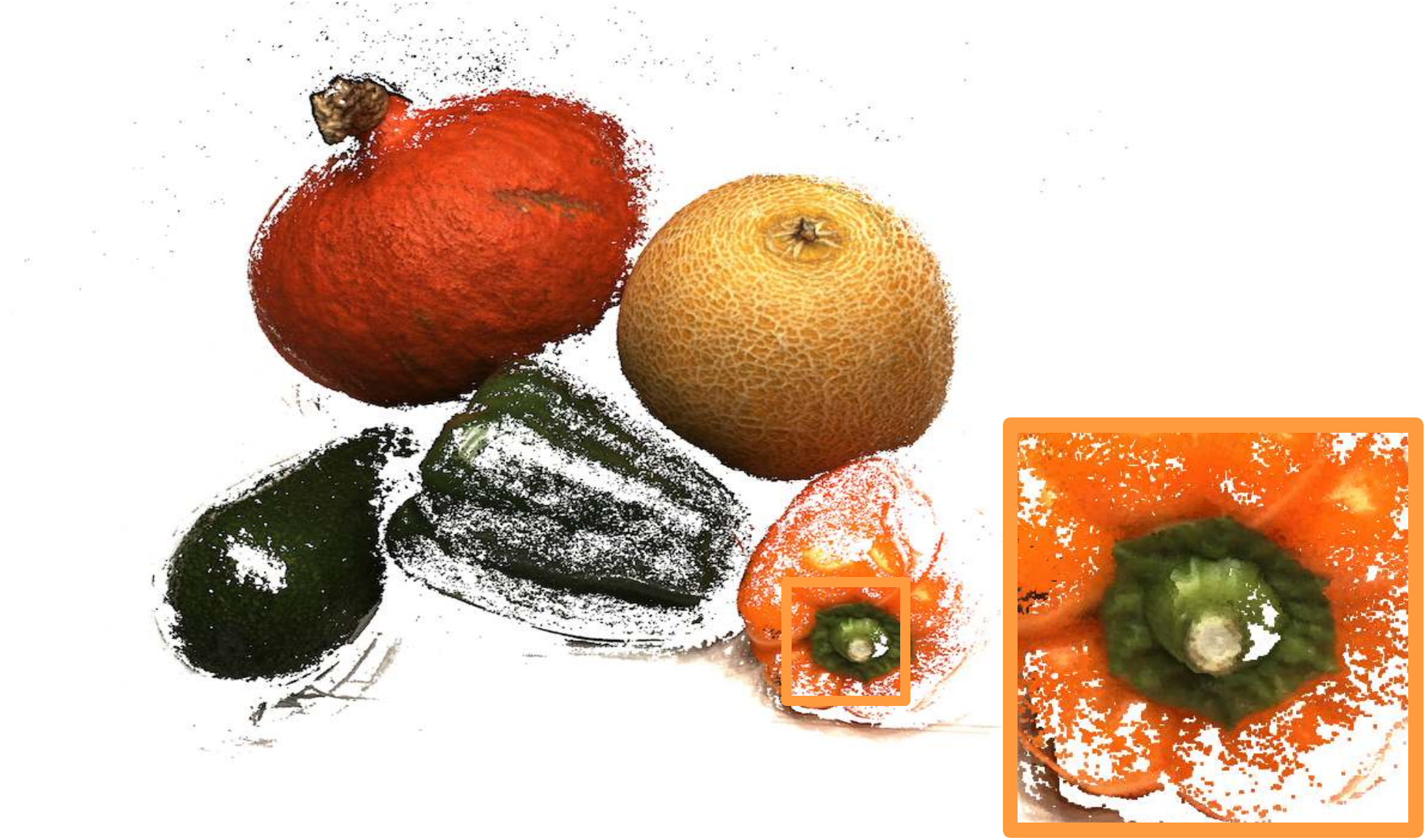}
      & \includegraphics[width=0.2\linewidth]{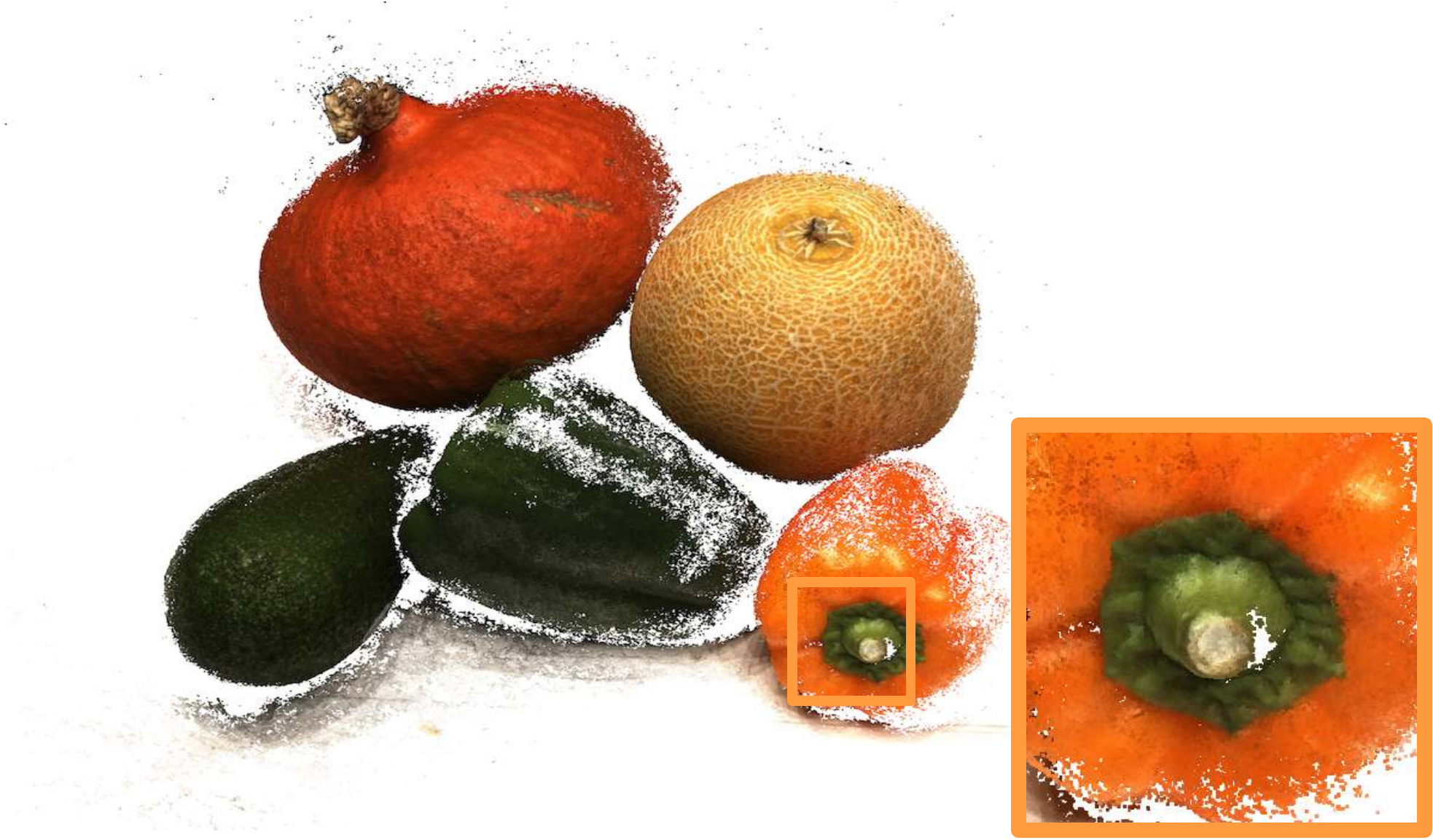}
      & \includegraphics[width=0.2\linewidth]{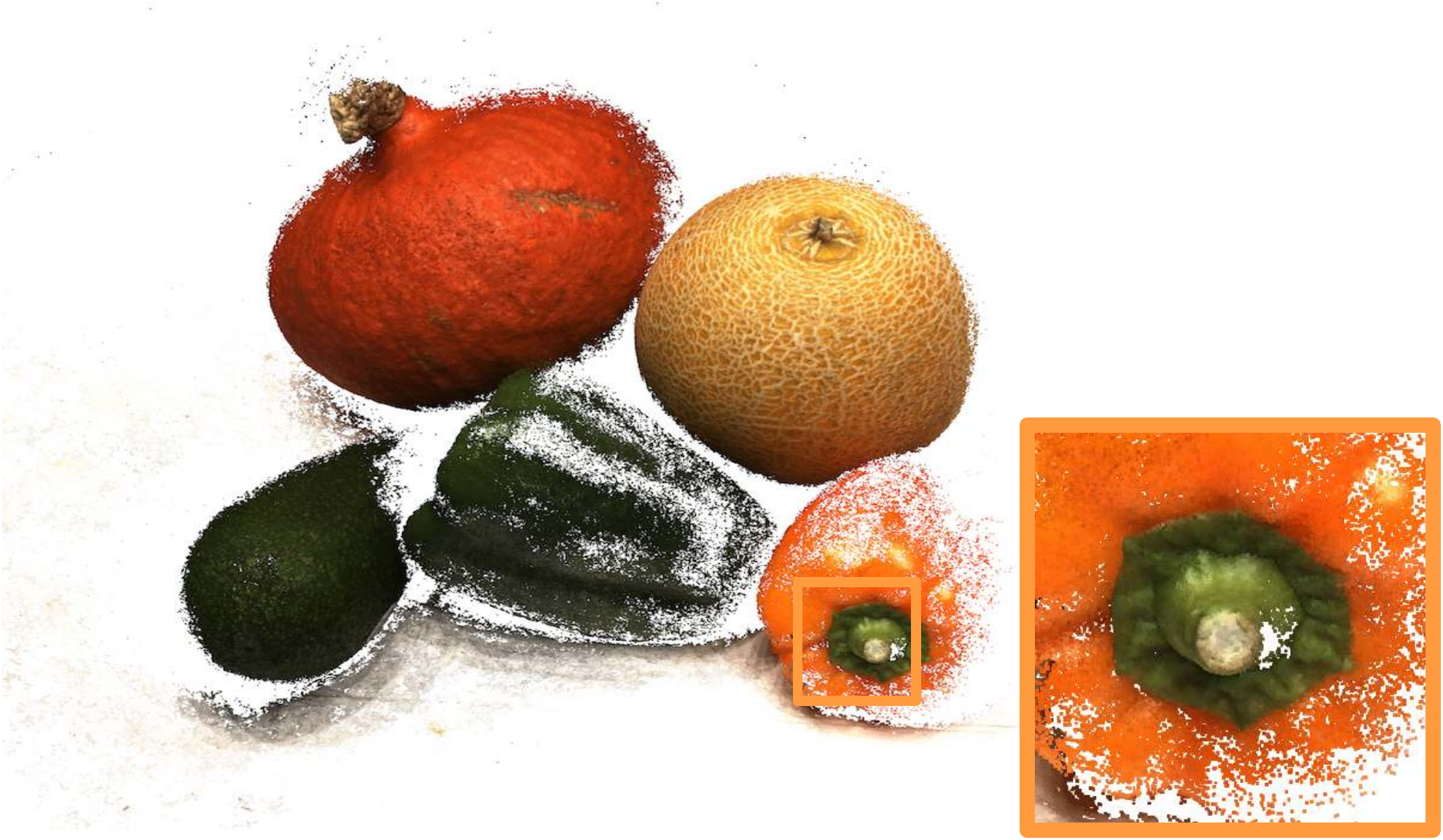}\\
      \includegraphics[width=0.2\linewidth]{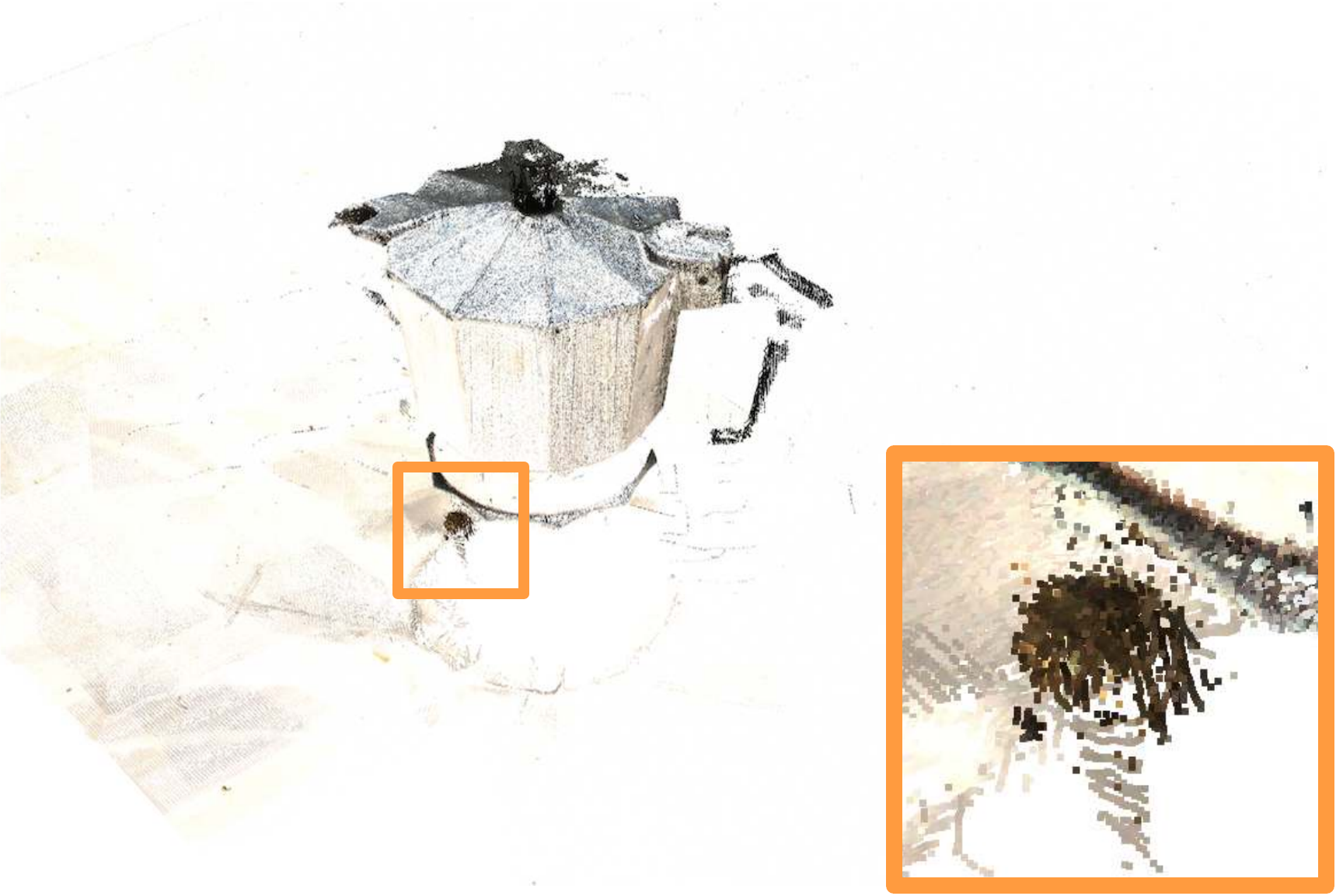}
      & \includegraphics[width=0.2\linewidth]{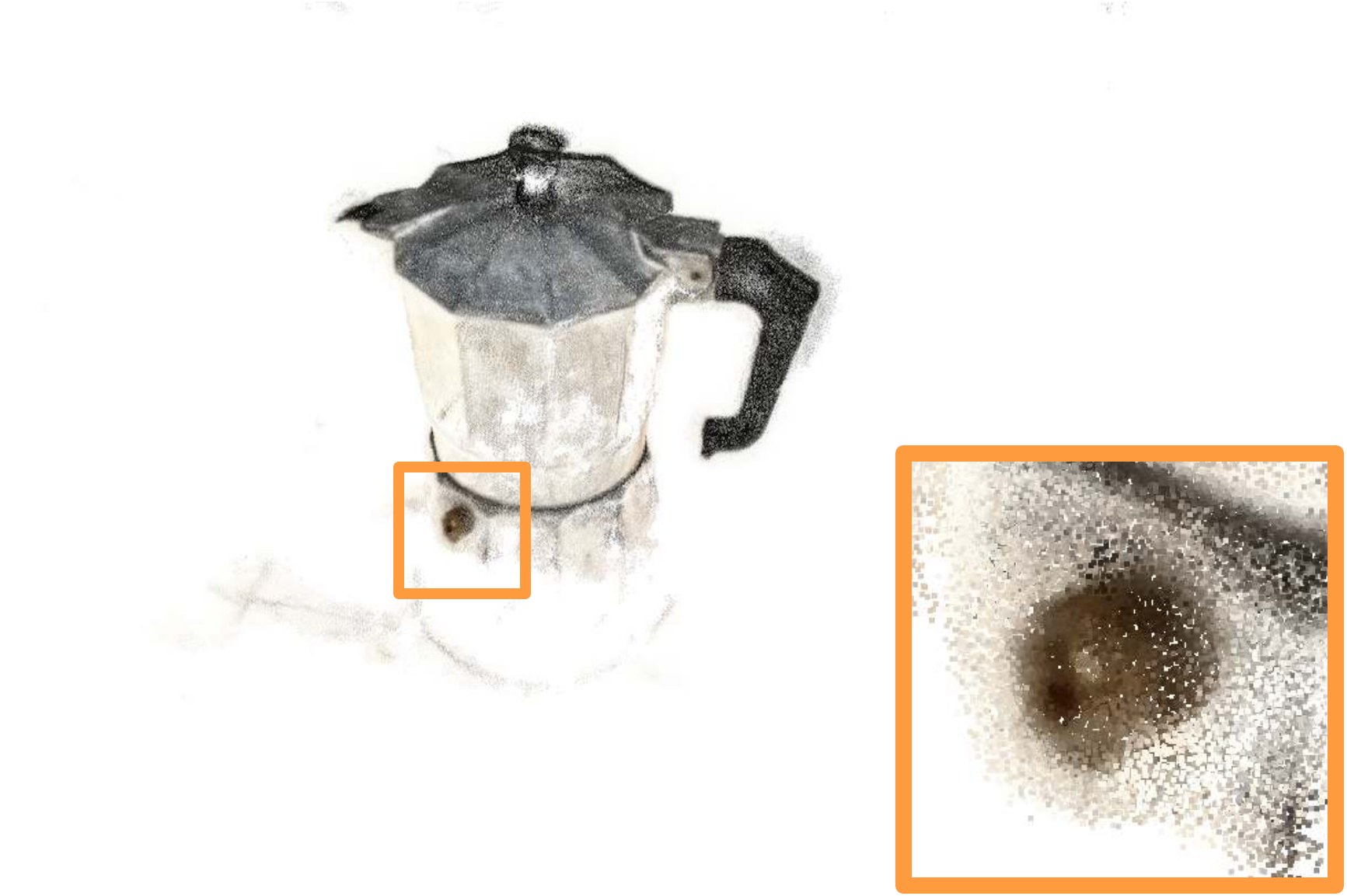}
      & \includegraphics[width=0.2\linewidth]{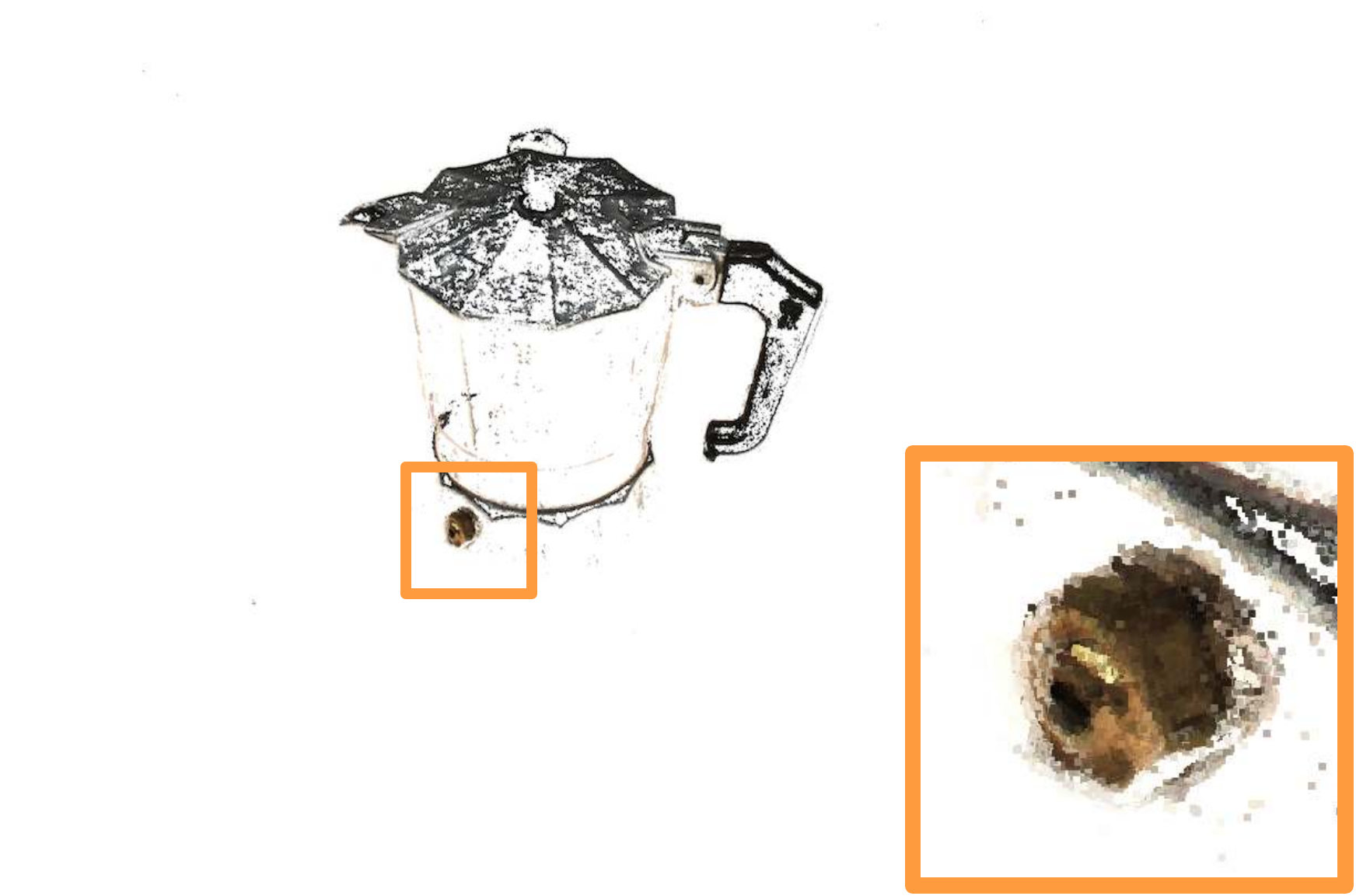}
      & \includegraphics[width=0.2\linewidth]{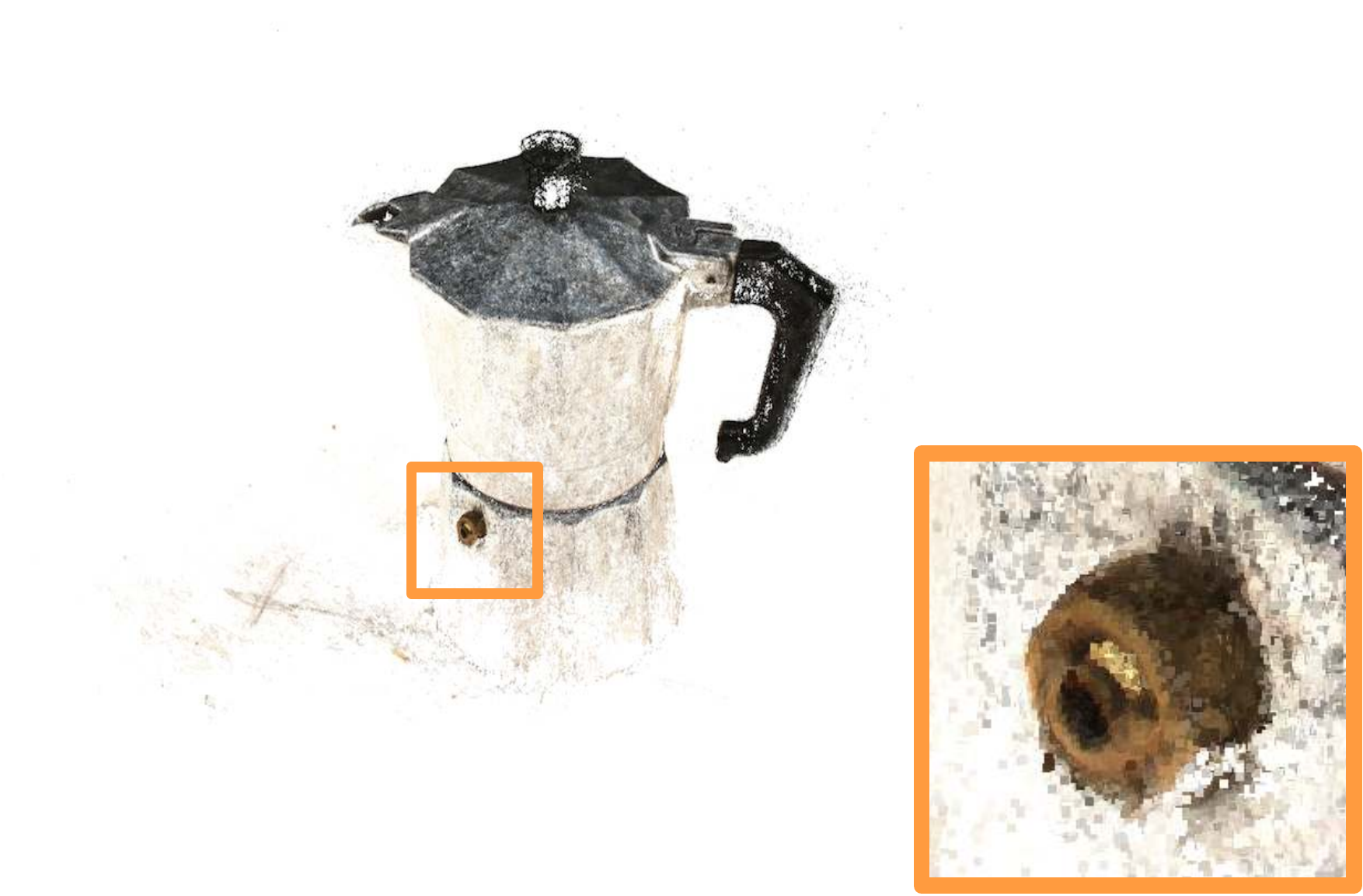}
      & \includegraphics[width=0.2\linewidth]{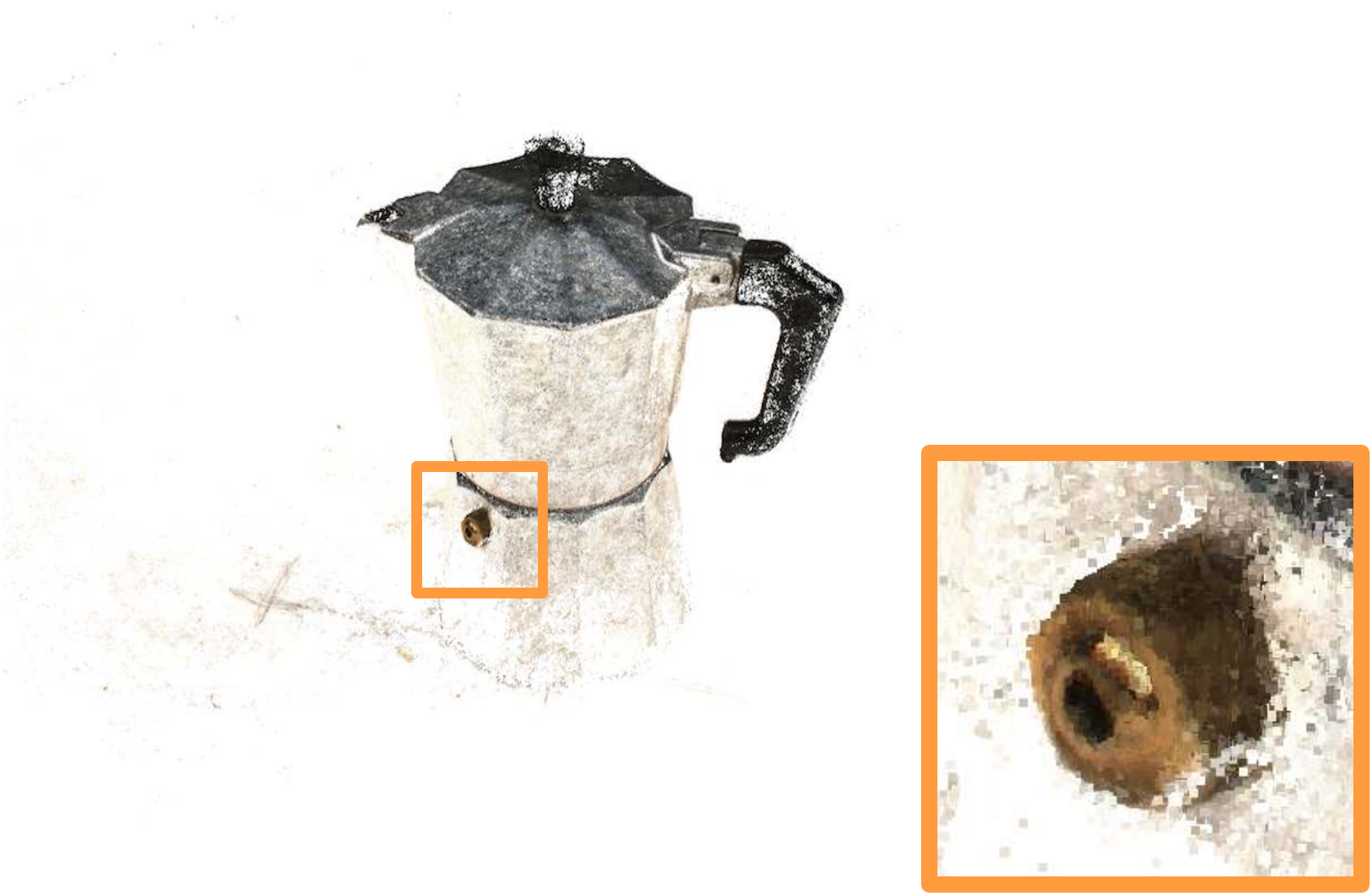}\\
      
      \multicolumn{1}{c}{Ground-truth} & \multicolumn{1}{c}{Khot \etal~\cite{Khot2019learning}} & \multicolumn{1}{c}{Ours} & \multicolumn{1}{c}{Ours} & \multicolumn{1}{c}{CVP-MVSNet~\cite{Yang2020CVP}}\\
      \multicolumn{1}{c}{Point Cloud} & \multicolumn{1}{c}{(Unsupervised)} & \multicolumn{1}{c}{(Unsupervised)} & \multicolumn{1}{c}{(Self-supervised)} & \multicolumn{1}{c}{(Supervised)}
\end{tabular}
    \end{center}
    \caption{\textbf{DTU Dataset.} Representative point cloud results. Best viewed on screen.}
    \label{fig:more_dtu}
    \vspace{4cm}
\end{figure*}

Fig.~\ref{fig:more_dtu} shows additional reconstruction results by our self-supervised method on the DTU dataset. As shown, our self-supervised model can achieve similar reconstruction results comparing with the supervised model.

\section{Pseudo depth labels visualization}\label{sec:pseudo_many}
\begin{figure*}[htbp]
    \vspace{1cm}
    \begin{center}
    \setlength\tabcolsep{0pt}
    \includegraphics[width=\linewidth]{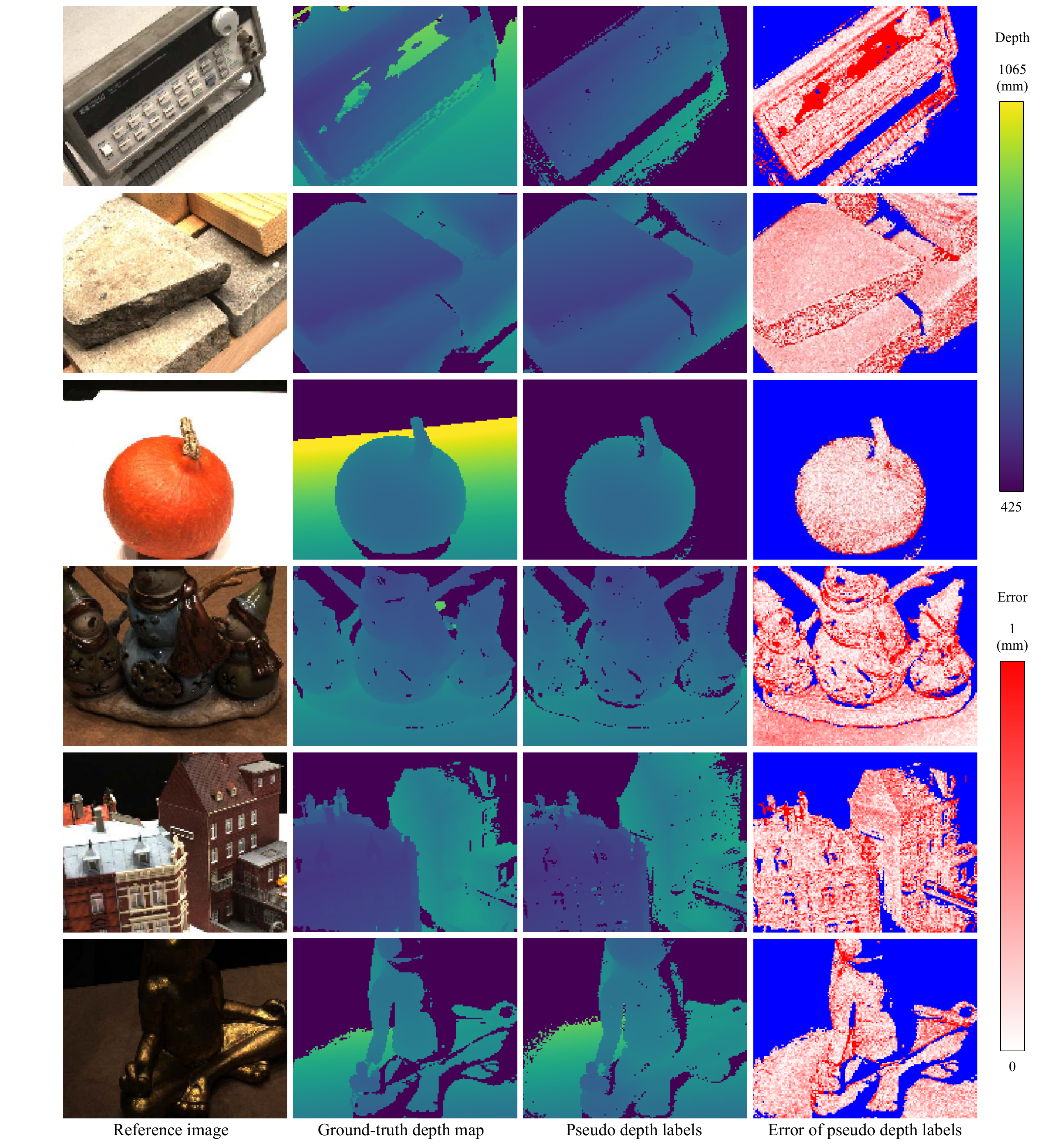}
    \end{center}
    \vspace{-0.5cm}
    \caption{Pseudo depth labels generated by the self-supervised learning framework. Areas with no pseudo depth labels or no ground-truth depth are marked as blue in the error visualization. Best viewed on screen.}
    \label{fig:pseudo_depth_many}
    \vspace{4cm}
\end{figure*}

\begin{figure*}[htbp]
    \vspace{1cm}
    \begin{center}
    \setlength\tabcolsep{0pt}
    \includegraphics[width=\linewidth]{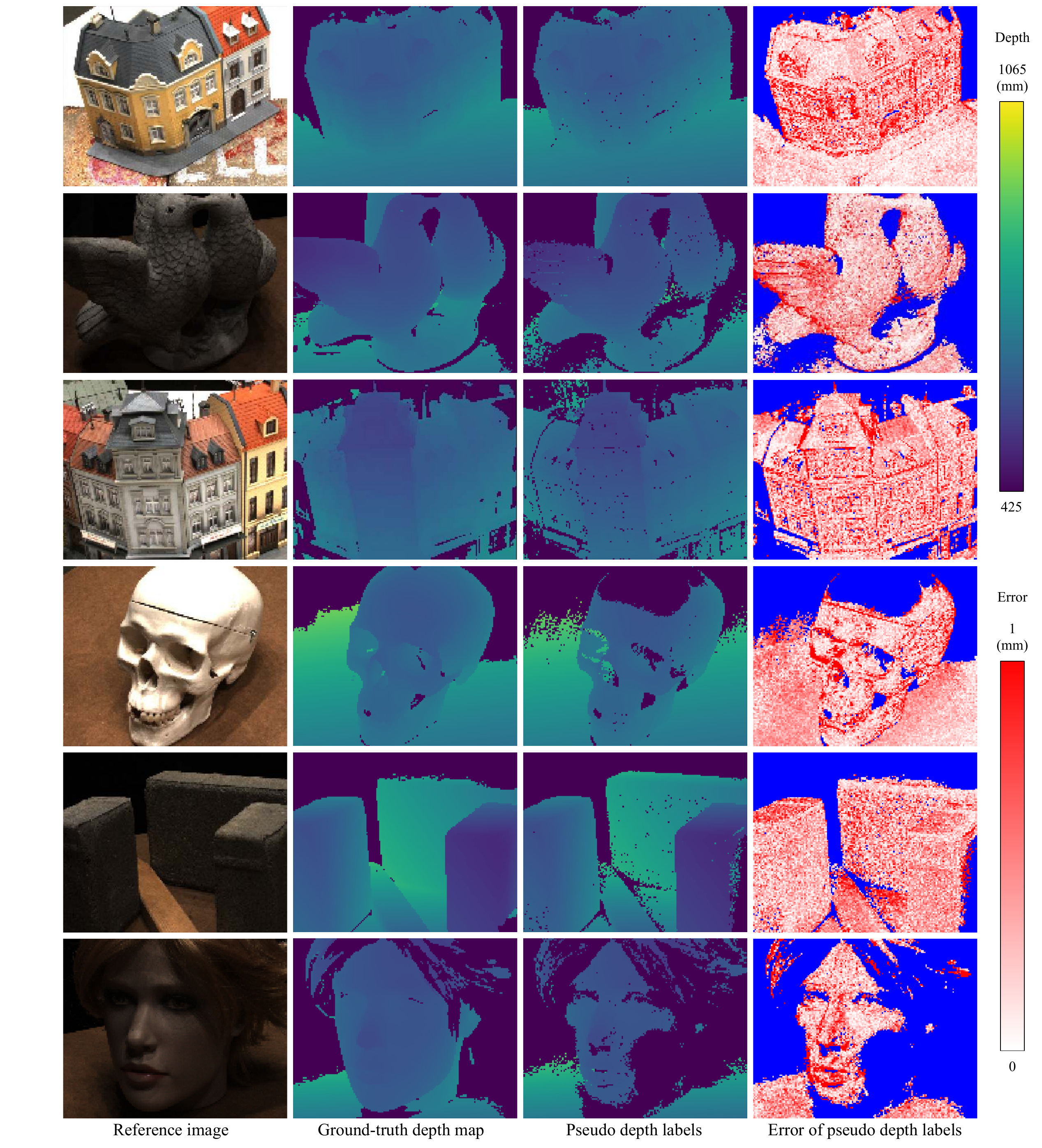}
    \end{center}
    \vspace{-0.5cm}
    \caption{Pseudo depth labels generated by the self-supervised learning framework. Areas with no pseudo depth labels or no ground-truth depth are marked as blue in the error visualization. Best viewed on screen.}
    \label{fig:pseudo_depth_many2}
    \vspace{4cm}
\end{figure*}

Fig.~\ref{fig:pseudo_depth_many} and Fig.~\ref{fig:pseudo_depth_many2} provide visualization of pseudo depth labels generated by our self-supervised learning framework. As shown, our method can generate high quality pseudo depth labels for rich-texture areas. However, our method can not generate pseudo depth labels for severely texture-less regions.

\begin{figure*}[htbp]
    \begin{center}
    \setlength\tabcolsep{0pt}
    \includegraphics[width=\linewidth]{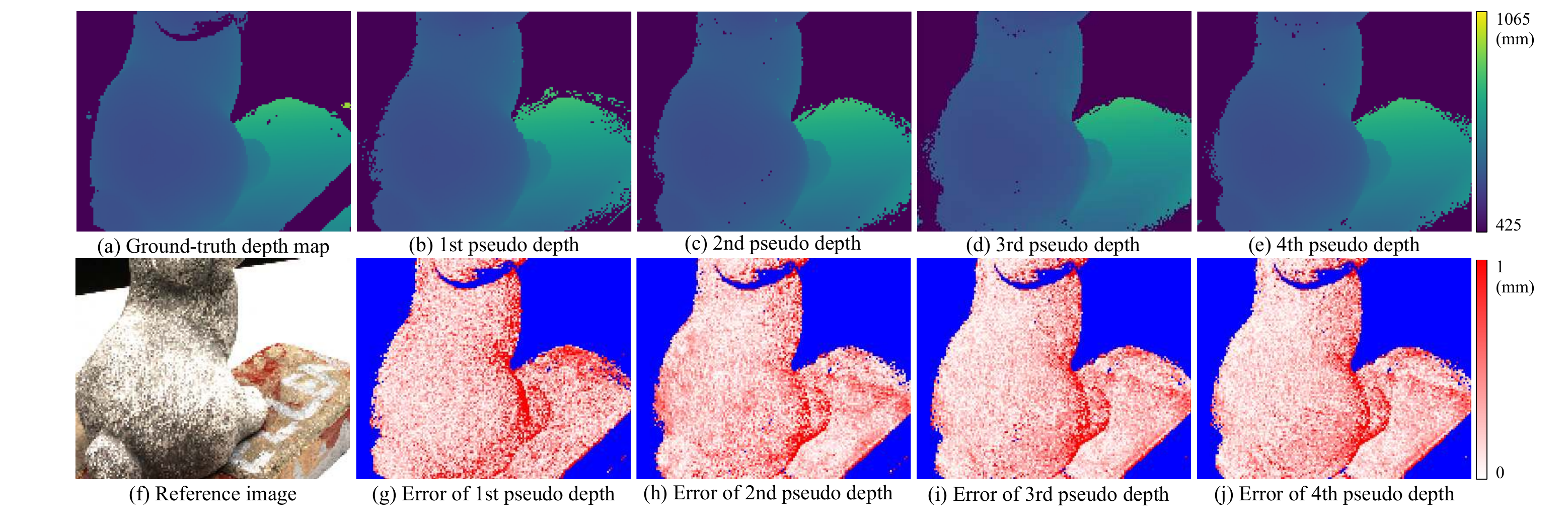}
    \end{center}
    \vspace{-0.5cm}
    \caption{Pseudo depth labels generated by each iteration of the iterative self-supervised learning framework. (a) Ground-truth depth map. (b-e) Pseudo depth labels generated by 1-4 iteration of the self-supervised learning framework. (f) Reference image. (g-j) Error of each iteration of the pseudo depth labels. Areas with no pseudo depth labels are marked as blue in the error visualization. Best viewed on screen.}
    \label{fig:pseudo_depth_itr}
    \vspace{-0.4cm}
\end{figure*}

\section{Pseudo labels generated on each iteration}\label{sec:pseudo_itr}
Fig.~\ref{fig:pseudo_depth_itr} shows visualization of pseudo depth labels generated by the different iterations of the self-supervised learning framework. As shown, pseudo depth labels tend to become stable after the second iteration.

\section{Additional Ablation Experiments}\label{sec:ablations}

\noindent\textbf{Use geometric MVS methods as initialization}.

We use a traditional MVS method OpenMVS to generate the pseudo labels to replace our unsupervised learning process on DTU dataset. Specifically, we render a depth map from the mesh generated by OpenMVS and treat it as the label for the first iteration of the iterative training. As shown in Tab.~\ref{table:openmvs_init}, using OpenMVS outputs as pseudo labels can achieve similar performance as our proposed initialization method after 3 iterations of self-supervised learning.

\begin{table}[!ht]
\vspace{0.1cm}
\begin{center}
\footnotesize
\begin{tabular}{ll|c|cccc}              
\hline
\multicolumn{2}{c|}{Method$\backslash$\textit{f-score}} & Init & Iter. 1 & Iter. 2 & Iter. 3\\
\hline\hline
& OpenMVS Init. & 73.70\% & 76.86\% & 87.95\% & 88.02\%\\\hline
& Ours Init. & 77.06\% & 88.16\%  & 88.42\% & 88.49\% \\\hline
\end{tabular}
\end{center}
\vspace{-0.3cm}
\caption{Performance with different initialization method.}
\label{table:openmvs_init}
\vspace{-0.5cm}
\end{table}

\section{Discussions}\label{sec:discussion}

\noindent\textbf{Runtime}
Despite the limitation on texture-less area mentioned in main paper, another limitation appears on the runtime of proposed self-supervised learning method. Each iteration of the self-training process takes around 15 hours on our machine, which adds up to days for several iterations of self-training or fine-tuning on novel data. For comparison, the supervised CVP-MVSNet takes around 10 hours. A classical MVS method such as the OpenMVS even does not need any training to achieve compromised results. Improving the efficiency of the proposed learning method can be an direction of future research.

\noindent\textbf{Limitation of geometric processing}
Another concern appears on the geometric filtering and fusion methods we used to refine the pseudo depth label. The traditional geometric methods such as consistency check and Screened Poission Surface Reconstruction have their limitations. Specifically, the SPSR is a non-learning, indifferentiable and time-consuming step, which might be too heavy for fine-tuning on novel data. It also have limited performance on very complex scenes. Improving pseudo label processing methods can be a direction of future research.


\end{document}